\pdfoutput=1

\documentclass[11pt]{article}

\usepackage[final]{acl}

\usepackage{times}
\usepackage{latexsym}

\usepackage[T1]{fontenc}

\usepackage[utf8]{inputenc}

\usepackage{microtype}

\usepackage{inconsolata}

\usepackage{graphicx}

\usepackage{caption}               
\usepackage{subcaption}
\usepackage{threeparttable}        
\usepackage{wrapfig}               
\usepackage{booktabs}
\usepackage{multirow}
\usepackage{xcolor}
\usepackage{bbding}                
\usepackage{lipsum}                
\usepackage[cjk]{kotex}            
\usepackage{amsmath}               
\usepackage{accents}               
\usepackage{multirow}              
\usepackage{mathtools}             
\usepackage{footnote}
\usepackage{tablefootnote}
\usepackage{subfiles}
\usepackage{colortbl}              
\usepackage{xcolor}                
\usepackage{wrapfig}               
\usepackage{amssymb}               
\usepackage{comment}               
\usepackage{cleveref}              
\usepackage{pifont}
\usepackage{tocbibind}             
\usepackage{appendix}              
\usepackage[export]{adjustbox}     

\usepackage{tikz}
\usepackage{pgfplots}
\usetikzlibrary{spy,calc}
\usepackage{hyperref}

\newif\ifblackandwhitecycle
\gdef\patternnumber{0}

\pgfkeys{/tikz/.cd,
    zoombox paths/.style={
        draw=orange,
        very thick
    },
    black and white/.is choice,
    black and white/.default=static,
    black and white/static/.style={ 
        draw=white,   
        zoombox paths/.append style={
            draw=white,
            postaction={
                draw=black,
                loosely dashed
            }
        }
    },
    black and white/static/.code={
        \gdef\patternnumber{1}
    },
    black and white/cycle/.code={
        \blackandwhitecycletrue
        \gdef\patternnumber{1}
    },
    black and white pattern/.is choice,
    black and white pattern/0/.style={},
    black and white pattern/1/.style={    
            draw=white,
            postaction={
                draw=black,
                dash pattern=on 2pt off 2pt
            }
    },
    black and white pattern/2/.style={    
            draw=white,
            postaction={
                draw=black,
                dash pattern=on 4pt off 4pt
            }
    },
    black and white pattern/3/.style={    
            draw=white,
            postaction={
                draw=black,
                dash pattern=on 4pt off 4pt on 1pt off 4pt
            }
    },
    black and white pattern/4/.style={    
            draw=white,
            postaction={
                draw=black,
                dash pattern=on 4pt off 2pt on 2 pt off 2pt on 2 pt off 2pt
            }
    },
    zoomboxarray inner gap/.initial=5pt,
    zoomboxarray columns/.initial=2,
    zoomboxarray rows/.initial=2,
    subfigurename/.initial={},
    figurename/.initial={zoombox},
    zoomboxarray/.style={
        execute at begin picture={
            \begin{scope}[
                spy using outlines={%
                    zoombox paths,
                    width=\imagewidth / \pgfkeysvalueof{/tikz/zoomboxarray columns} - (\pgfkeysvalueof{/tikz/zoomboxarray columns} - 1) / \pgfkeysvalueof{/tikz/zoomboxarray columns} * \pgfkeysvalueof{/tikz/zoomboxarray inner gap} -\pgflinewidth,
                    height=\imageheight / \pgfkeysvalueof{/tikz/zoomboxarray rows} - (\pgfkeysvalueof{/tikz/zoomboxarray rows} - 1) / \pgfkeysvalueof{/tikz/zoomboxarray rows} * \pgfkeysvalueof{/tikz/zoomboxarray inner gap}-\pgflinewidth,
                    magnification=3,
                    every spy on node/.style={
                        zoombox paths
                    },
                    every spy in node/.style={
                        zoombox paths
                    }
                }
            ]
        },
        execute at end picture={
            \end{scope}
            \node at (image.north) [anchor=north, inner sep=0pt] {
                \subcaptionbox{
                    \label{\pgfkeysvalueof{/tikz/figurename}-image}
                }{\phantomimage}
            };
            \node at (zoomboxes container.north) [anchor=north, inner sep=0pt] {
                \subcaptionbox{
                    \label{\pgfkeysvalueof{/tikz/figurename}-zoom}
                }{\phantomimage}
            };
     \gdef\patternnumber{0}
        },
        spymargin/.initial=0.5em,
        zoomboxes xshift/.initial=1,
        zoomboxes right/.code=\pgfkeys{/tikz/zoomboxes xshift=1},
        zoomboxes left/.code=\pgfkeys{/tikz/zoomboxes xshift=-1},
        zoomboxes yshift/.initial=0,
        zoomboxes above/.code={
            \pgfkeys{/tikz/zoomboxes yshift=1},
            \pgfkeys{/tikz/zoomboxes xshift=0}
        },
        zoomboxes below/.code={
            \pgfkeys{/tikz/zoomboxes yshift=-1},
            \pgfkeys{/tikz/zoomboxes xshift=0}
        },
        caption margin/.initial=4ex,
    },
    adjust caption spacing/.code={},
    image container/.style={
        inner sep=0pt,
        at=(image.north),
        anchor=north,
        adjust caption spacing
    },
    zoomboxes container/.style={
        inner sep=0pt,
        at=(image.north),
        anchor=north,
        name=zoomboxes container,
        xshift=\pgfkeysvalueof{/tikz/zoomboxes xshift}*(\imagewidth+\pgfkeysvalueof{/tikz/spymargin}),
        yshift=\pgfkeysvalueof{/tikz/zoomboxes yshift}*(\imageheight+\pgfkeysvalueof{/tikz/spymargin}+\pgfkeysvalueof{/tikz/caption margin}),
        adjust caption spacing
    },
    calculate dimensions/.code={
        \pgfpointdiff{\pgfpointanchor{image}{south west} }{\pgfpointanchor{image}{north east} }
        \pgfgetlastxy{\imagewidth}{\imageheight}
        \global\let\imagewidth=\imagewidth
        \global\let\imageheight=\imageheight
        \gdef\columncount{1}
        \gdef\rowcount{1}
        
    },
    image node/.style={
        inner sep=0pt,
        name=image,
        anchor=south west,
        append after command={
            [calculate dimensions]
            node [image container,subfigurename=\pgfkeysvalueof{/tikz/figurename}-image] {\phantomimage}
            node [zoomboxes container,subfigurename=\pgfkeysvalueof{/tikz/figurename}-zoom] {\phantomimage}
        }
    },
    color code/.style={
        zoombox paths/.append style={draw=#1}
    },
    connect zoomboxes/.style={
    spy connection path={\draw[draw=none,zoombox paths] (tikzspyonnode) -- (tikzspyinnode);}
    },
    help grid code/.code={
        \begin{scope}[
                x={(image.south east)},
                y={(image.north west)},
                font=\footnotesize,
                help lines,
                overlay
            ]
            \foreach \x in {0,1,...,9} { 
                \draw(\x/10,0) -- (\x/10,1);
                \node [anchor=north] at (\x/10,0) {0.\x};
            }
            \foreach \y in {0,1,...,9} {
                \draw(0,\y/10) -- (1,\y/10);                        \node [anchor=east] at (0,\y/10) {0.\y};
            }
        \end{scope}    
    },
    help grid/.style={
        append after command={
            [help grid code]
        }
    },
}

\newcommand\phantomimage{%
    \phantom{%
        \rule{\imagewidth}{\imageheight}%
    }%
}
\newcommand\zoombox[2][]{
    \begin{scope}[zoombox paths]
        \pgfmathsetmacro\xpos{
            (\columncount-1)*(\imagewidth / \pgfkeysvalueof{/tikz/zoomboxarray columns} + \pgfkeysvalueof{/tikz/zoomboxarray inner gap} / \pgfkeysvalueof{/tikz/zoomboxarray columns} ) + \pgflinewidth
        }
        \pgfmathsetmacro\ypos{
            (\rowcount-1)*( \imageheight / \pgfkeysvalueof{/tikz/zoomboxarray rows} + \pgfkeysvalueof{/tikz/zoomboxarray inner gap} / \pgfkeysvalueof{/tikz/zoomboxarray rows} ) + 0.5*\pgflinewidth
        }
        \edef\dospy{\noexpand\spy [
            #1,
            zoombox paths/.append style={
                black and white pattern=\patternnumber
            },
            every spy on node/.append style={#1},
            x=\imagewidth,
            y=\imageheight
        ] on (#2) in node [anchor=north west] at ($(zoomboxes container.north west)+(\xpos pt,-\ypos pt)$);}
        \dospy
        \pgfmathtruncatemacro\pgfmathresult{ifthenelse(\columncount==\pgfkeysvalueof{/tikz/zoomboxarray columns},\rowcount+1,\rowcount)}
        \global\let\rowcount=\pgfmathresult
        \pgfmathtruncatemacro\pgfmathresult{ifthenelse(\columncount==\pgfkeysvalueof{/tikz/zoomboxarray columns},1,\columncount+1)}
        \global\let\columncount=\pgfmathresult
        \ifblackandwhitecycle
            \pgfmathtruncatemacro{\newpatternnumber}{\patternnumber+1}
            \global\edef\patternnumber{\newpatternnumber}
        \fi
    \end{scope}
}

\definecolor{darkergreen}{RGB}{21, 152, 56}
\definecolor{red2}{RGB}{252, 54, 65}
\definecolor{Gray}{gray}{0.6}
\definecolor{LavenderBlush}{rgb}{1.0, 0.94, 0.96}

\newcommand{\yesmark}{\textcolor{darkergreen}{\ding{52}}}
\newcommand{\nomark}{\textcolor{red2}{\ding{56}}}

\usepackage{titletoc}    

\def\addcontentsline#1#2#3{}

\hypersetup{
    colorlinks=true
}
\usepackage{url}
\usepackage{cuted}                 

\usepackage{tabularx}    

\usepackage{algorithm}
\usepackage{algorithmicx}
\usepackage{algpseudocode}

\usepackage{minitoc}

\usepackage[most]{tcolorbox} 
\definecolor{myDarkGreen}{HTML}{0A542D}   
\definecolor{myLightGreen}{HTML}{EDFBD2}  

\usepackage{newfloat}
\usepackage{listings}
\DeclareCaptionStyle{ruled}{labelfont=normalfont,labelsep=colon,strut=off} 
\floatstyle{ruled}
\newfloat{listing}{tb}{lst}{}
\floatname{listing}{Listing}
\AtBeginEnvironment{tcolorbox}{\small}

\title{PosterForest: Hierarchical Multi-Agent Collaboration\\for Scientific Poster Generation}

\author {
    Jiho~Choi\textsuperscript{\textrm{1}}\thanks{Equal contribution},
    Seojeong~Park\textsuperscript{\textrm{1}}\footnotemark[1],
    Seongjong~Song\textsuperscript{\textrm{2}},
    Hyunjung~Shim\textsuperscript{\textrm{1}}\thanks{Corresponding author} \\
    \textsuperscript{\textrm{1}}Graduate School of Artificial Intelligence, KAIST, Republic of Korea \\
    \textsuperscript{\textrm{2}}School of Integrated Technology, Yonsei University, Republic of Korea \\
    {\tt\small {\{jihochoi, seojeong.park, kateshim\}@kaist.ac.kr, \{bell\}@yonsei.ac.kr}} \\
}

\begin{document}

\maketitle

\begin{abstract}
    Automating scientific poster generation requires hierarchical document understanding and coherent content-layout planning.
    Existing methods often rely on flat summarization or optimize content and layout separately.
    As a result, they often suffer from information loss, weak logical flow, and poor visual balance.
    We present \textbf{PosterForest}, a training-free framework for scientific poster generation.
    Our method introduces the \emph{Poster Tree}, a structured intermediate representation that captures document hierarchy and visual-textual semantics across multiple levels.
    Building on this representation, content and layout agents perform hierarchical reasoning and recursive refinement, progressively optimizing the poster from global organization to local composition.
    This joint optimization improves semantic coherence, logical flow, and visual harmony.
    Experiments show that PosterForest outperforms prior methods in both automatic and human evaluations, without additional training or domain-specific supervision.
    Code: \url{https://github.com/kaist-cvml/poster-forest}
\end{abstract}

\section{Introduction}
\label{sec:Introduction}

With the rapid advancement of science and technology, there has been an exponential increase~\cite{hanson2024strain, larsen2010rate} in the number of academic papers and technical reports with complex structures.
As these documents are often difficult to interpret quickly, readers are required to invest significant time and cognitive resources to understand their main arguments.
In this context, scientific posters have emerged as an effective medium for summarizing and presenting complex information in a concise and visually intuitive manner.
By combining textual and visual elements, posters facilitate more accessible communication of technical content.
However, manually crafting high-quality posters is a labor-intensive process that requires both domain knowledge and design expertise~\cite{qiang2019learning_PGM2,wang2024scipostlayout}.
Automating scientific poster generation (SPG) is therefore a critical research problem, as it can accelerate the dissemination of specialized knowledge and reduce the burden on researchers.

\begin{figure}[t]
    \centering
    \begin{subfigure}[t]{0.49\linewidth}
        \centering
        \fbox{\includegraphics[width=0.95\linewidth]{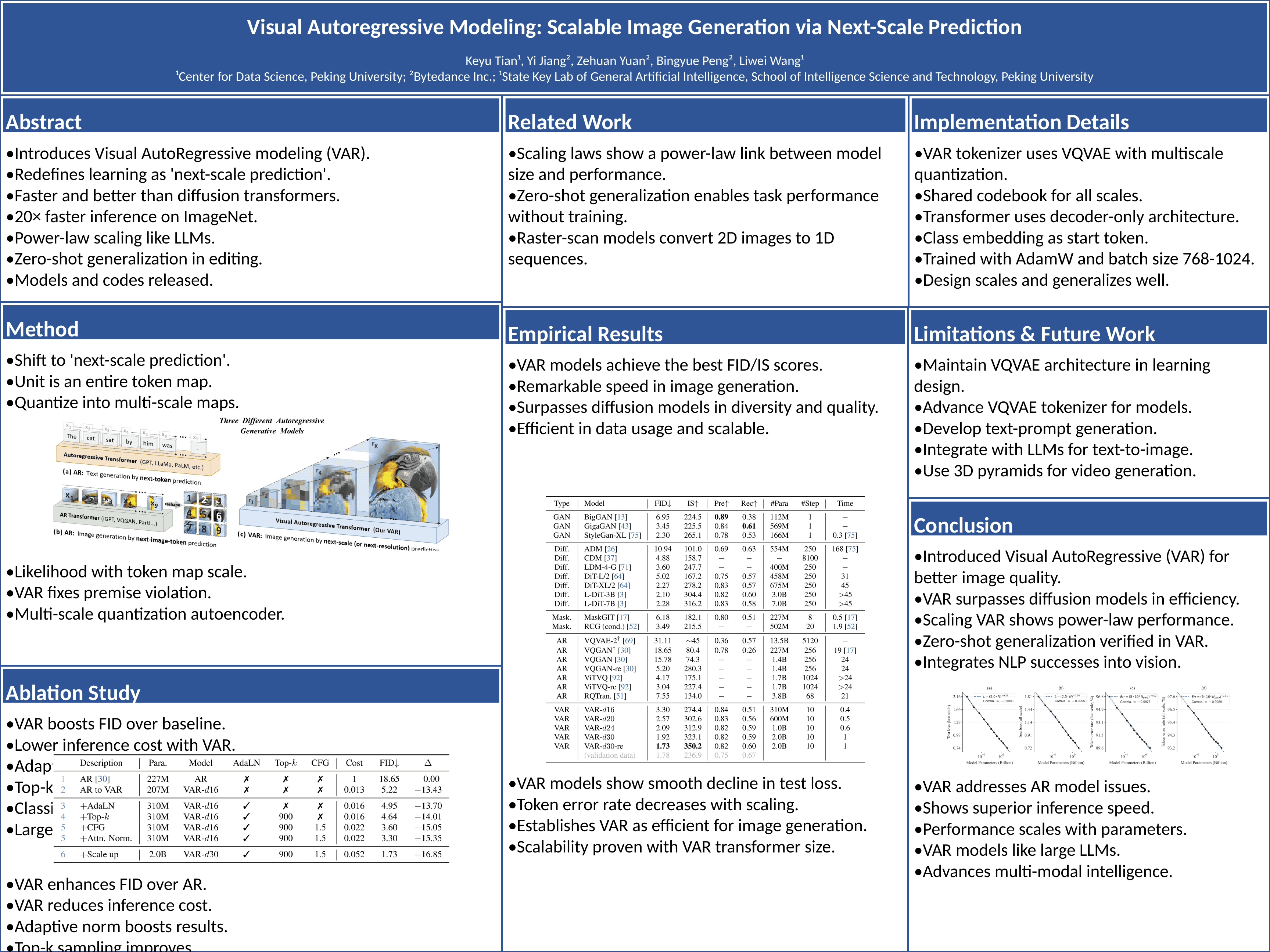}}
        \caption{\normalsize Paper2Poster}
        \label{subfig:limitation_Paper2Poster}
    \end{subfigure}
    \hfill
    \begin{subfigure}[t]{0.49\linewidth}
        \centering
        \fbox{\includegraphics[width=0.95\linewidth]{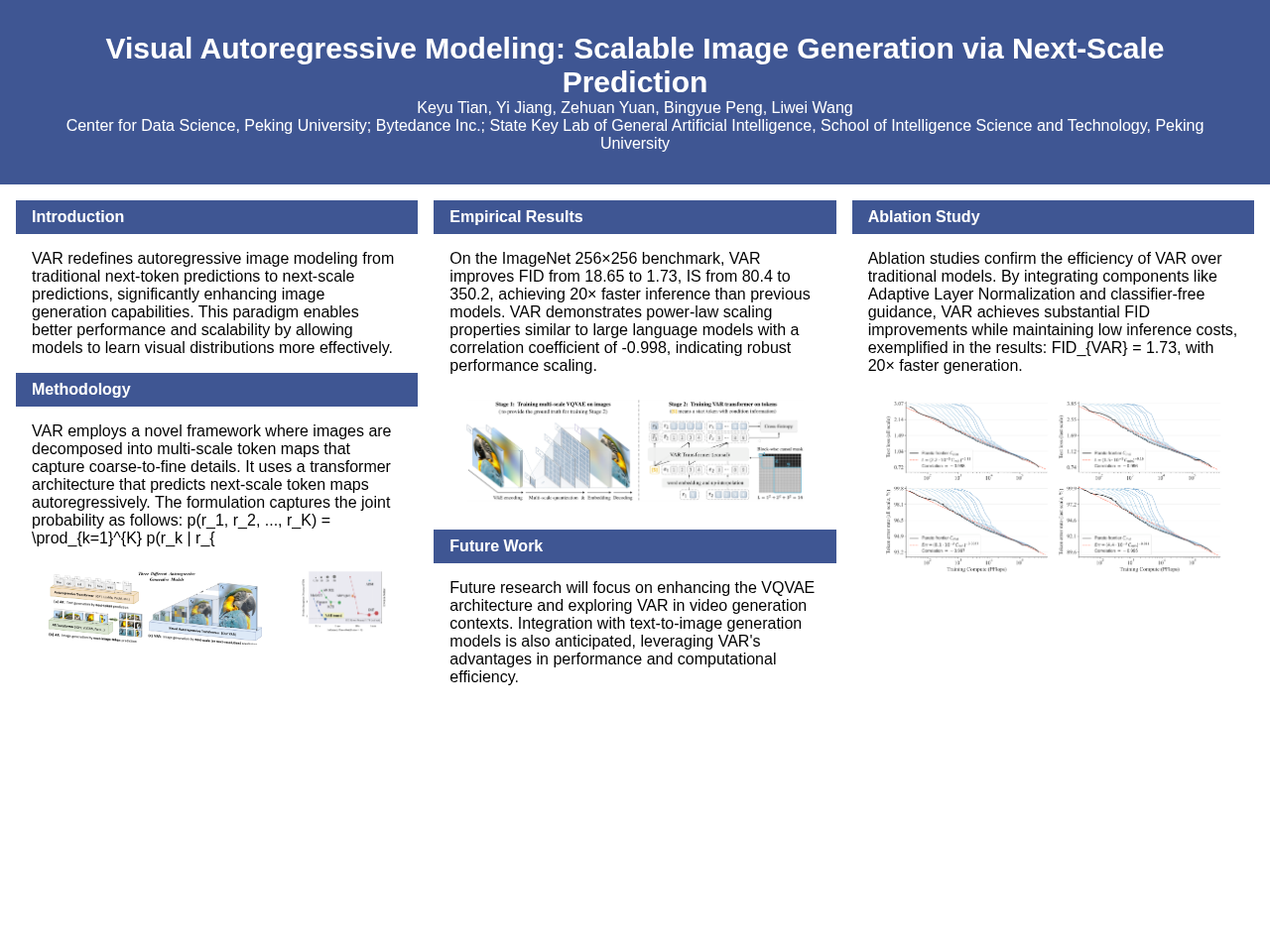}}
        \caption{\normalsize P2P}
        \label{subfig:limitation_P2P}
    \end{subfigure}
    \caption{
        \textbf{Limitations of Current SPG Methods.}
        Existing state-of-the-art scientific poster generation (SPG) methods, including P2P~\cite{sun2025p2p} and Paper2Poster~\cite{pang2025paper2poster}, lack \textit{hierarchical} document understanding, resulting in errors in both \textit{content} and \textit{layout}.
        (a) shows an example where an experiment table is incorrectly placed in the conclusion section.
        (b) illustrates an overly simplified poster, where paragraphs are merely summarized and assigned to fixed panels with fixed-sized figures.
    }
    \label{fig:intro_limitations_p2p}
    \vspace{-5mm}
\end{figure}



Pioneering works such as PGM~\cite{qiang2016learning_PGM, qiang2019learning_PGM2}, NCE~\cite{xu2021neural_NCE}, and PostDoc~\cite{jaisankar2024postdoc} approached automated poster generation by extracting text and figures from scientific documents and heuristically arranging them within poster panels. 
However, these approaches rely on fixed rules and struggle to handle the complexity of long, structured documents and the interplay between textual and visual content. 
To address this, more recent methods, including P2P~\cite{sun2025p2p} and Paper2Poster~\cite{pang2025paper2poster}, adopt multi-agent pipelines that decompose the task into specialized sub-problems such as parsing, content summarization, layout planning, and rendering. 
This modular design improves flexibility and coordination across stages, but typically requires explicit model training, such as instruction tuning or regressor-based optimization.


Despite these advances, current approaches suffer from several critical limitations.
(1) \textbf{Shallow Document Understanding:} They primarily depend on surface-level text features, lacking a deep grasp of the hierarchical structure inherent to scientific documents and the semantic associations between textual and visual components.
Consequently, they often exhibit an interrupted logical flow and weak integration of visual elements as in~\Cref{fig:intro_limitations_p2p} (a), ultimately reducing the effectiveness of posters in conveying information quickly and accurately.
This limitation significantly increases users’ cognitive load.
(2) \textbf{Weak Content-Layout Integration:} Existing approaches often adopt a sequential pipeline in which the layout is determined before content placement.
This decoupled strategy overlooks the intrinsic interdependence between content and layout, treating them as isolated components rather than pursuing their integrated organization.
As a result, critical content may be truncated or misplaced, and the logical flow between textual and visual elements is frequently disrupted.
Moreover, as shown in~\Cref{fig:intro_limitations_p2p} (b), this often results in posters that are overly simplified and fail to capture the complexity of the original document, diminishing their practical value.
It diminishes the practical value of automated poster generation systems. 
(3) \textbf{Training Overhead:} Existing methods requiring instruction tuning or regression training pipelines add complexity and resource demands, limiting practical deployment.


In this study, we aim to address these limitations, which overlook both the hierarchical organization of scientific documents and the semantic alignment between content and layout.
Such limitations hinder holistic understanding and often result in reduced clarity, visual incoherence, and discrepancies between visual elements and their explanatory context.
To address these limitations, we present \textit{PosterForest}, a novel framework with two core components. 
For limitation (1), we introduce the \textit{Poster Tree}, a hierarchical intermediate representation.
It prunes and merges document content across the section–subsection–paragraph hierarchy to preserve salient information while reducing redundancy, and explicitly links text with figures and tables.
Each node jointly encodes \textbf{content} and \textbf{layout} attributes.
This representation, tailored for scientific poster generation, preserves logical flow and strengthens text–visual associations.



For limitation (2), we propose hierarchical modification planning with multi-agent collaboration. 
Scientific poster generation requires balancing multiple interdependent objectives, such as content fidelity, layout efficiency, and visual coherence, which are difficult to optimize jointly within a single reasoning process.
To address this, we decompose the task into specialized roles and employ multiple agents that focus on complementary aspects, including content summarization, layout planning, and visual material placement across different levels of the hierarchy.
Through iterative coordination and feedback, these agents collaboratively refine the Poster Tree, enabling effective joint optimization of both content and structure. 
This results in visually balanced and semantically coherent posters with improved integration of textual and visual information. 
Finally, addressing limitation (3), the proposed pipeline is training-free and relies only on standard APIs and publicly available checkpoints, enabling practical deployment.



Overall, \textit{PosterForest} delivers high-quality summarization, visual coherence, and consistent information flow, overcoming the limitations of prior methods. 
Extensive experiments show that our method consistently outperforms prior approaches in both automatic and human evaluations, achieving up to 59.2\% preference in human studies (vs. 27.2\% for prior work) while remaining training-free.

\begin{figure*}[t]
    \centering
    \includegraphics[width=1.0\textwidth]{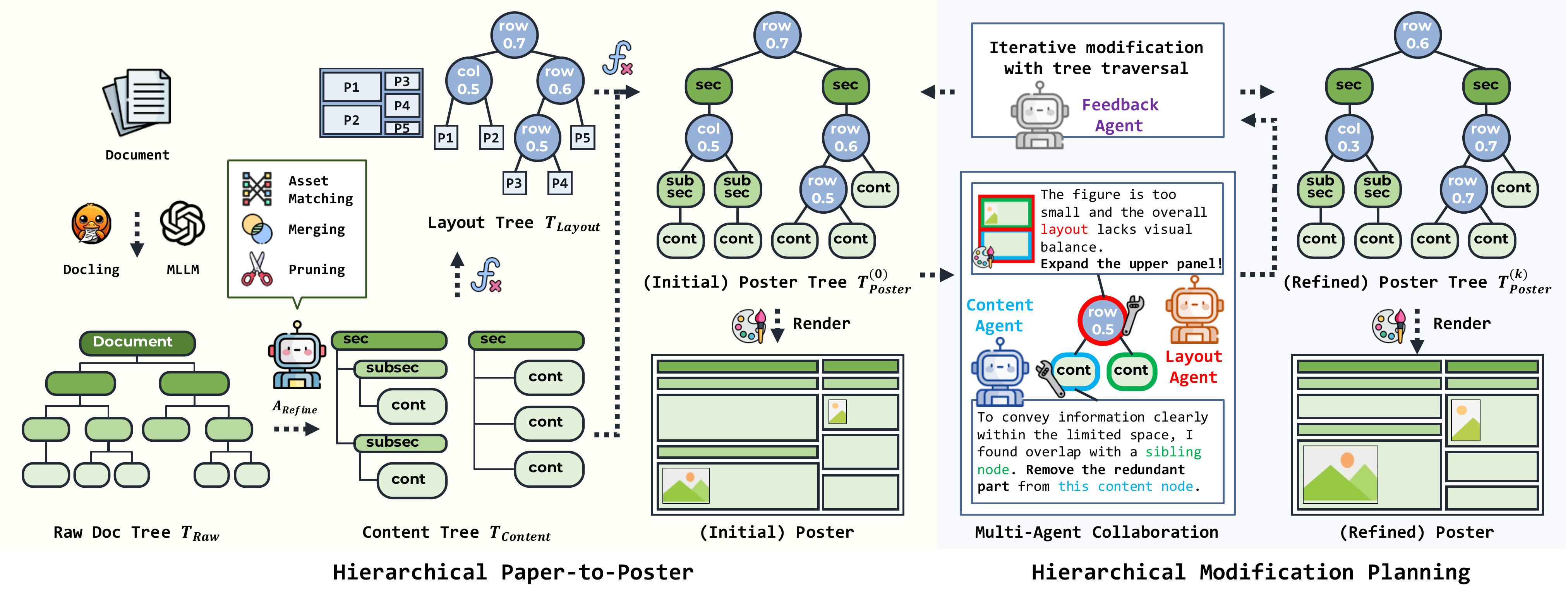}
    \caption{
        \textbf{Overview of PosterForest.}
        PosterForest first constructs a hierarchical Poster Tree that integrates document semantics and layout (\Cref{subsec:Hierarchical Paper-to-Poster}), then iteratively refines it through collaboration between agents to optimize structure and visual coherence (\Cref{subsec:Hierarchical Modification Planning}).
    }
    \label{fig:overall}
\end{figure*}

\section{Related Work}
\label{sec:Related Work}

\textbf{Scientific Poster Generation.}
Although generic (i.e., movie, commercial) poster generation is a pervasively researched topic~\cite{gupta2021layouttransformer, li2020attribute, zheng2023layoutdiffusion, inoue2023layoutdm, gao2025postermaker,hsu2025postero}, scientific poster generation is more challenging and crucial due to its deliverability of extensive information and reasoning.
It has been studied as a layout-driven summarization problem~\cite{qiang2016learning_PGM, qiang2019learning_PGM2}, where key elements such as panel size, position, and hierarchy are learned from examples.
Earlier work, such as \citet{xu2021neural_NCE}, emphasized the importance of content extraction, proposing a pipeline to select representative text and visuals. An instruction-tuning based P2P~\cite{sun2025p2p} and a regressor fitting-based Paper2Poster~\cite{pang2025paper2poster} are recent approaches of introducing LLM-based multi-agent frameworks to handle parsing, planning, and rendering in a modular way, extending the framework to generate slides such as PPTAgent~\cite{zheng2025pptagent}.
These methods are supported by new benchmarks~\cite{wang2024scipostlayout, saxena2025postersum} and evaluation protocols utilizing vision-language models~\cite{lee2024prometheus}, enabling fine-grained assessment of visual coherence and content fidelity.
While these approaches achieve baseline-level automation, they often treat sections independently, simply mapping them to panels.
As a result, the semantic flow and hierarchical connections across sections in the original paper are not well preserved in the generated poster.

\noindent
\textbf{Hierarchical Document Parsing and Understanding.}
Scientific poster generation requires a robust understanding of document structure and content, compared to naive rank-based text extraction \cite{mihalcea2004textrank}.
Early approaches to document understanding leveraged hierarchical parsing methods, such as DocParser~\cite{rausch2021docparser} and PDF-to-Tree~\cite{zhang2024tree}, to recover logical section structures from rendered pages.
These techniques enable semantic segmentation of long documents, which is crucial for downstream tasks like summarization and visualization.
More recent works employ pre-trained multimodal models~\cite{huang2022layoutlmv3, lin2023layoutprompter} or graph-based representations~\cite{gemelli2022doc2graph} to jointly model textual and visual elements. OWL~\cite{hu2025owl} integrates multiple LLMs for reasoning and understanding documents.
Such advances form the basis for extracting salient content needed for poster generation.




\noindent
\textbf{Multi-Agent Reasoning and Collaboration.}
Single-agent reasoning techniques such as Chain-of-Thought (CoT) \cite{wei2022chain} and its variations \cite{yao2023tree, besta2024graph, chen2022program, gao2023pal} enabled logical thinking of models, representatively academic \cite{team2024gemma, zhang2023multimodal} and mathematical \cite{shao2024deepseekmath} reasoning.
Recent researches take a step forward and explore multi-agent collaboration to further enhance problem-solving capabilities.
Instead of relying on a single reasoning path, multi-agent systems assign specialized roles to LLM agents and promote iterative feedback, critique, and coordination~\cite{li2023camel, zhang2024chain, hong2023metagpt, tran2025multi_collaboration, li2024chain}, successfully simulating collaborative software engineers \cite{qian2023chatdev}, or peer-reviewers of scientific papers \cite{yu2024researchtown} even generating code from papers \cite{seo2025paper2code}.
In the poster generation, this paradigm enables specialized agents to perform document analysis, content summarization, and layout composition~\cite{sun2025p2p, pang2025paper2poster}.

\section{Proposed Method}
\label{sec:Proposed Method}


\subsection{Preliminaries}

Recent advances in scientific poster generation (SPG), including P2P~\cite{sun2025p2p} and Paper2Poster~\cite{pang2025paper2poster}, have introduced multi-agent pipelines that automatically synthesize posters from research papers.
These methods leverage multimodal large language models (MLLMs), such as GPT-4~\cite{achiam2023_gpt4} and Qwen~\cite{bai2023_qwen}, to extract textual and visual content, summarize key information, and organize it into structured panel layouts.

Among them, Paper2Poster adopts a modular approach.
It comprises:
(a) a parser that constructs an asset library of textual and visual elements,
(b) a planner that matches text content with relevant figures and tables, and
(c) a painter-commenter loop that iteratively refines contents inside the panel through vision-language feedback.
P2P further leverages instruction tuning to enhance coordination among multiple agents, whereas Paper2Poster employs regressor-based learning to optimize content arrangement and visual composition.

Despite their effectiveness, existing approaches typically treat scientific papers as linear text sequences, disregarding structural relationships among textual units and semantic alignment between text and visuals.
Consequently, they capture only shallow associations within and across modalities.
Structural cues such as section and subsection boundaries, paragraph-level semantics, and cross-references to figures and tables are often underutilized or entirely ignored.
These limitations result in logical discontinuities across the content and weakened correspondence among different elements, ultimately diminishing overall clarity and informativeness.

To address these limitations, we introduce \textbf{PosterForest}, a training-free framework for SPG as shown in~\Cref{fig:overall}.
PosterForest operates in two main stages: (1) constructing a hierarchical \textbf{Poster Tree} that jointly encodes the document’s semantic content and the poster’s layout structure, and (2) iteratively refining this \textit{Poster Tree} through multi-agent collaboration between specialized content and layout experts.
The following sections provide a detailed description of each stage.


\subsection{Hierarchical Paper-to-Poster}
\label{subsec:Hierarchical Paper-to-Poster}


Given an input paper (or document) $\mathcal{D}$, we first parse it into a \textbf{Raw Doc Tree}, $\mathcal{T}_{\mathrm{Raw}}$.
We define $\mathcal{T}_{\mathrm{Raw}} = (\mathcal{V}_{\mathrm{Raw}}, \mathcal{E}_{\mathrm{Raw}})$ as a rooted tree that represents the structural hierarchy of the document contents.
Each node $v \in \mathcal{V}_{\mathrm{Raw}}$ corresponds to a document element such as title, section, subsection, paragraph, figure, or table, which contains its raw semantic content.
Each directed edge $(u \to v) \in \mathcal{E}_{\mathrm{Raw}}$ denotes a parent–child relation, indicating that $v$ is a subcomponent of $u$ in the hierarchy.
Let $\mathcal{A}_\mathrm{{Parser}}$ be a parsing agent that extracts this structure as:
\begin{equation}
    \mathcal{T}_{\mathrm{Raw}} \;=\; \mathcal{A}_\textsc{{Parse}} (D)\,.
    \label{eq:parse}
\end{equation}
The resulting tree explicitly captures both hierarchical organization and referential links, ensuring that figures and tables appear as children of the textual nodes that reference them.


We then refine $\mathcal{T}_{\mathrm{Raw}}$ into a \textbf{Content Tree}, $\mathcal{T}_{\mathrm{Content}} = (\mathcal{V}_{\mathrm{Content}}, \mathcal{E}_{\mathrm{Content}})$, which preserves the essential information to construct a scientific poster.
Guided by the MLLM agent, this process involves \textit{pruning} less important nodes, \textit{merging} redundant or closely related content, and \textit{summarizing} lengthy textual content, ensuring the resulting structure remains concise, coherent, and focused on key information.
During this process, the node-edge relationships are updated to reflect the revised structure.
Let $\mathcal{A}_{\mathrm{Refine}}$ be a content refinement agent that performs these operations as:
\begin{equation}
    \mathcal{T}_{\mathrm{Content}} = \mathcal{A}_{\textsc{Refine}} (\mathcal{T}_{\mathrm{Raw}}).
    \label{eq:raw_to_content}
\end{equation}
In $\mathcal{T}_{\mathrm{Content}}$, each node corresponds to a concise textual unit or a visual asset (e.g., figures and tables), with minor details removed. 
Each node ${c} \in \mathcal{V}_{\mathrm{Content}}$ is represented as ${c} = (t, s)$ with $t \in \mathcal{T}_{\mathrm{semantic}}$ denoting the semantic type (e.g., paragraph, figure, table) and $s \in \mathcal{S}_{\mathrm{semantic}}$ denoting the semantic content (summarized text, caption, or visual data).
This produces a compact and informative representation tailored for poster generation.


Based on the Content Tree, we establish a \textbf{Layout Tree}, $\mathcal{T}_{\mathrm{Layout}} = (\mathcal{V}_{\mathrm{Layout}}, \mathcal{E}_{\mathrm{Layout}})$, which specifies the poster’s spatial organization.
The Layout Tree follows a widely adopted approach~\cite{qiang2016learning_PGM, qiang2019learning_PGM2, pang2025paper2poster} in poster layout modeling, where the canvas is hierarchically partitioned into regions organized by rows and columns.
Unlike previous methods that derive such structures directly from the document layout, our Layout Tree is initialized from $\mathcal{T}_{\mathrm{Content}}$, inheriting the hierarchical relationships among content elements, and aligning the layout with the intended content structure of the poster as:
\begin{equation}
    \mathcal{T}_{\mathrm{Layout}} = \mathcal{O}_\textsc{Layout\_Init} (\mathcal{T}_{\mathrm{Content}}).
    \label{eq:layout_init}
\end{equation}
Each layout node $l \in \mathcal{V}_{\mathrm{Layout}}$ is represented as $l = (r, x)$ with $r \in \mathcal{R}_{\mathrm{spatial}}$ (region type: row split, column split, panel) and $x \in \mathcal{X}_{\mathrm{spatial}}$ (spatial attributes: normalized position, width, height, aspect ratio).
The initial allocation of regions is {deterministically} derived from content statistics. 


Finally, we integrate content and layout into a unified representation, the \textbf{Poster Tree}, $\mathcal{T}_{\mathrm{Poster}}$. 
We define $\mathcal{T}_{\mathrm{Poster}} = (\mathcal{V}_{\mathrm{Poster}}, \mathcal{E}_{\mathrm{Poster}})$ by merging the Content Tree, $\mathcal{T}_{\mathrm{Content}}$, and the Layout Tree, $\mathcal{T}_{\mathrm{Layout}}$, 
where each semantic node is mapped to a spatial region of the poster as:
\begin{equation}
    \mathcal{T}_{\mathrm{Poster}} = \mathcal{O}_\textsc{Merge} \big(\mathcal{T}_{\mathrm{Content}},\, \mathcal{T}_{\mathrm{Layout}}\big).
    \label{eq:merge}
\end{equation}
Each poster node $w \in \mathcal{V}_{\mathrm{Poster}}$ is a heterogeneous node that jointly encodes semantic attributes (e.g., a summarized paragraph or key visual element from $\mathcal{T}_{\mathrm{Content}}$) and spatial attributes placement from $\mathcal{T}_{\mathrm{Layout}}$. 
The merge operation aligns the hierarchical structure of the content with the corresponding layout partition, producing nodes that specify both \emph{what} information is displayed and \emph{where and how} it appears on the canvas.
This unified tree representation provides an inductive bias tailored for poster generation, as it tightly couples the logical document hierarchy with the visual layout structure.


\subsection{Hierarchical Modification Planning}
\label{subsec:Hierarchical Modification Planning}

After constructing the initial Poster Tree, $\mathcal{T}_\texttt{poster}^{(0)}$, we introduce a hierarchical refinement phase that is designed to jointly optimize both content quality and layout organization.
In contrast to prior methods~\cite{sun2025p2p, pang2025paper2poster} that fix the layout and subsequently adjust only the content, our approach traverses the heterogeneous nodes in Poster Tree and performs node-specific updates by leveraging both local attributes and hierarchical context, while further incorporating global evaluation to achieve a more coherent and polished final result.

\begin{figure}[t]
    \centering
    \includegraphics[width=0.47\textwidth]{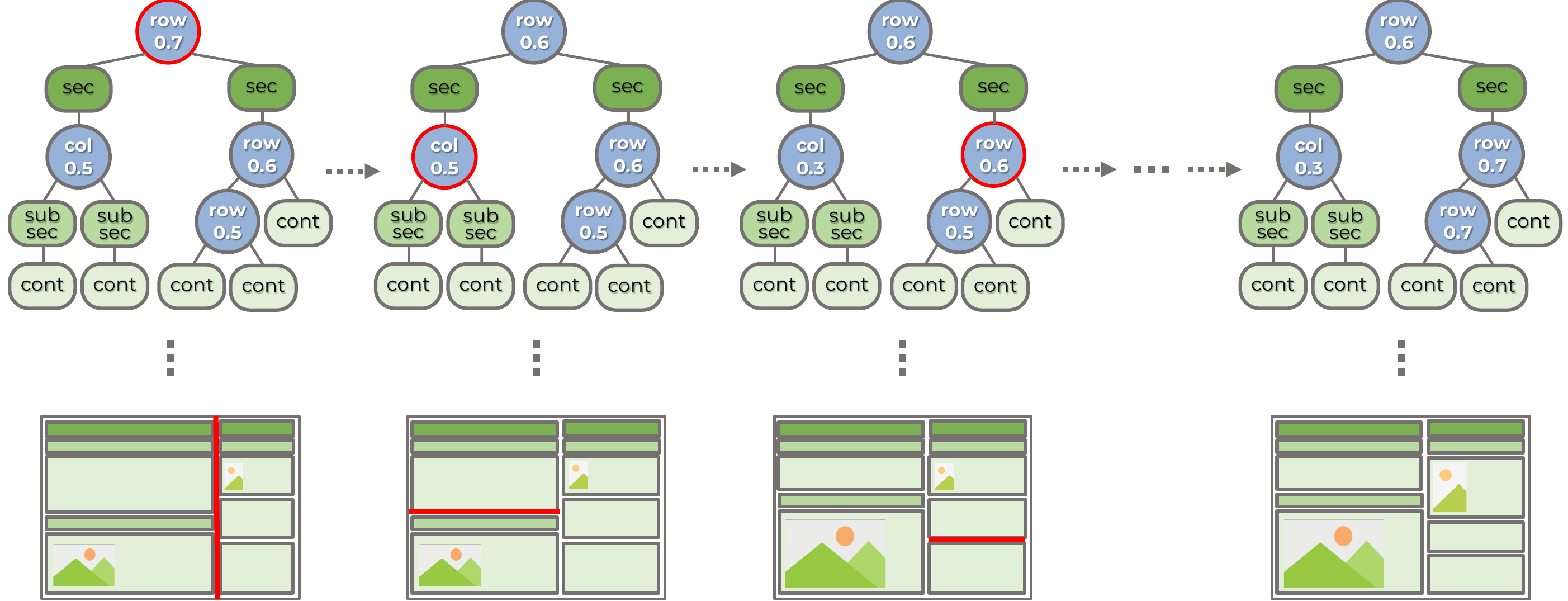}
    \caption{
        \textbf{Poster Tree Traversal (Node-level).}
        The Poster Tree and layout are iteratively updated through the shared decision of the layout and Content Agent.
    }
    \label{fig:main_iterative_refinement}
\end{figure}


\subsubsection{Poster Tree Traversal (Node-level)}

Given the initialized Poster Tree, $\mathcal{T}_{\mathrm{Poster}}^{(0)}$, refinement begins through a hierarchical traversal from the root toward the leaves as shown in~\Cref{fig:main_iterative_refinement}.
Each node is updated by jointly considering its intrinsic attributes, the propagated information from its parent, and the structural context defined by its descendants.
This top-down propagation ensures that modifications applied at higher levels are consistently reflected throughout the tree.

\paragraph{Layout Agent.}
For each {layout node} $l_i \in \mathcal{V}_{\mathrm{Poster}}$, the Layout Agent, $\mathcal{A}_\textsc{Layout}$ optimizes geometric attributes such as region ratios, alignment, and spatial distribution.
The optimization is performed by aggregating the structural and semantic statistics of all descendant nodes as:
\begin{equation}
    {l}_i^{(t+1)} =
    \mathcal{A}_\textsc{Layout}\big(
        {l}_i^{(t)}, \,
        \tilde{\mathcal{P}}(l_i^{(t)}), \,
        \mathcal{D}(l_i^{(t)})
    \big),
    \label{eq:layout_update}
\end{equation}
where $\tilde{\mathcal{P}}(l_i)$ represents the updated information of the parent node after its refinement, and $\mathcal{D}(l_i)$ denotes the set of all descendants of $l_i$.

\paragraph{Content Agent.}
For each {content node} $c_i \in \mathcal{V}_{\mathrm{Poster}}$, $\mathcal{A}_\textsc{Content}$ refines textual density and semantic abstraction by referencing both the updated configuration of its parent layout node and the descendant layout context of that parent:
\begin{equation}
    {c}_i^{(t+1)} =
    \mathcal{A}_\textsc{Content}\big(
        {c}_i^{(t)}, \,
        \tilde{\mathcal{P}}(c_i^{(t)}), \,
        \mathcal{D}(
            \mathcal{P}(c^{(t)}_{i})
        )\}
    \big),
    \label{eq:content_update}
\end{equation}
where $\tilde{\mathcal{P}}(c_i)$ represents the updated information of the parent node after its refinement, and $\mathcal{D}(\mathcal{P}(c_i))$ denotes the set of all descendants of the parent node of $c_i$.
This hierarchical dependency allows local content updates to reflect global layout constraints and parent-level refinements.

The traversal proceeds until every node has been updated once, yielding an intermediate tree $\mathcal{T}_{\mathrm{Poster}}^{(t+1)}$ that captures coherent modifications across both spatial and semantic dimensions.
The resulting representation serves as the basis for subsequent global evaluation.


\subsubsection{Iterative Tree Refinement (Tree-level)}

After completing full node-level traversal, PosterForest performs iterative refinement at the tree level to progressively enhance the overall poster structure.
Each tree-level iteration corresponds to a complete pass of node-level updates, resulting in a refined Poster Tree that jointly enhances semantic clarity and spatial organization. This process may be repeated up to a maximum of $K$ iterations to incrementally enhance layout quality and content coherence. In practice, we set $K = 2$, which empirically yields stable and visually balanced results with a single additional refinement step.

\paragraph{Global Feedback Agent.}
At each iteration $t$, the rendering of the current Poster Tree $\mathcal{T}_{\mathrm{Poster}}^{(t)}$ is evaluated by a multimodal large language model (MLLM) acting as a {Global Feedback Agent}, denoted as $\mathcal{A}_\textsc{Feedback}$.
This agent analyzes the poster’s visual organization, textual structure, and hierarchical balance, and provides structured global feedback to determine whether an additional tree-level traversal should be executed:
\begin{equation}
    \big[\hat{\mathcal{F}}_\textsc{Global}^{(t)},\, \pi_\textsc{Continue}^{(t)}\big]
    = \mathcal{A}_\textsc{Feedback}\!\big(\mathcal{T}_{\mathrm{Poster}}^{(t)}\big),
    \label{eq:mllm_feedback}
\end{equation}
where $\hat{\mathcal{F}}_\textsc{Global}^{(t)}$ denotes the structured global feedback extracted from the MLLM, and $\pi_\textsc{Continue}^{(t)} \in \{0,1\}$ is a binary signal indicating whether another refinement iteration should be performed.

If $\pi_\textsc{Continue}^{(t)} = 1$, the next tree-level traversal is triggered using the propagated feedback:
\begin{equation}
    \mathcal{T}_{\mathrm{Poster}}^{(t+1)} =
    \mathcal{O}_\textsc{Traverse}\!\big(
        \mathcal{T}_{\mathrm{Poster}}^{(t)}, \,
        \hat{\mathcal{F}}_\textsc{Global}^{(t)}
    \big),
    \label{eq:iterative_refine}
\end{equation}
where $\mathcal{O}_\textsc{Traverse}$ denotes one complete pass of the propagation-based node-level refinement defined in \Cref{eq:layout_update} and \Cref{eq:content_update}.
Otherwise, the iterative refinement loop terminates, and the final Poster Tree is obtained as $\mathcal{T}_{\mathrm{Poster}}^{*} = \mathcal{T}_{\mathrm{Poster}}^{(t)}$.

\begin{figure*}[ht]
    \begin{subfigure}[t]{0.245\textwidth}
        \centering
        \fbox{\includegraphics[valign=c,width=\textwidth]{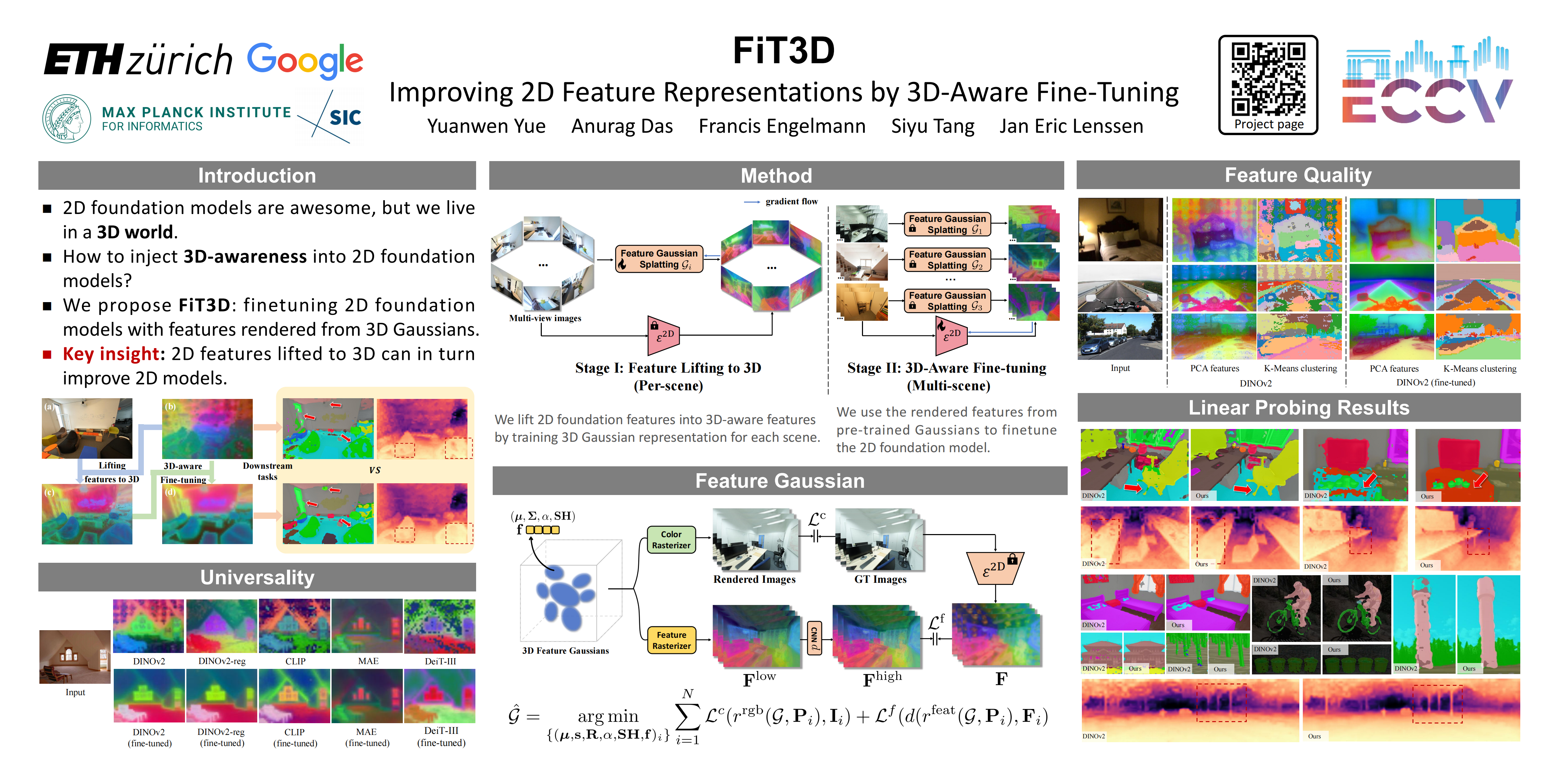}}
    \end{subfigure}
    \begin{subfigure}[t]{0.245\textwidth}
        \centering
        \fbox{\includegraphics[valign=c,width=\textwidth]{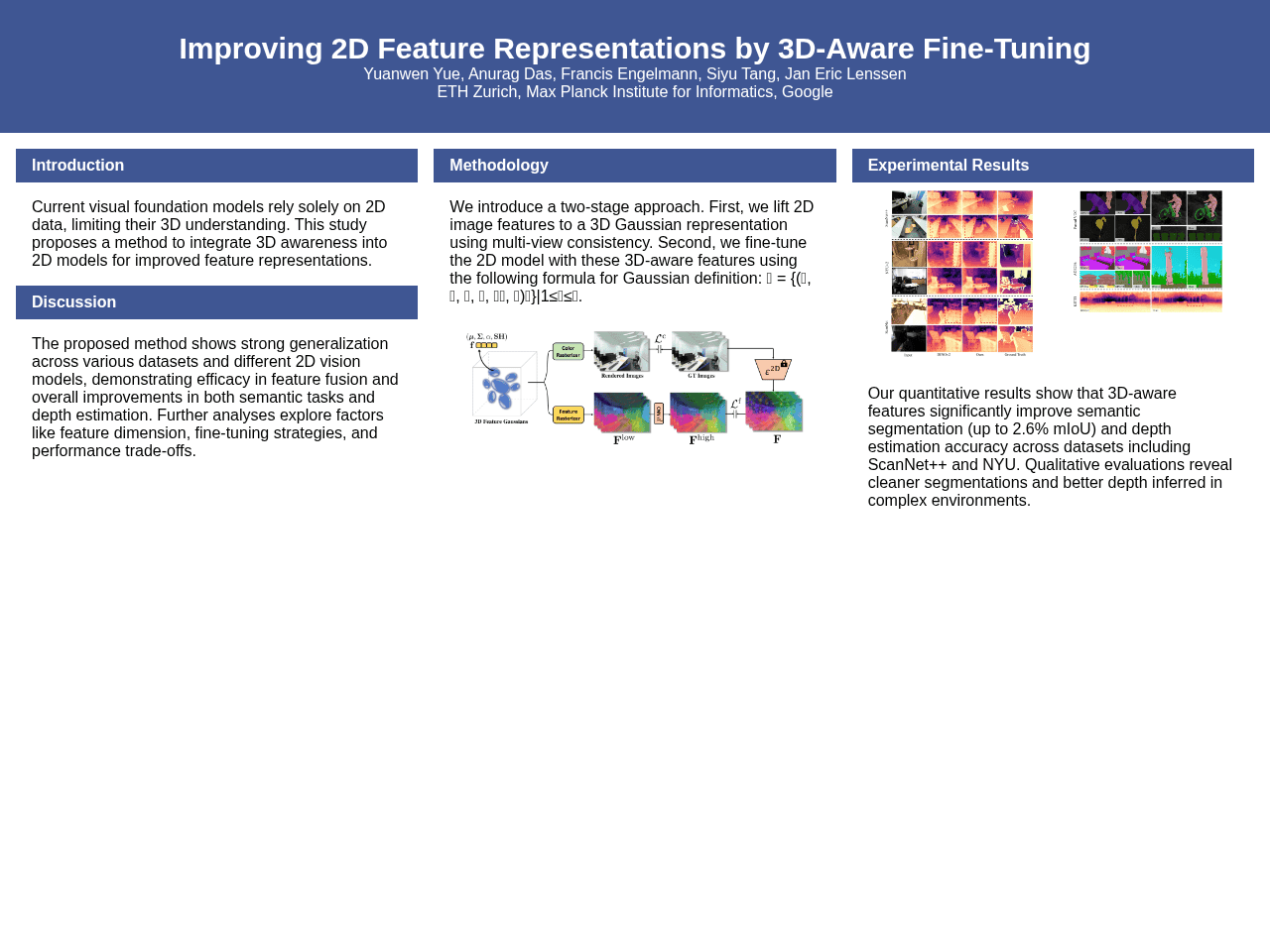}}
    \end{subfigure}
    \begin{subfigure}[t]{0.245\textwidth}
        \centering
        \fbox{\includegraphics[valign=c,width=\textwidth]{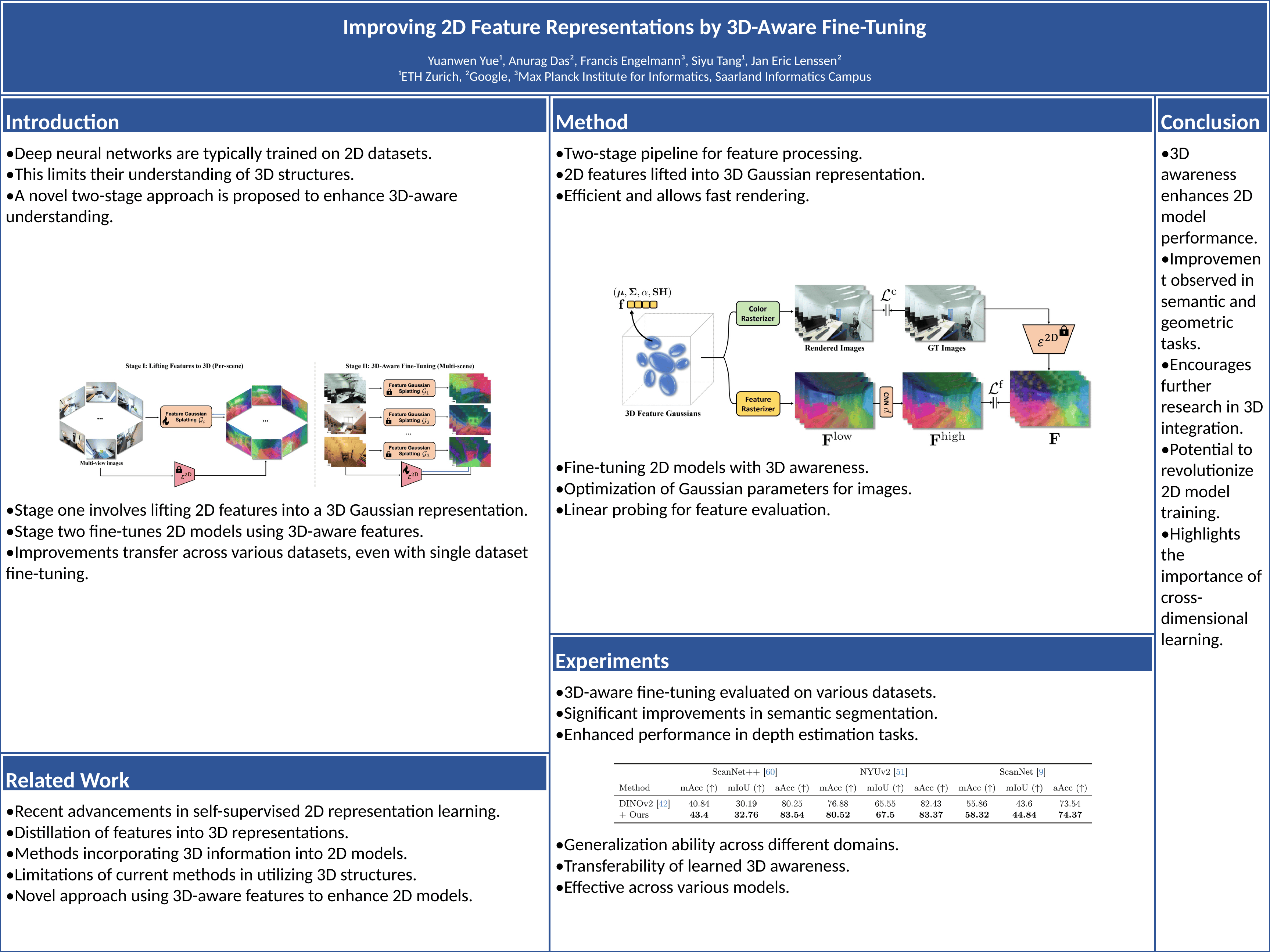}}
    \end{subfigure}
    \begin{subfigure}[t]{0.245\textwidth}
        \centering
        \fbox{\includegraphics[valign=c,width=\textwidth]{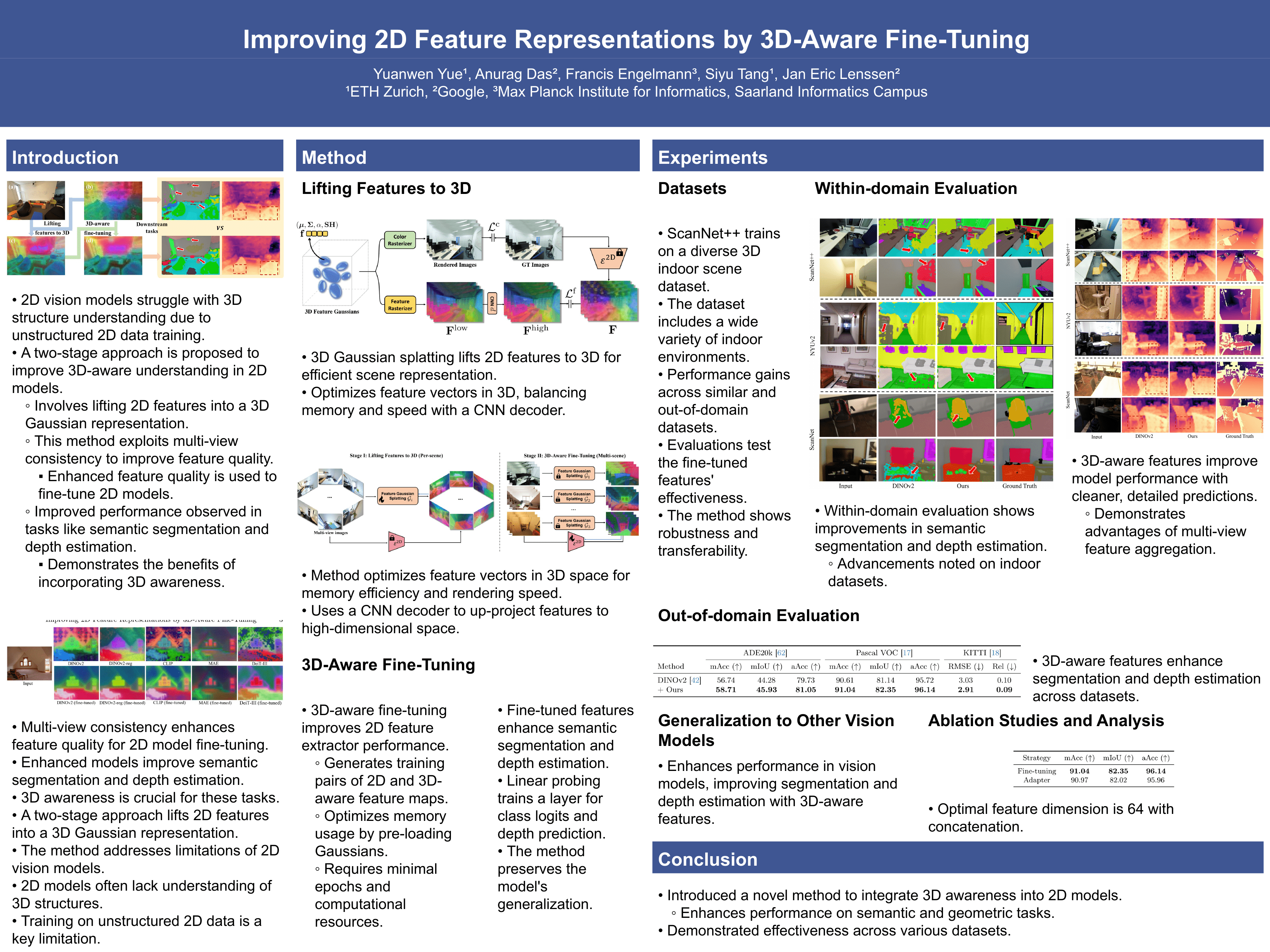}}
    \end{subfigure}
    \begin{subfigure}[t]{0.245\textwidth}
        \centering
        \fbox{\includegraphics[valign=c,width=\textwidth]{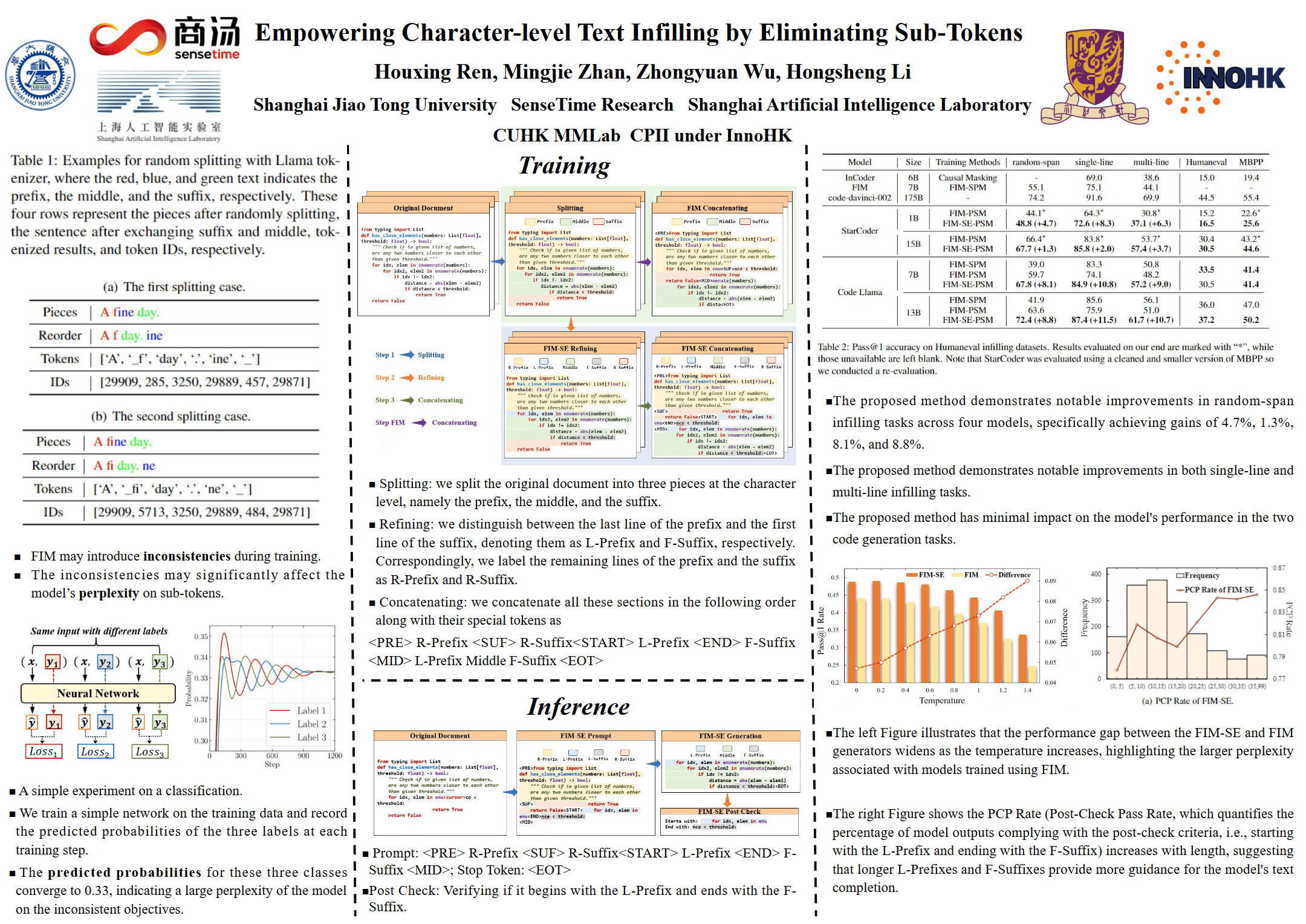}}
    \end{subfigure}
    \begin{subfigure}[t]{0.245\textwidth}
        \centering
        \fbox{\includegraphics[valign=c,width=\textwidth]{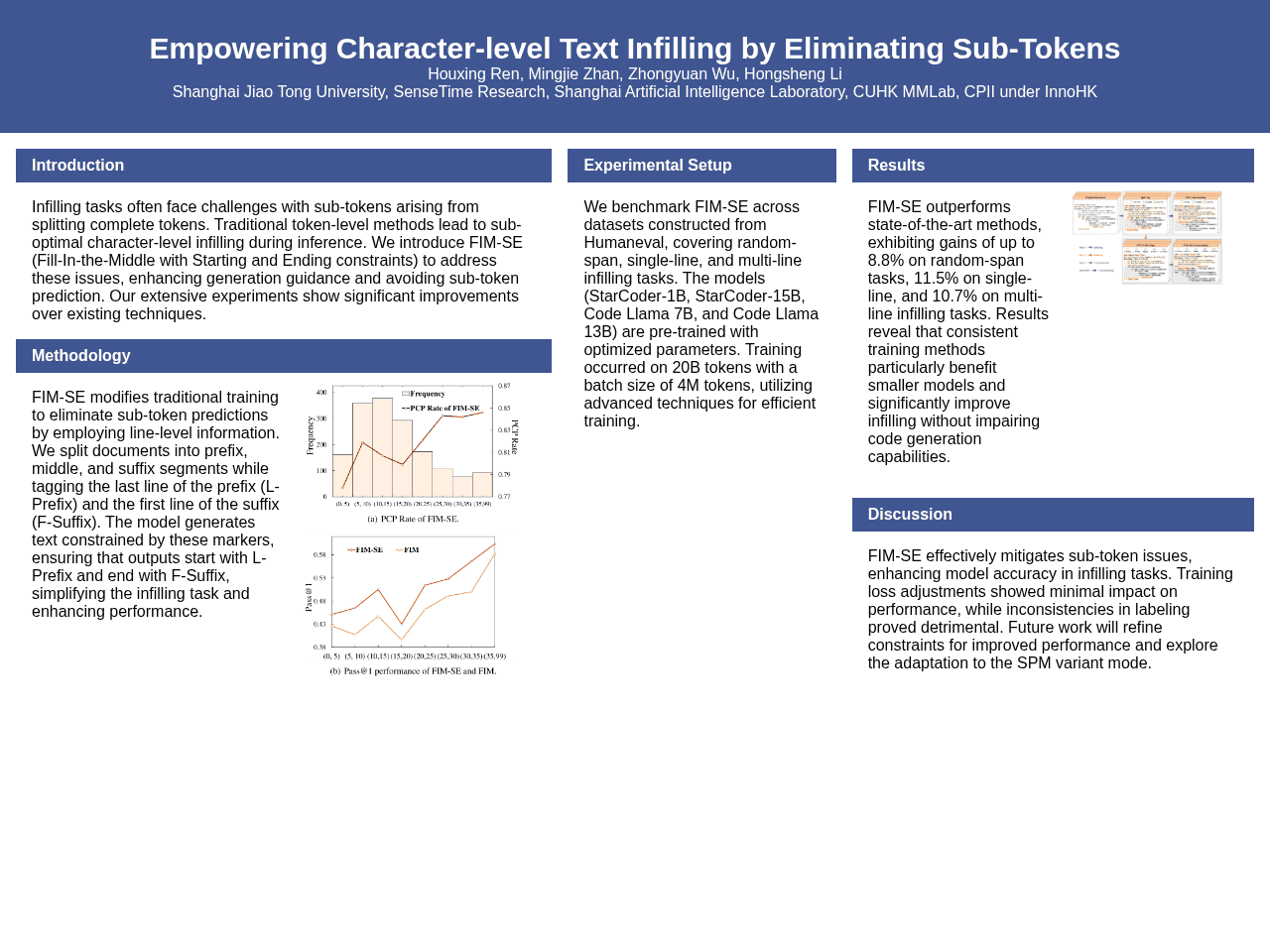}}
    \end{subfigure}
    \begin{subfigure}[t]{0.245\textwidth}
        \centering
        \fbox{\includegraphics[valign=c,width=\textwidth]{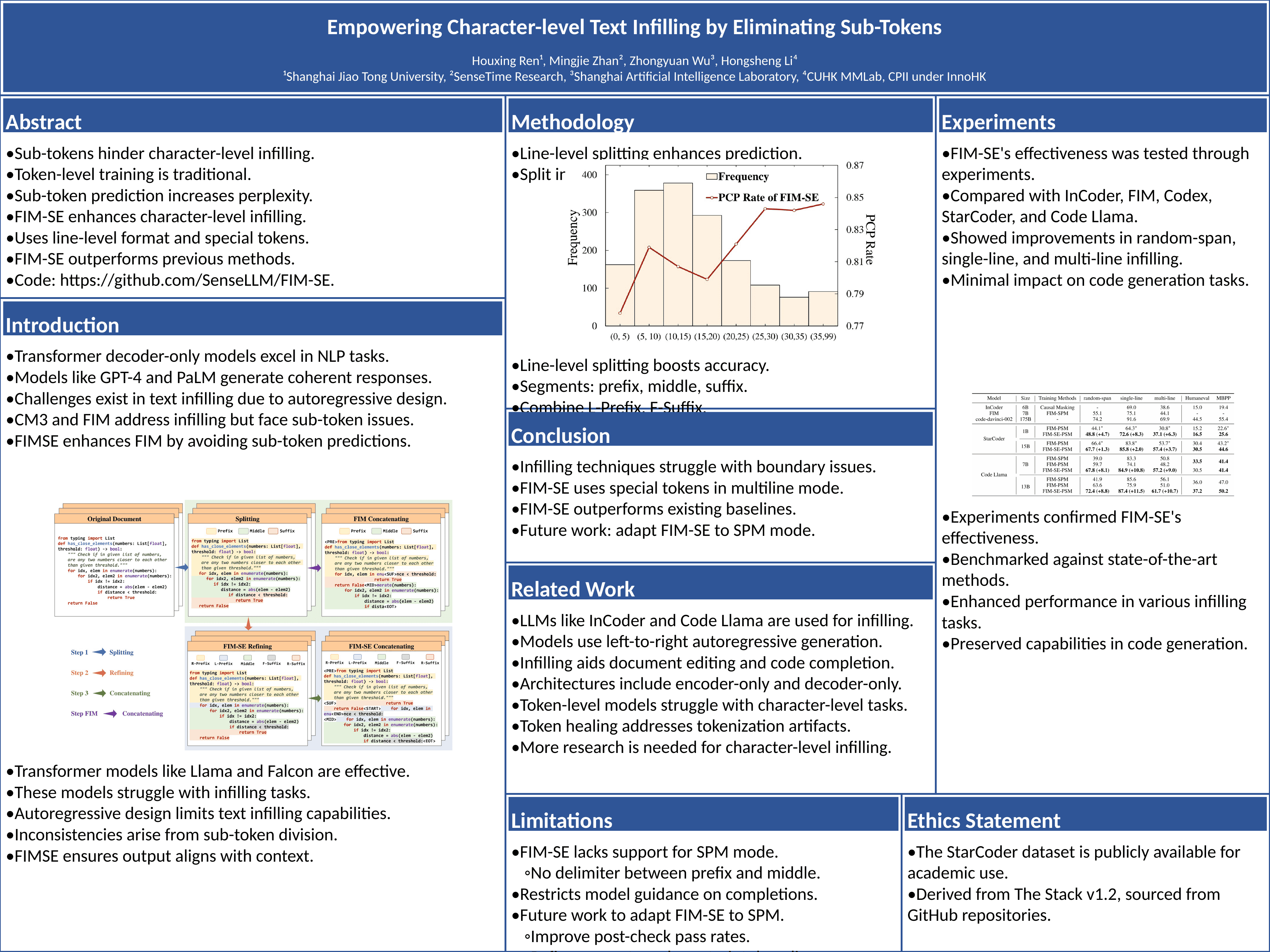}}
    \end{subfigure}
    \begin{subfigure}[t]{0.245\textwidth}
        \centering
        \fbox{\includegraphics[valign=c,width=\textwidth]{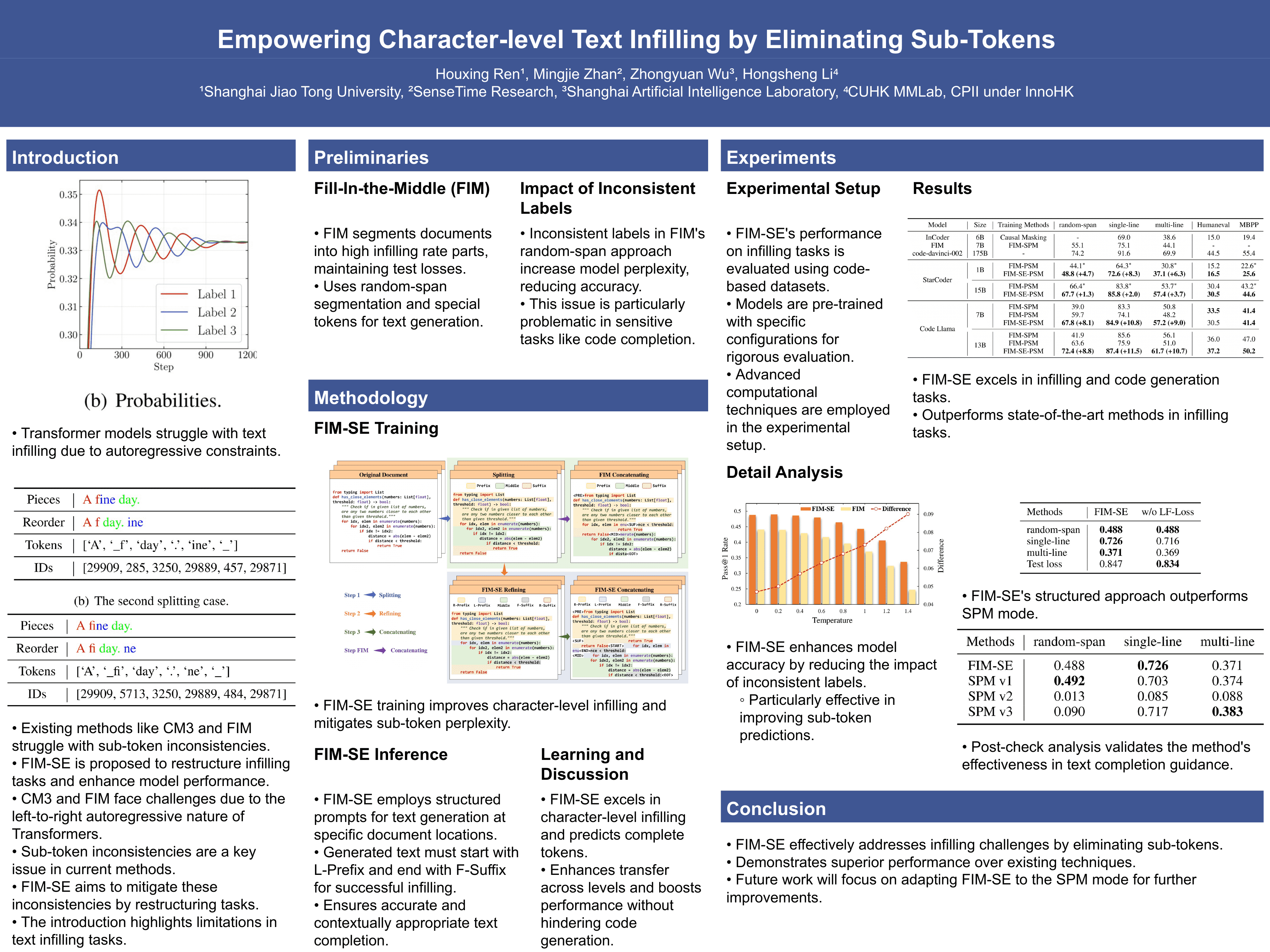}}
    \end{subfigure}
    \begin{subfigure}[t]{0.245\textwidth}
        \centering
        \fbox{\includegraphics[valign=c,width=\textwidth]{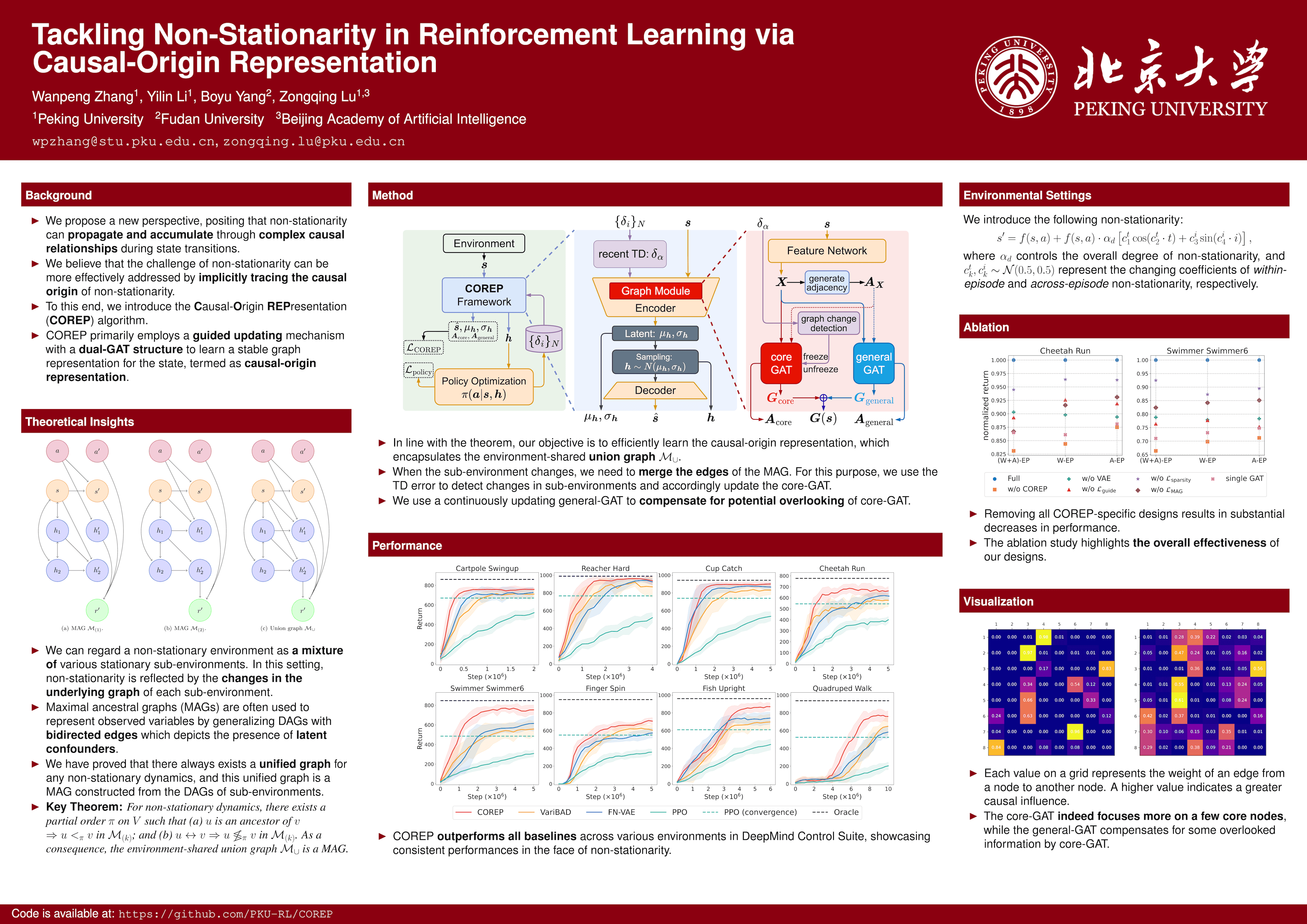}}
    \end{subfigure}
    \begin{subfigure}[t]{0.245\textwidth}
        \centering
        \fbox{\includegraphics[valign=c,width=\textwidth]{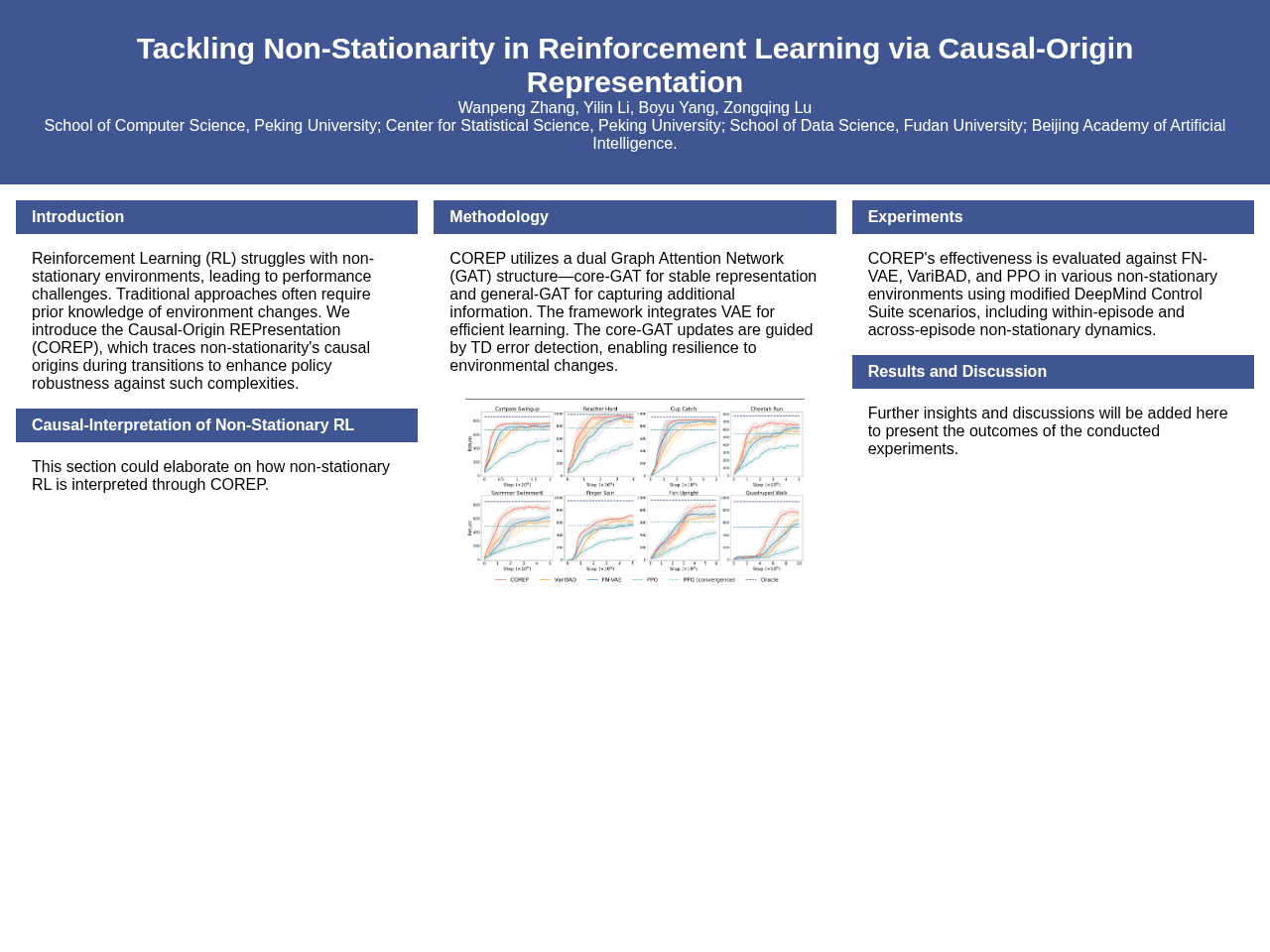}}
    \end{subfigure}
    \begin{subfigure}[t]{0.245\textwidth}
        \centering
        \fbox{\includegraphics[valign=c,width=\textwidth]{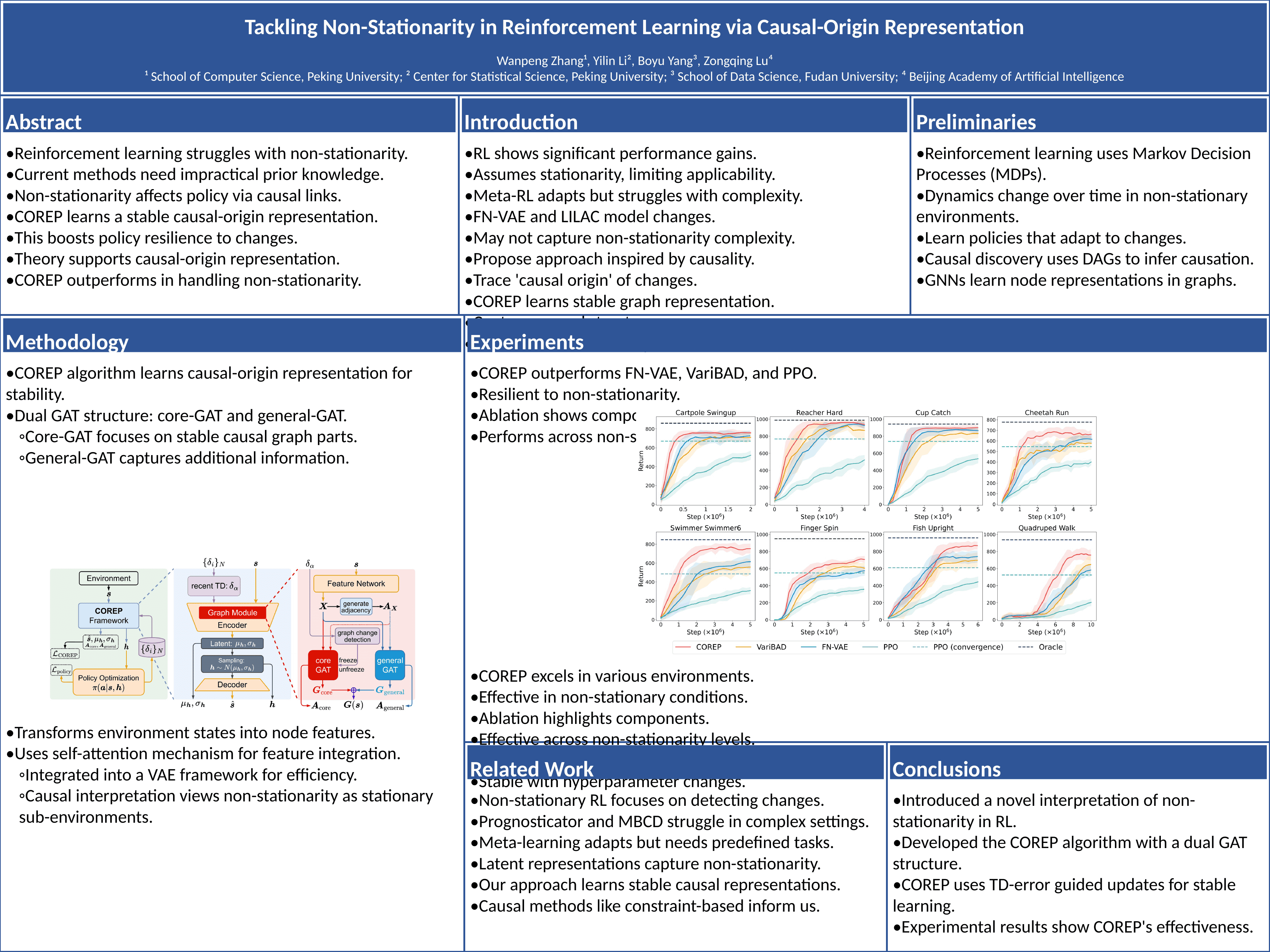}}
    \end{subfigure}
    \begin{subfigure}[t]{0.245\textwidth}
        \centering
        \fbox{\includegraphics[valign=c,width=\textwidth]{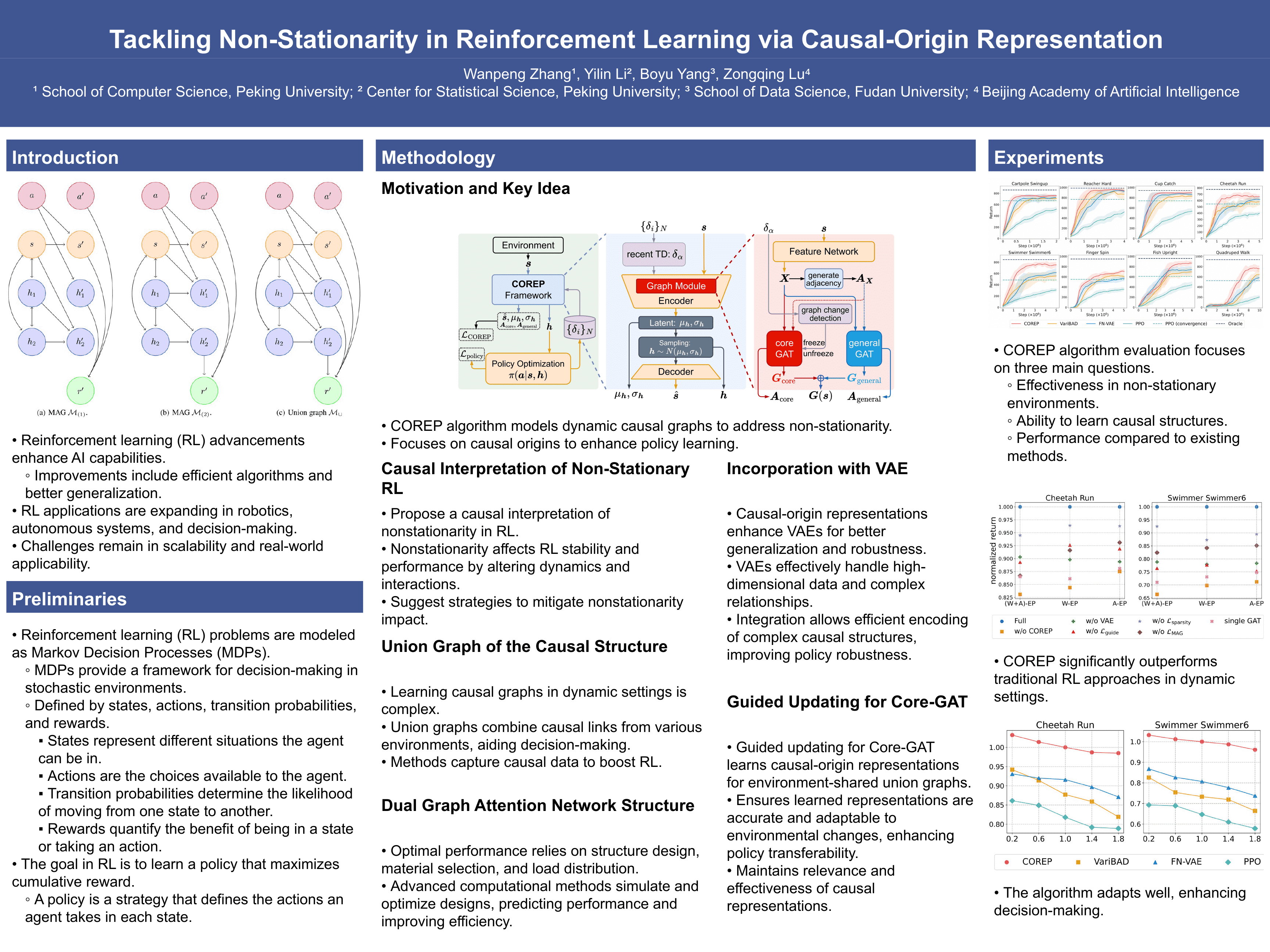}}
    \end{subfigure}
    \begin{subfigure}[t]{0.245\textwidth}
        \centering
        \fbox{\includegraphics[valign=c,width=\textwidth]{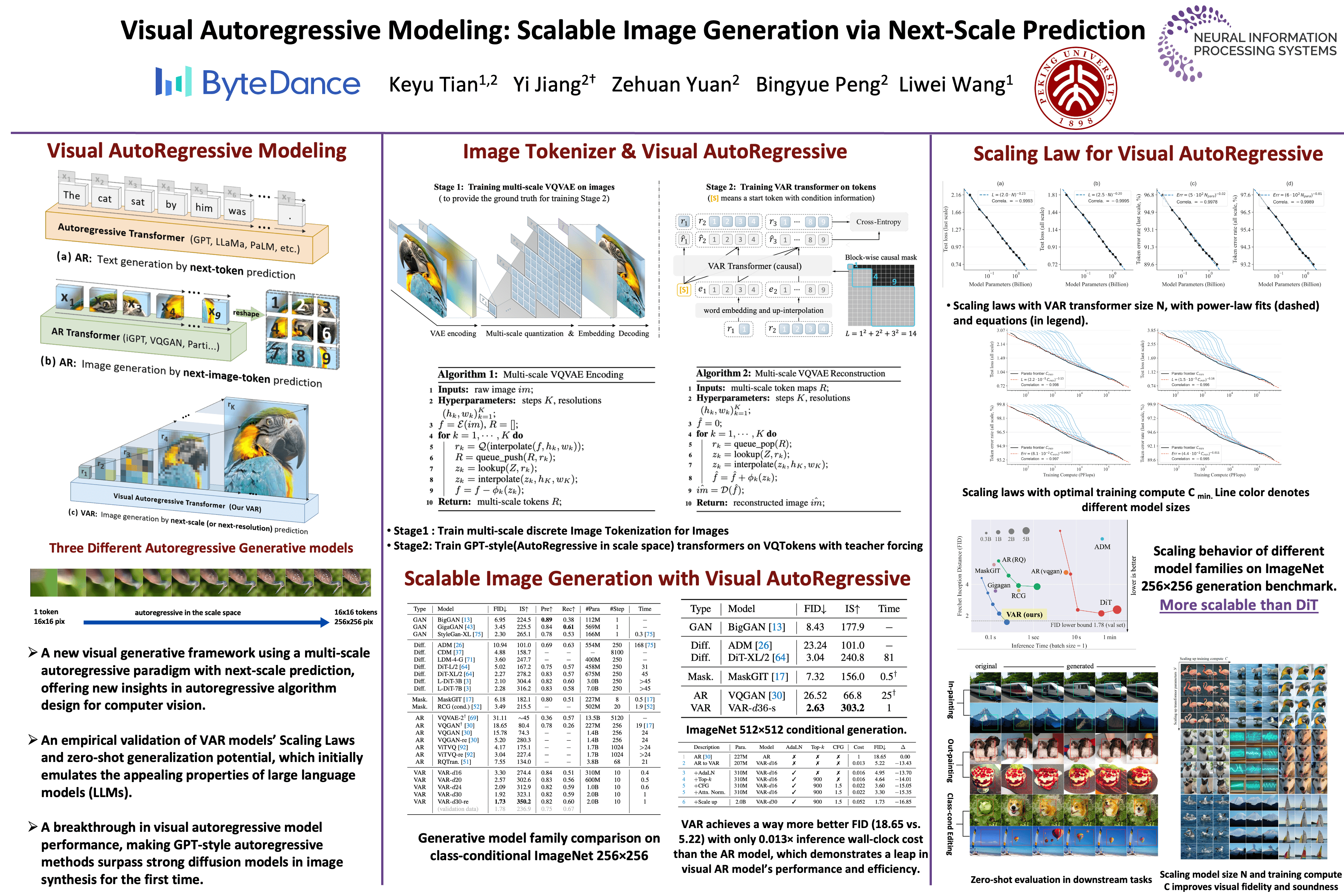}}
    \end{subfigure}
    \begin{subfigure}[t]{0.245\textwidth}
        \centering
        \fbox{\includegraphics[valign=c,width=\textwidth]{assets/04_qualitative_results/NeurIPS2024_VAR/poster_p2p.png}}
    \end{subfigure}
    \begin{subfigure}[t]{0.245\textwidth}
        \centering
        \fbox{\includegraphics[valign=c,width=\textwidth]{assets/04_qualitative_results/NeurIPS2024_VAR/poster_paper2poster.png}}
    \end{subfigure}
    \begin{subfigure}[t]{0.245\textwidth}
        \centering
        \fbox{\includegraphics[valign=c,width=\textwidth]{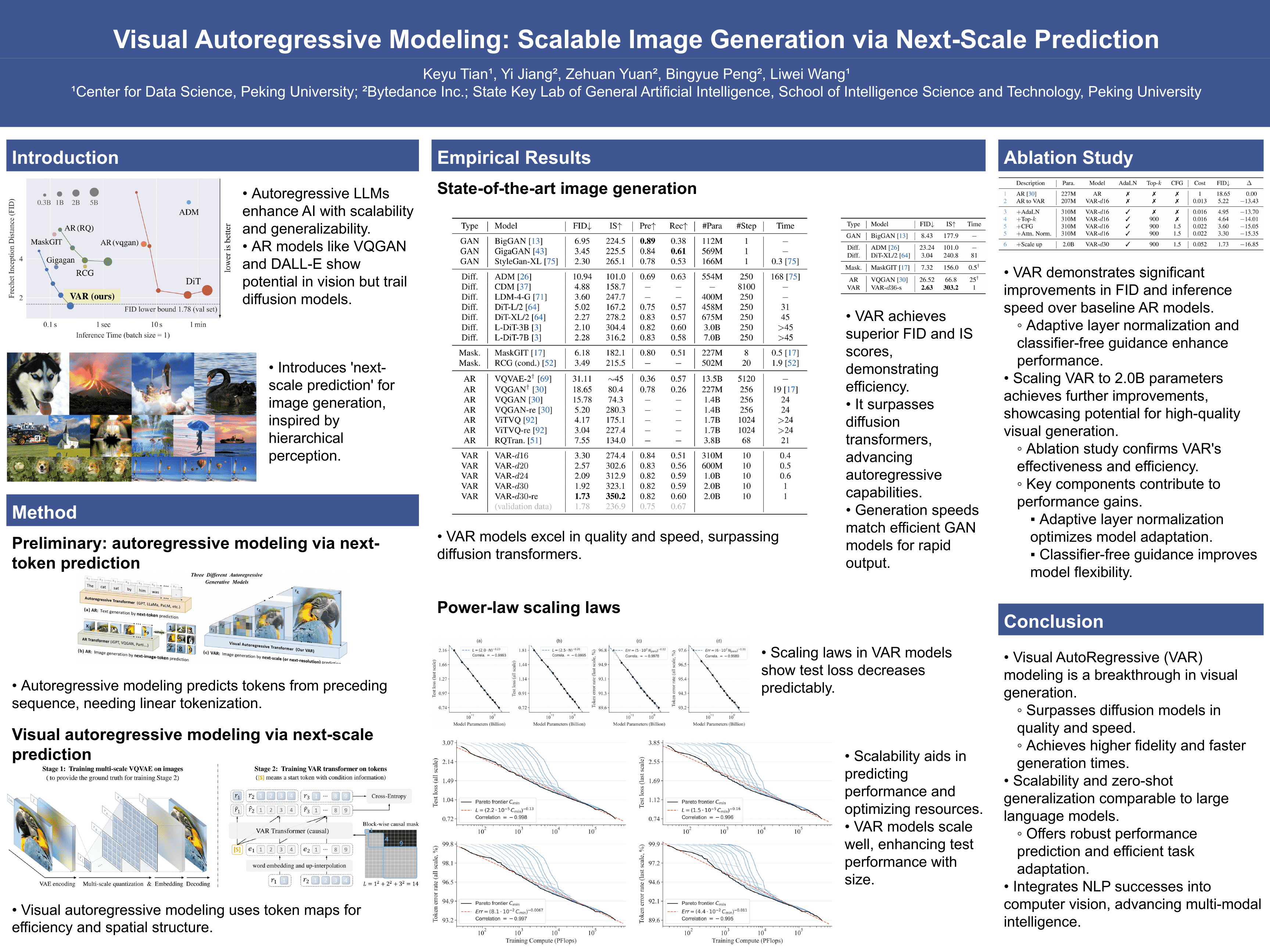}}
    \end{subfigure}
    
    \caption*{
        \makebox[0.245\textwidth]{(a) GT}\hfill
        \makebox[0.245\textwidth]{(b) P2P}\hfill
        \makebox[0.245\textwidth]{(c) Paper2Poster}\hfill
        \makebox[0.245\textwidth]{(d) PosterForest (Ours)}\hfill
    }
    \caption{
        \textbf{Qualitative Comparison.}
        Posters generated by the SoTA baseline methods and \textit{PosterForest}, based on papers from various AI fields (NLP, CV, RL), along with the original posters (GT) designed by the authors.
    }
    \label{fig:qualitative_comparison}
\end{figure*}

\section{Experiments}
\label{sec:Experiments}

\subsection{Experimental Setup}
\label{subsec:Experimental Setup}

\textbf{Baselines.}
Following the evaluation protocols of P2P~\cite{sun2025p2p} and Paper2Poster~\cite{pang2025paper2poster}, we compare four categories of baseline methods.
First, $\texttt{Oracle}$ methods represent upper bounds.
The original $\texttt{Paper}$ represents the upper bound for content fidelity, while the author-created $\texttt{GT~Poster}$ indicates the optimal layout and clarity achievable by human experts.
Second, end-to-end methods employ GPT-4o to generate posters directly.
Specifically, $\texttt{GPT-4o-HTML}$ renders posters by converting the paper into HTML, whereas $\texttt{GPT-4o-Image}$ produces poster images in a single step using GPT-4o.
Third, multi-agent workflows encompass general-purpose converters and algorithmic generators.
For this category, we evaluate the PDF-to-HTML conversion toolkit of $\texttt{OWL}$~\cite{hu2025owl} and Python-pptx conversion results of $\texttt{PPTAgent}$.
Finally, poster-specialized agents include $\texttt{P2P}$~\cite{sun2025p2p}, Paper2Poster, and our proposed method.
To ensure that visual factors did not influence qualitative evaluations and user studies, we standardized the color scheme and font across all posters.

\noindent\textbf{Datasets.}
For quantitative evaluation, we used the 100 paper–poster pairs provided by the Paper2Poster benchmark~\cite{pang2025paper2poster}, which is an extension of the PosterSum dataset~\cite{saxena2025postersum}.
For qualitative results and user studies, we additionally collected 15 recent paper–poster pairs from the AI conferences (e.g., NeurIPS, CVPR, and ACL).
More comprehensive experimental details are provided in the Appendix.

\noindent\textbf{Implementation Details.}
For further details regarding model architecture and evaluation protocols, please refer to the Appendix.


\subsection{Qualitative Evaluation}
\label{subsec:Qualitative Evaluation}

In~\Cref{fig:qualitative_comparison}, we compare our method with state-of-the-art baselines, P2P and Paper2Poster, and the poster made by the authors (Ground Truth).
All experiments were conducted using the GPT-4o framework to ensure consistency in model performance.
Our approach dynamically adjusts column widths and panel sizes, resulting in a balanced distribution of content.
Compared to P2P and Paper2Poster, our method makes more efficient use of space, prevents the inclusion of oversized or undersized figures, and avoids excessively long or verbose paragraphs through strategic hierarchical organization.
Notably, our method excels at preserving information: for the VAR paper (4th row), both P2P and Paper2Poster omit either the result table or graph, whereas our method retains both, ensuring that critical information is maintained.
Furthermore, our approach demonstrates robust performance across diverse paper formats and academic domains, as illustrated by the examples from 3D Vision-ECCV (1st row), Language Processing-ACL (2nd row), and Reinforcement Learning-ICML (3rd row).

\subsection{Quantitative Evaluation}

Following the evaluation protocol introduced in P2P and Paper2Poster, we employ MLLM-as-a-Judge metrics for quantitative evaluation.
The GPT-4o model is prompted to act as six independent judges, each assigning a score from 1 to 5 based on the following criteria: element quality, layout balance, engagement, clarity, content completeness, and logical flow.
The first three criteria assess aesthetics, while the latter three evaluate the informativeness of the generated poster.
As shown in~\Cref{tab:judge_table}, the judges indicate that our method is comparable to other baselines in terms of aesthetics, and demonstrates superior performance in informativeness.
Importantly, these scores are the closest to those of the author-created ground truth posters (GT), demonstrating the effectiveness of our approach.
Details of the MLLM-as-a-Judge are provided in the supplementary material.

While MLLM-based evaluation provides a scalable and objective means for poster assessment, it still has inherent limitations in fully capturing subjective preferences and subtle qualities valued by human readers.
Therefore, to complement the quantitative results, we further conduct a user study to obtain human judgments and validate the practical effectiveness of our method.


\begin{table}[t]
    \centering
    \resizebox{\linewidth}{!}
    {
        \begin{tabular}{l c cccc cccc c}
        \toprule
        \multirow{2}{*}{\textbf{Model}} &  \multirow{2}{*}{\textbf{TF}} & \multicolumn{4}{c}{\textbf{Aesthetic Score ↑}} & \multicolumn{4}{c}{\textbf{Information Score ↑}} & \multirow{2}{*}{\textbf{Overall ↑}} \\ 
                                        & & Elem.  & Lay.  & Eng.  & \textbf{Avg.}  & Cla.   & Cont.   & Logic   & \textbf{Avg.}  &                                     \\
        \midrule
        Paper                           & - & 4.05     & 3.89    & 2.80     & 3.58           & 4.00      & 4.68      & 3.98    & 4.22           & 3.90                                \\
        GT Poster                       & - & 4.07     & 3.90    & 2.70     & 3.56           & 4.09      & 3.96      & 3.89    & 3.98           & 3.77                                \\
        \midrule
        4o-HTML                         & \yesmark & 3.53     & 3.82    & 2.72     & 3.36           & 3.94      & 3.64      & 3.47    & 3.68           & 3.52                                \\
        4o-Image                        & \yesmark & 2.93     & 3.02    & 2.75     & 2.90           & 1.05      & 2.04      & 2.22    & 1.77           & 2.33                                \\
        OWL-4o                          & \yesmark & 2.76     & 3.62    & 2.56     & 2.98           & 3.92      & 2.89      & 3.36    & 3.39           & 3.19                                \\
        PPTAgent-4o                     & \yesmark & 2.49     & 3.05    & 2.45     & 2.66           & 2.05      & 1.26      & 1.38    & 1.56           & 2.11                                \\
        P2P-4o                          & \nomark & 3.63     & \textbf{4.01}    & \underline{2.96}     & \textbf{3.91}           & 3.80      & \textbf{3.99}      & 3.48    & \textbf{3.94}           & \underline{3.72}                                \\
        PosterAgent-Qwen                & \nomark & 3.93     & 3.67    & 2.89     & 3.50           & 3.95      & 3.85      & \underline{3.68}    & 3.83           & 3.66                                \\
        PosterAgent-4o                  & \nomark & 3.95     & 3.86    & 2.93     & 3.58           & \textbf{4.03}      & 3.96      & 3.60    & 3.86           & \underline{3.72}                                \\
        PosterForest-Qwen (Ours)                  & \yesmark & \textbf{4.02}     & 3.85    & \textbf{2.99}     & 3.62           & 3.98      & 3.93      & 3.54    & 3.82           & \underline{3.72}                                \\
        PosterForest-4o (Ours)                  & \yesmark & \textbf{4.02}     & \underline{3.96}    & \underline{2.96}     & \underline{3.65}           & \underline{4.00}      & \underline{3.88}      & \textbf{3.71}    & \underline{3.87}           & \textbf{3.76}                                \\

        \bottomrule
        \end{tabular}
    
    }
    \caption{
        \textbf{MLLM-as-a-Judge score across four categories of baselines.}
        The average score serves as a fine-grained assessment of 6 different perspectives.
        The best score is \textbf{bold}, and the second is \underline{underlined} for each criterion.
        “TF” denotes \textit{Training-free} methods.
    }
    \label{tab:judge_table}
    \vspace{-1mm}
\end{table}

\begin{table}[ht]
    \centering
    \resizebox{0.9\linewidth}{!}
    {
        \begin{tabular}{lrrrr}
        \toprule
        \textbf{Method} & Content & Esthetics & Structure & Overall \\
        \midrule
        4o-HTML             &  2.0 \%    &  1.6 \%      &  2.4 \%      &  1.6 \%             \\
        P2P             & 9.2 \%    & 21.2 \%      & 13.2 \%      & 12.0 \%             \\
        Paper2Poster    & 32.8 \%    & 24.0 \%      & 24.8 \%      & 27.2 \%             \\
        \textbf{Ours}   & \textbf{56.0 \%}    & \textbf{53.2 \%}      & \textbf{59.6 \%}      & \textbf{59.2 \%}             \\
        \bottomrule
        \end{tabular}
    }
    \caption{
        \textbf{Human Evaluation.}
        Numbers represent the proportion of times each method was ranked first for each criterion.
    }
    \label{tab:user_study}
    \vspace{-1em}
\end{table}


\captionsetup[subfigure]{labelformat=empty}
\begin{figure*}[ht]
    \centering
    \begin{minipage}{0.245\linewidth}
        \centering
        \begin{tikzpicture}[zoomboxarray, figurename=base, zoomboxes below, zoomboxarray columns=1, zoomboxarray rows=1, caption margin=-1mm, zoomboxarray inner gap=0mm, connect zoomboxes] 
            \node [image node] { \fbox{\includegraphics[width=1.0\linewidth]{assets/04_ablation_results/poster_base_new.jpg}}};
            \zoombox[color code=orange,magnification=2, width=1.0\linewidth, height=0.50\linewidth]{0.625,0.685}
        \end{tikzpicture}
        \vspace{-20mm}
        \caption*{(a) Base}
    \end{minipage}
    \hfill
    \begin{minipage}{0.245\linewidth}
        \centering
        \begin{tikzpicture}[zoomboxarray, figurename=content, zoomboxes below, zoomboxarray columns=1, zoomboxarray rows=1, caption margin=-1mm, zoomboxarray inner gap=0mm] 
            \node [image node] { \fbox{\includegraphics[width=1.0\linewidth]{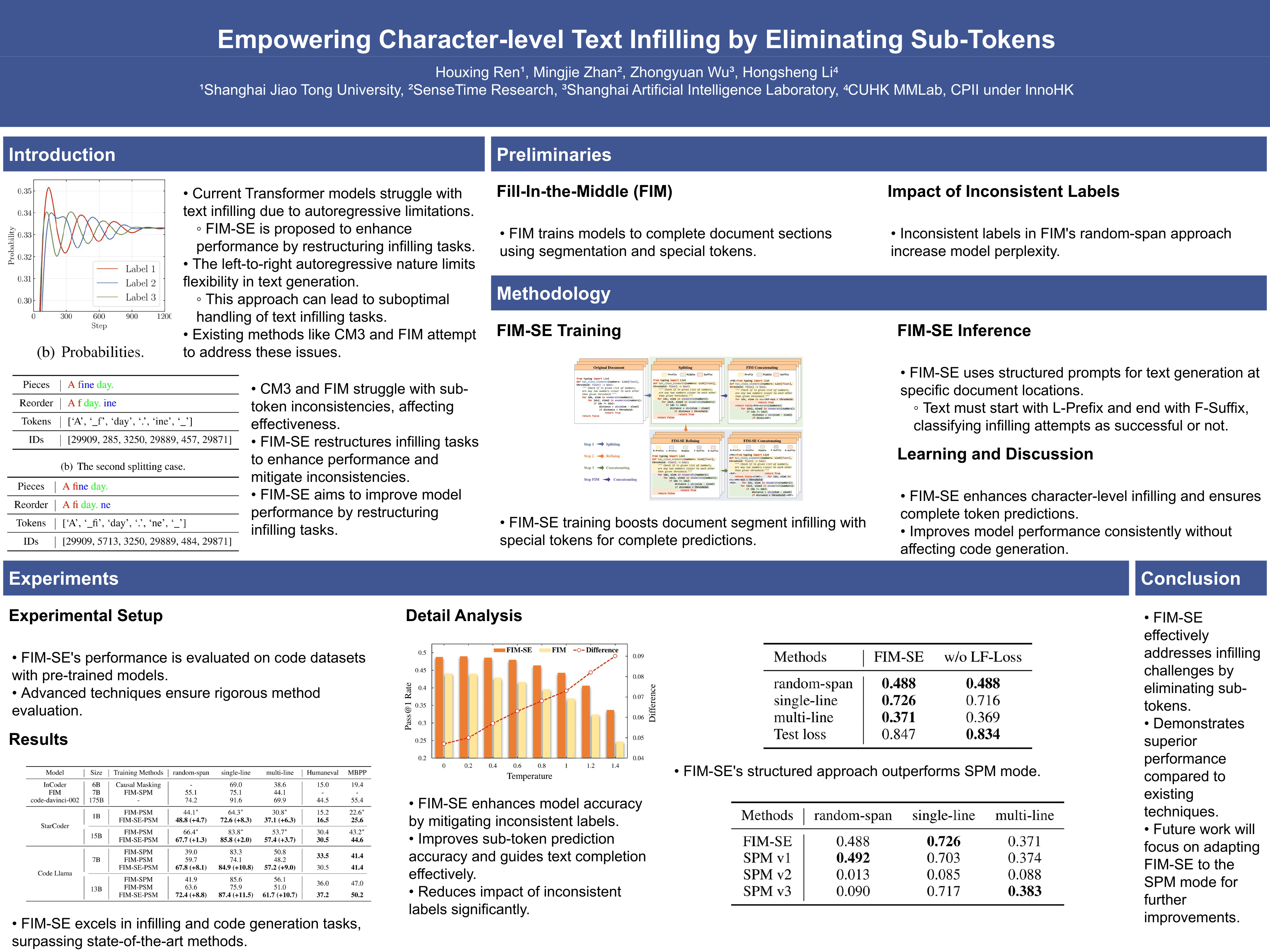}} };
            \zoombox[color code=orange,magnification=2, width=1.0\linewidth, height=0.50\linewidth]{0.625,0.685}
        \end{tikzpicture}
        \vspace{-20mm}
        \caption*{(b) Only Content Agent}
    \end{minipage}
    \hfill
    \begin{minipage}{0.245\linewidth}
        \centering
        \begin{tikzpicture}[zoomboxarray, figurename=layout, zoomboxes below, zoomboxarray columns=2, zoomboxarray rows=1, caption margin=-1mm, zoomboxarray inner gap=-3mm] %
            \node [image node] { \fbox{\includegraphics[width=1.0\linewidth]{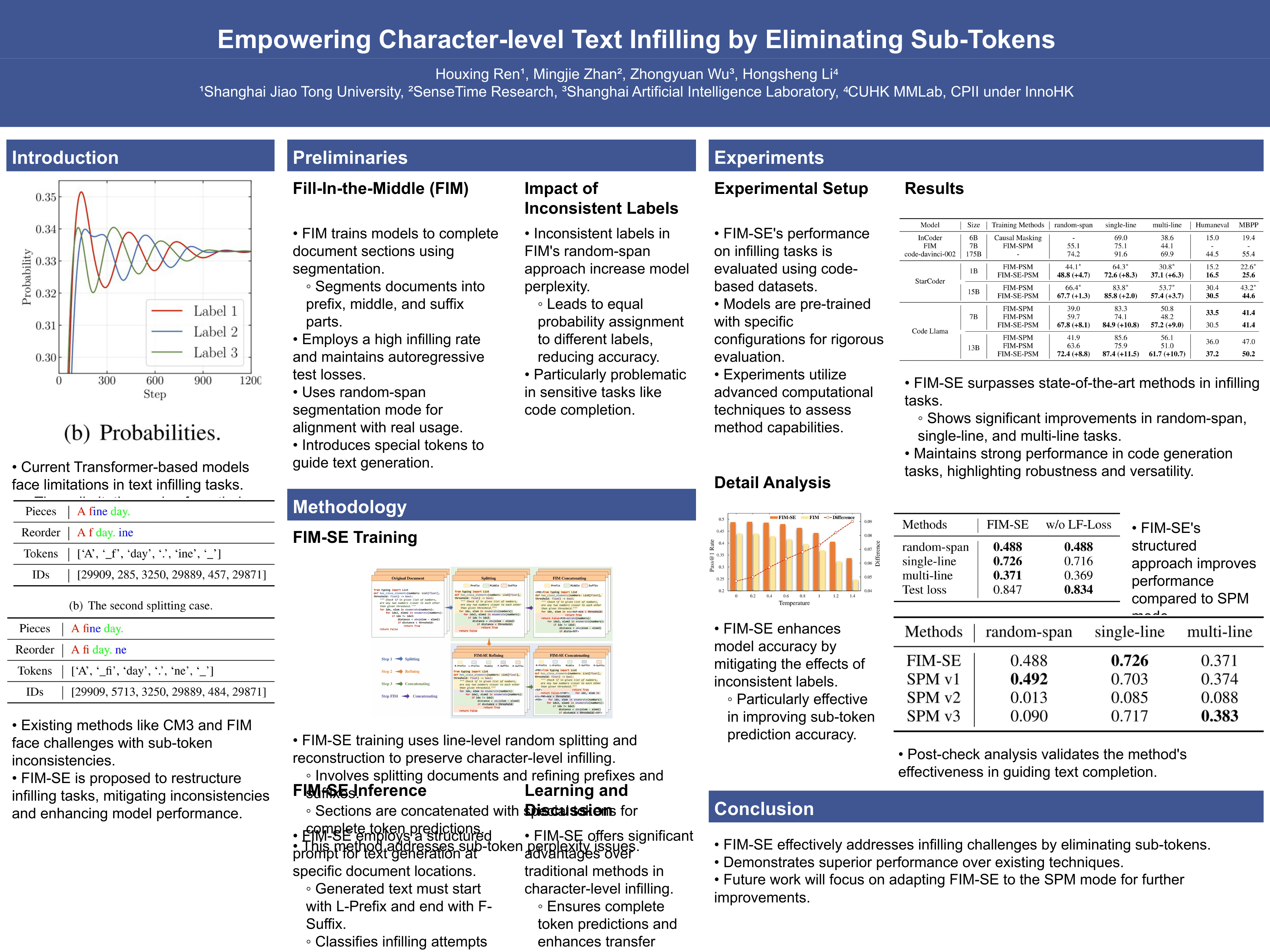}} };
            \zoombox[color code=orange,magnification=1.75, width=0.45\linewidth, height=0.50\linewidth]{0.35,0.45}
            \zoombox[color code=magenta,magnification=2.3, width=0.53\linewidth, height=0.45\linewidth]{0.34,0.13}
        \end{tikzpicture}
        \vspace{-20mm}
        \caption*{(c) Only Layout Agent}
    \end{minipage}
    \hfill
    \begin{minipage}{0.245\linewidth}
        \centering
        \begin{tikzpicture}[zoomboxarray, figurename=both, zoomboxes below, zoomboxarray columns=2, zoomboxarray rows=1, caption margin=-1mm, zoomboxarray inner gap=-3mm] 
            \node [image node] { \fbox{\includegraphics[width=1.0\linewidth]{assets/04_qualitative_results/ACL2024_FIM/poster_ours_new.png}} };
            \zoombox[color code=orange,magnification=1.75, width=0.45\linewidth, height=0.50\linewidth]{0.365,0.45}
                \zoombox[color code=magenta,magnification=2.3, width=0.53\linewidth, height=0.45\linewidth]{0.355,0.13}
        \end{tikzpicture}
        \vspace{-20mm}
        \caption*{(d) Both Agents}
    \end{minipage}
    \caption{
    \textbf{Effect of Content and Layout Agents.}
    Using both agents balances layout (\textcolor{orange}{orange}) and removes redundancy (\textcolor{magenta}{magenta}), yielding well-organized posters with proper information density and strong visual harmony.}
    \label{fig:ablation_study_2_agents}
\end{figure*}

\captionsetup[subfigure]{labelformat=empty}
\begin{figure}[ht]
    \centering
    \begin{minipage}{0.48\linewidth}
        \centering
        \begin{tikzpicture}[zoomboxarray, figurename=without_hierarchical, zoomboxes below, zoomboxarray columns=2, zoomboxarray rows=1, caption margin=-1mm, zoomboxarray inner gap=-6.3mm] 
            \node [image node] { \fbox{\includegraphics[width=1.0\linewidth]{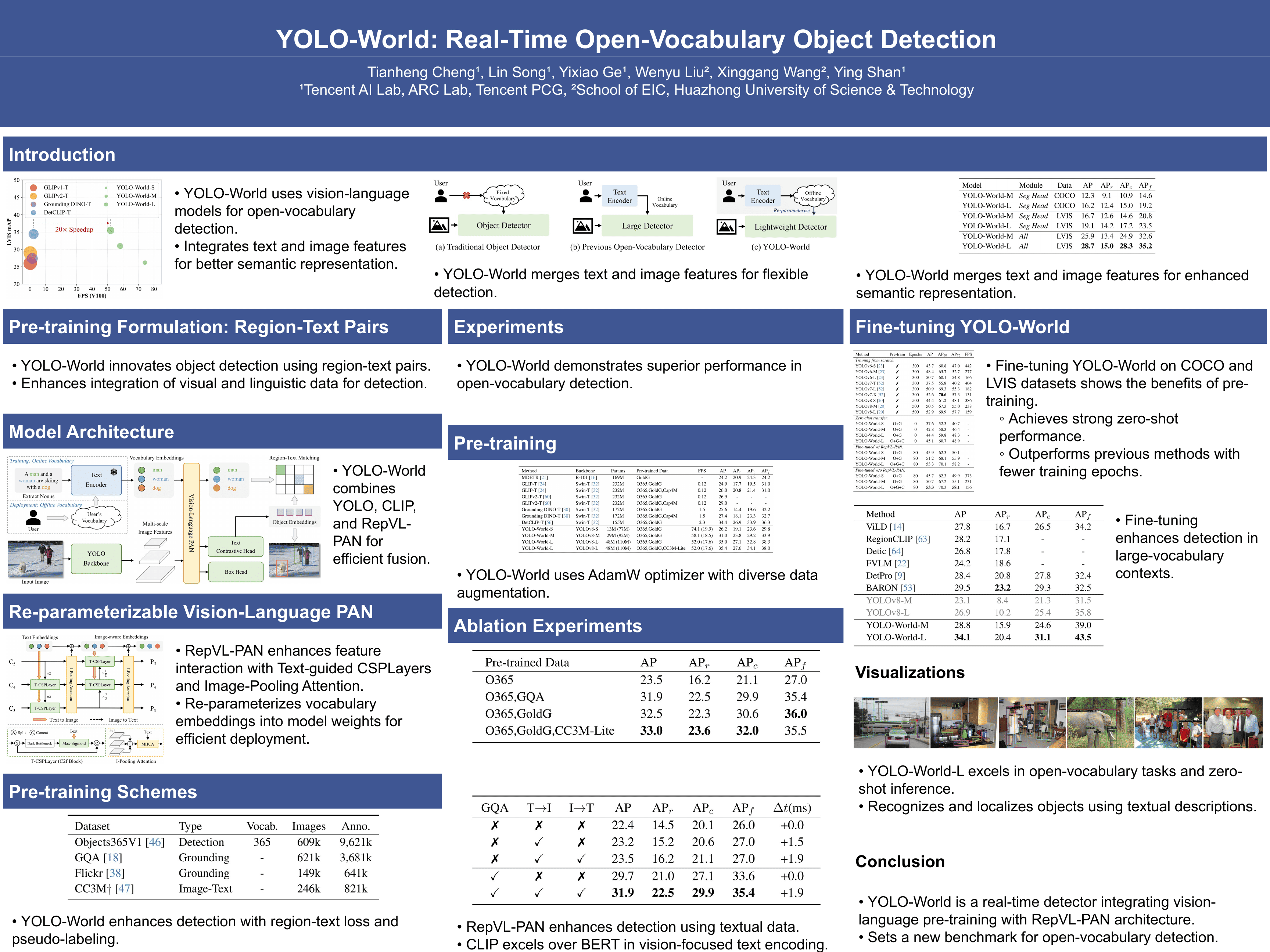}}};
            \zoombox[color code=orange,magnification=3, width=0.40\linewidth, height=0.47\linewidth]{0.415,0.57}
            \zoombox[color code=magenta,magnification=3.2, width=0.58\linewidth, height=0.47\linewidth]{0.83,0.76}
        \end{tikzpicture}
        \vspace{-18mm}
    \end{minipage}
    \hfill
    \begin{minipage}{0.48\linewidth}
        \centering
        \begin{tikzpicture}[zoomboxarray, figurename=with_hierarchical, zoomboxes below, zoomboxarray columns=2, zoomboxarray rows=1, caption margin=-1mm, zoomboxarray inner gap=-6.3mm] 
            \node [image node] { \fbox{\includegraphics[width=1.0\linewidth]{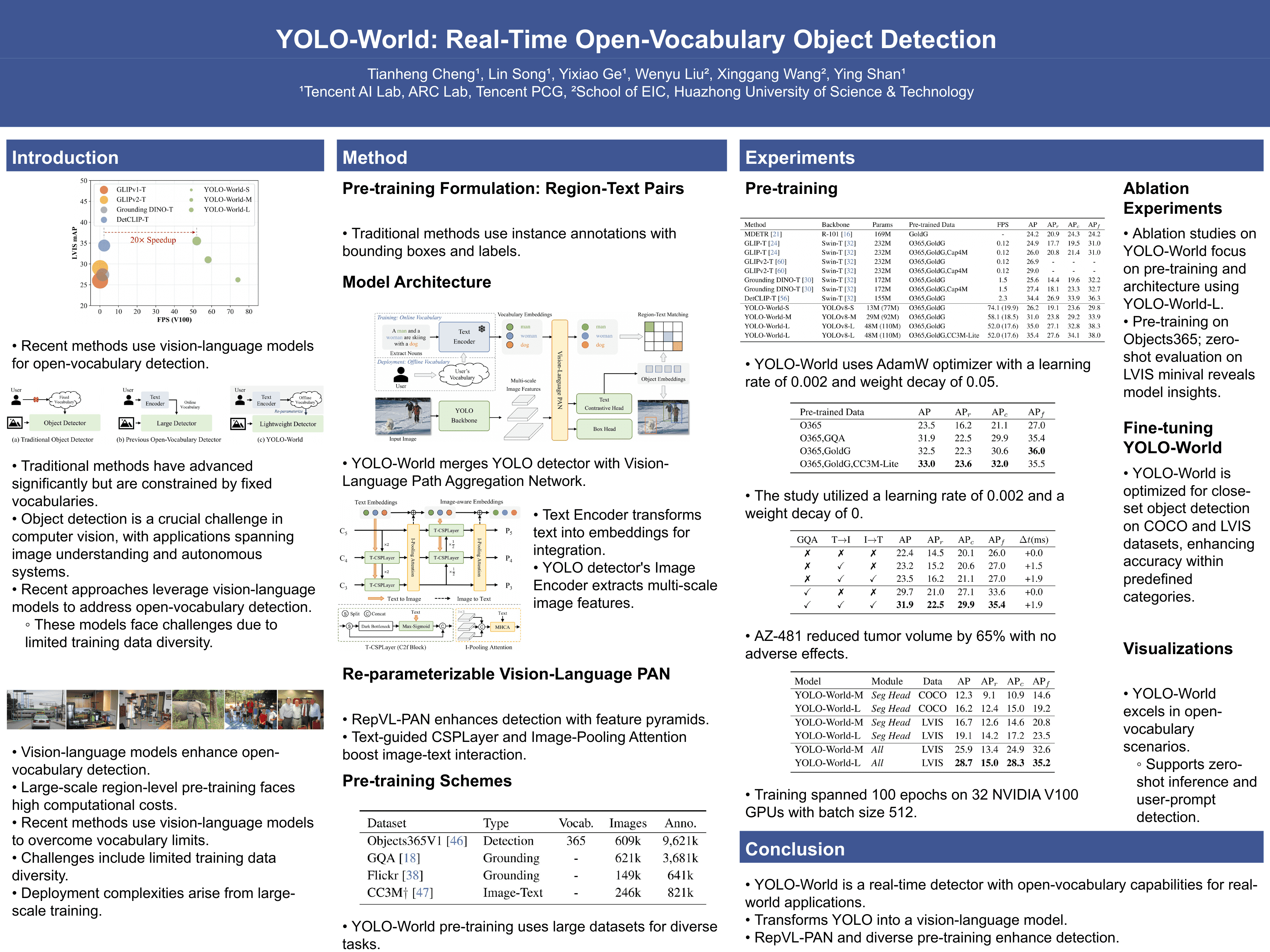}} };
            \zoombox[color code=orange,magnification=3, width=0.40\linewidth, height=0.47\linewidth]{0.64,0.75}
            \zoombox[color code=magenta,magnification=2.5, width=0.58\linewidth, height=0.47\linewidth]{0.72,0.24}
        \end{tikzpicture}
        \vspace{-18mm}
    \end{minipage}
    \vspace{2mm}
    \begin{minipage}{0.48\linewidth}
        \lstset{
            basicstyle=\scriptsize\ttfamily,
            breaklines=true
        }
        \begin{lstlisting}
Root
 |-- Section 1
 |-- Subsection 1.1
 |-- Section 2
 |-- Subsection 2.1
 |-- Subsection 2.2
        \end{lstlisting}
    \end{minipage}
    \hfill
    \begin{minipage}{0.48\linewidth}
        \lstset{
            basicstyle=\scriptsize\ttfamily,
            breaklines=true
        }
        \begin{lstlisting}
Root
 |-- Section 1
 |   `-- Subsection 1.1
 `-- Section 2
 |   |-- Subsection 2.1
 |   `-- Subsection 2.2
        \end{lstlisting}
    \end{minipage}
    \vspace{-2mm}
    \caption*{
        \makebox[0.48\linewidth]{
            (a) w/o Hierarchical
        }\hfill
        \makebox[0.48\linewidth]{
            (b) w/ Hierarchical 
        }\hfill
    }
    \caption{
        \textbf{
            Effect of Hierarchical Content Tree. 
        }
        With a hierarchical structure, logical order and spatial coherence are preserved (\textcolor{orange}{orange}) and text–visual alignment improves; the performance table appears under Experiments rather than Introduction (\textcolor{magenta}{magenta}). 
    }
    \label{fig:ablation_study_1}
    \vspace{-1em}
\end{figure}

\subsection{User Study}
\label{subsec:User Study}

To conduct a user study to evaluate poster quality from a human perspective, we recruited 25 participants, all of whom were graduate students in the field of AI and had participated in scientific conferences.  
The study uses 10 sets (40 questions in total), each consisting of a group of posters and four evaluation questions.
Each poster group is generated with four GPT-4o-based methods: 4o-HTML, P2P, Paper2Poster, and our proposed method.
For each set, participants are asked to select one poster per question based on the following criteria: (1) \textit{content fidelity}, which poster best reflects the content of the paper; (2) \textit{aesthetic quality}, which poster is the most visually harmonious; (3) \textit{structural clarity}, which poster delivers information in the most structurally effective way; and (4) \textit{overall quality}, which poster appears most complete and well-polished overall.
As shown in~\Cref{tab:user_study}, our proposed method is strongly preferred over the other SoTA baselines across all four criteria.
Please refer to the Appendix for further details on the user study.

\subsection{Ablation Study}
\label{subsec:Ablation Study}

\subsubsection{Effect of Hierarchical Content Tree}

We conducted an ablation study to analyze the impact of incorporating hierarchical structure into the Content Tree, $\mathcal{T}_\texttt{content}$, during content generation.
When the hierarchical organization is omitted, as shown in~\Cref{fig:ablation_study_1} (a), sections and subsections are often disordered or mixed together on the poster.
This results in a loss of semantic grouping and spatial coherence between related elements such as figures and explanatory text.
In contrast, applying hierarchical parsing, as in~\Cref{fig:ablation_study_1} (b), preserves the logical relationships between sections and subsections, ensuring that related content is grouped and displayed in a consistent and interpretable manner.
This hierarchical structure enhances both the readability and spatial cohesion of the generated poster, supporting more effective information delivery.

\subsubsection{Effect of Content and Layout Agents}

To evaluate the effectiveness of the multi-agent collaboration, we conducted an ablation study by comparing four configurations: (a) the base model with only the initial Poster Tree, (b) Content Agent only, (c) Layout Agent only, and (d) both agents combined.
The Content Agent prunes cross-node redundancy and right-sizes the remaining text to the layout, as shown in~\Cref{fig:ablation_study_2_agents}(b), but often leads to suboptimal panel arrangements and unbalanced layouts.
In contrast, the Layout Agent focuses on optimizing the spatial arrangement at the layout level.
As illustrated in~\Cref{fig:ablation_study_2_agents} (c), this configuration achieves improved visual organization, but suboptimal figure scaling and text overflow frequently occur due to the lack of content adjustment.
When both agents are used together, as shown in~\Cref{fig:ablation_study_2_agents} (d), the system effectively addresses both content redundancy and layout imbalance, producing well-organized posters with appropriate information density and visual harmony.
These results confirm that the joint use of Content and Layout Agents is essential for achieving both semantic and structural quality in automated poster generation.

\section{Conclusion}
\label{sec:Conclusion}

This work proposed PosterForest, a hierarchical multi-agent framework for scientific poster generation that explicitly models the interplay between document structure and layout design.
The proposed Poster Tree serves as a unified intermediate representation, enabling integrated reasoning over semantic and spatial attributes.
Through hierarchical refinement driven by Content and Layout Agents, our method dynamically balances information density and visual harmony without any training or dataset-specific tuning.
Empirical results and user studies confirm that PosterForest substantially outperforms prior methods in informativeness, clarity, and structural quality.

\section*{Limitations}
\label{sec:Limitations}
While PosterForest demonstrates significant improvements, certain limitations persist. First, generated posters may not always achieve optimal content density, which can lead to less efficient space utilization.
Second, the lack of robust quality metrics may limit the comprehensiveness of quantitative evaluation, highlighting the need for further development of advanced evaluation methodologies in future research. 
Detailed failure cases and future directions are provided in the supplementary material.

\section*{Ethical Considerations}
\label{sec:ethical}
We rely only on officially released, publicly accessible models and APIs. In all experiments, we call GPT-4o through OpenAI’s official interface, and we also use publicly available Qwen checkpoints where noted. We do not fine-tune any models. Our framework is training-free and used strictly within the terms of the providers’ licenses. Source papers and posters are analyzed solely for non-commercial research under practices consistent with academic fair use, and we include references to the original sources to respect creator rights. Our human evaluation recruits 25 graduate-level participants with prior conference experience; assessments concern posters only, and no personal attributes are collected or analyzed. An AI assistant was used for sentence-level drafting and refining to improve clarity.

\section*{Acknowledgments}


This work was supported by Samsung Research, Samsung Electronics Co., Ltd.; the Basic Science Research Program through the National Research Foundation of Korea (NRF) funded by the Korea government (MSIT) (No. RS-2025-00520207); Institute of Information \& Communications Technology Planning \& Evaluation (IITP) grants funded by the Korea government (MSIT) (Nos. RS-2024-00457882, 2022-0-01045, 2022-0-00680); the Advanced GPU Utilization Support Program funded by the Government of the Republic of Korea (Ministry of Science and ICT); and a grant partly supported by both IITP (MSIT) and Korea Evaluation Institute of Industrial Technology (KEIT) (MOTIE) (No. RS-2025-02217259).

\bibliography{references}

\clearpage
\newpage
\appendix

\appendix

\section*{Supplementary Material}

\vspace{1em}
\hrule
\vspace{1em}

\setcounter{section}{0}
\setcounter{figure}{0}
\renewcommand{\thetable}{A\arabic{table}}
\setcounter{table}{0}
\renewcommand{\thefigure}{A\arabic{figure}}
\setcounter{equation}{0}
\renewcommand{\theequation}{A\arabic{equation}}





\setlength{\fboxsep}{0pt}

\begin{figure*}[t]


    \begin{subfigure}[t]{0.245\textwidth}
        \centering
        \fbox{\includegraphics[valign=c,width=\textwidth]{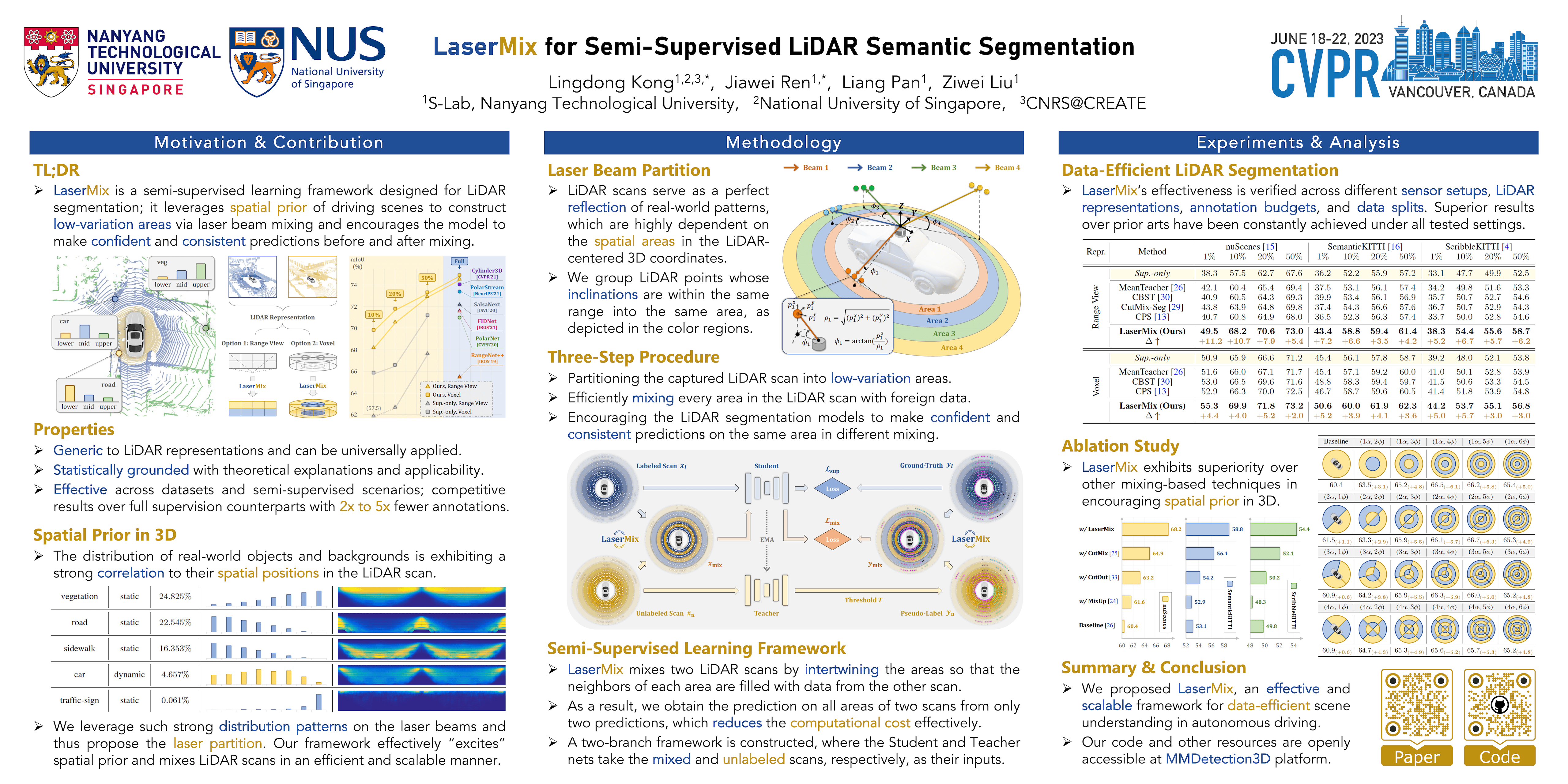}}
    \end{subfigure}
    \begin{subfigure}[t]{0.245\textwidth}
        \centering
        \fbox{\includegraphics[valign=c,width=\textwidth]{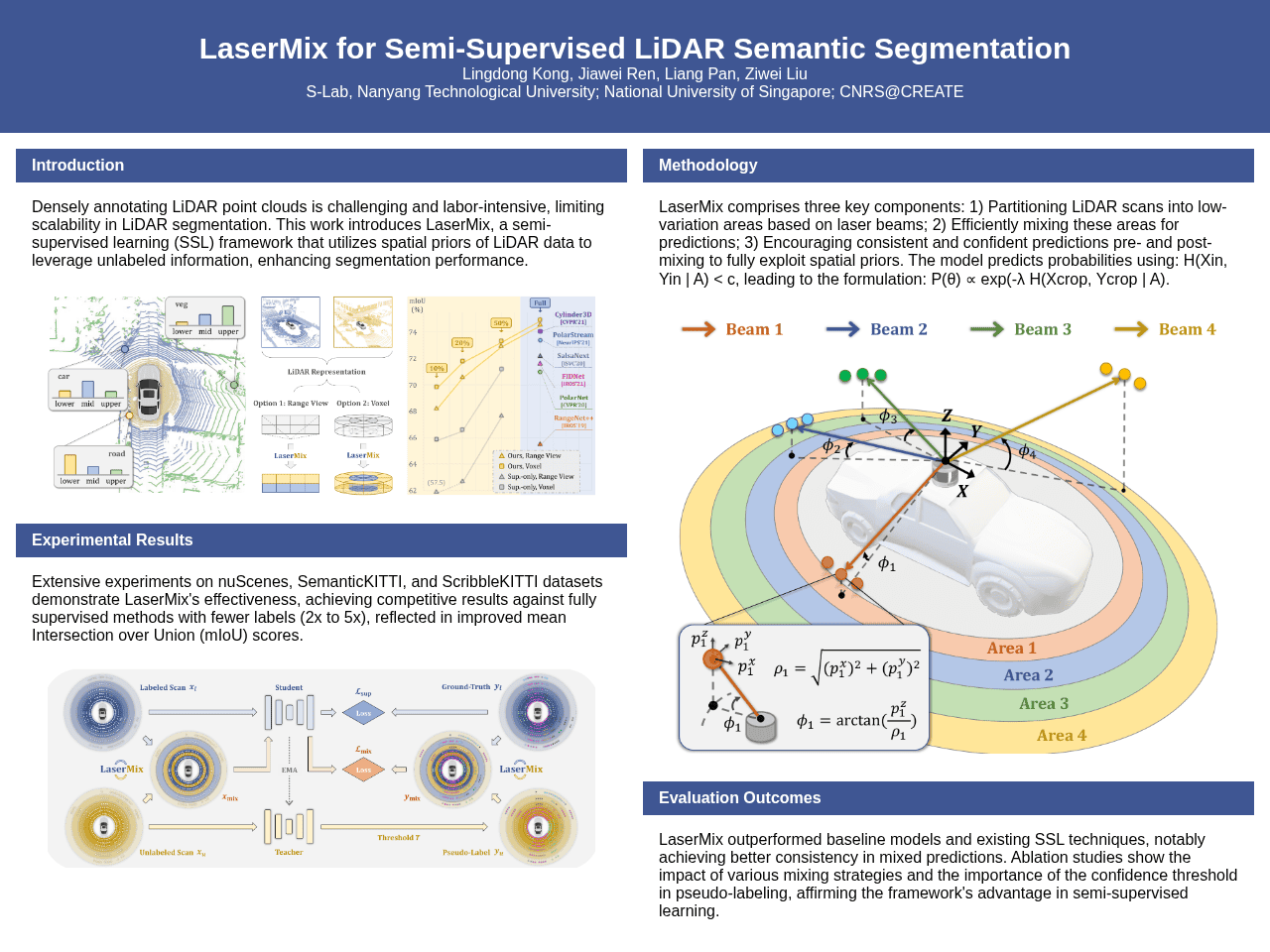}}
    \end{subfigure}
    \begin{subfigure}[t]{0.245\textwidth}
        \centering
        \fbox{\includegraphics[valign=c,width=\textwidth]{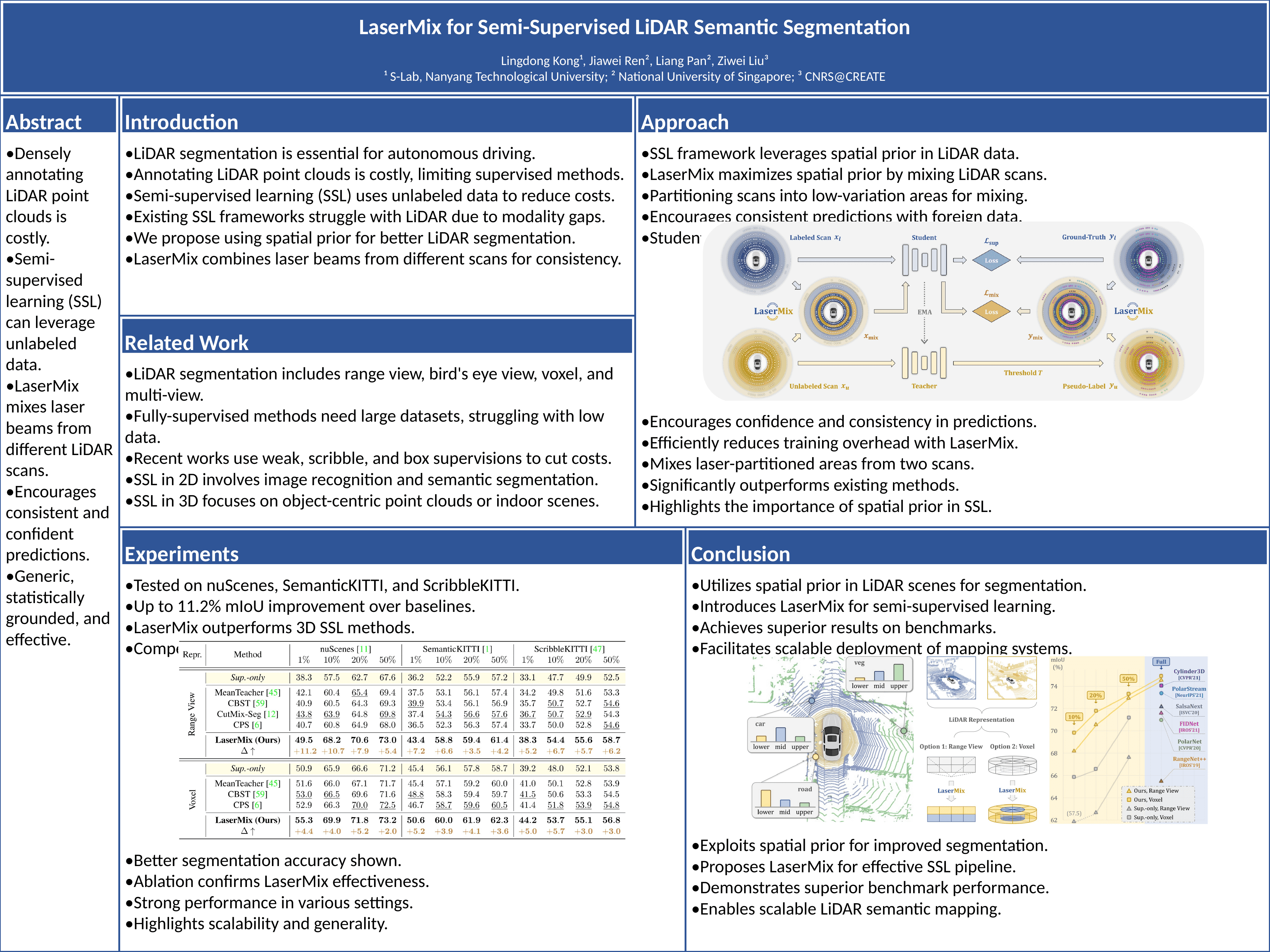}}
    \end{subfigure}
    \begin{subfigure}[t]{0.245\textwidth}
        \centering
        \fbox{\includegraphics[valign=c,width=\textwidth]{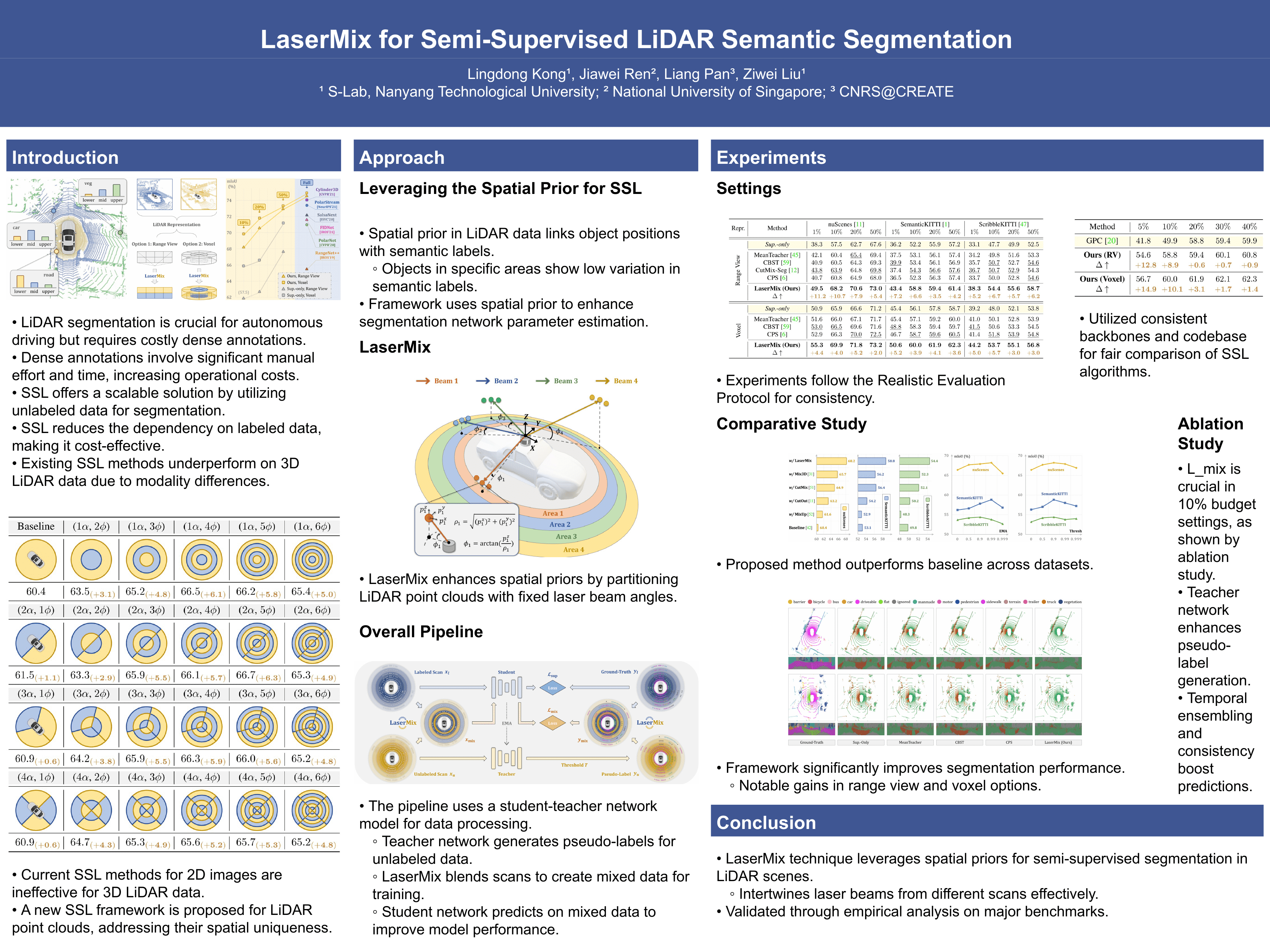}}
    \end{subfigure}


    \begin{subfigure}[t]{0.245\textwidth}
        \centering
        \fbox{\includegraphics[valign=c,width=\textwidth]{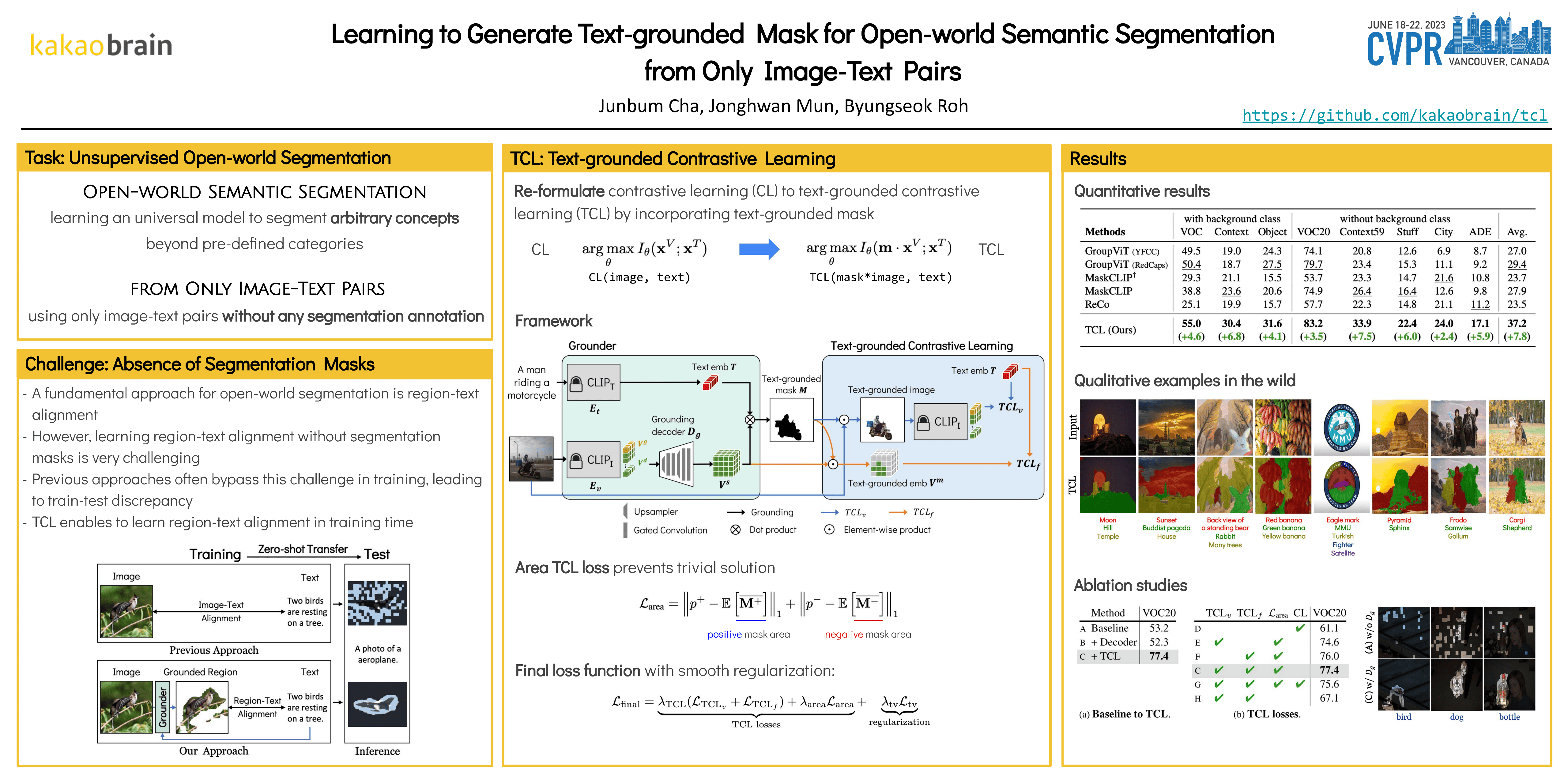}}
    \end{subfigure}
    \begin{subfigure}[t]{0.245\textwidth}
        \centering
        \fbox{\includegraphics[valign=c,width=\textwidth]{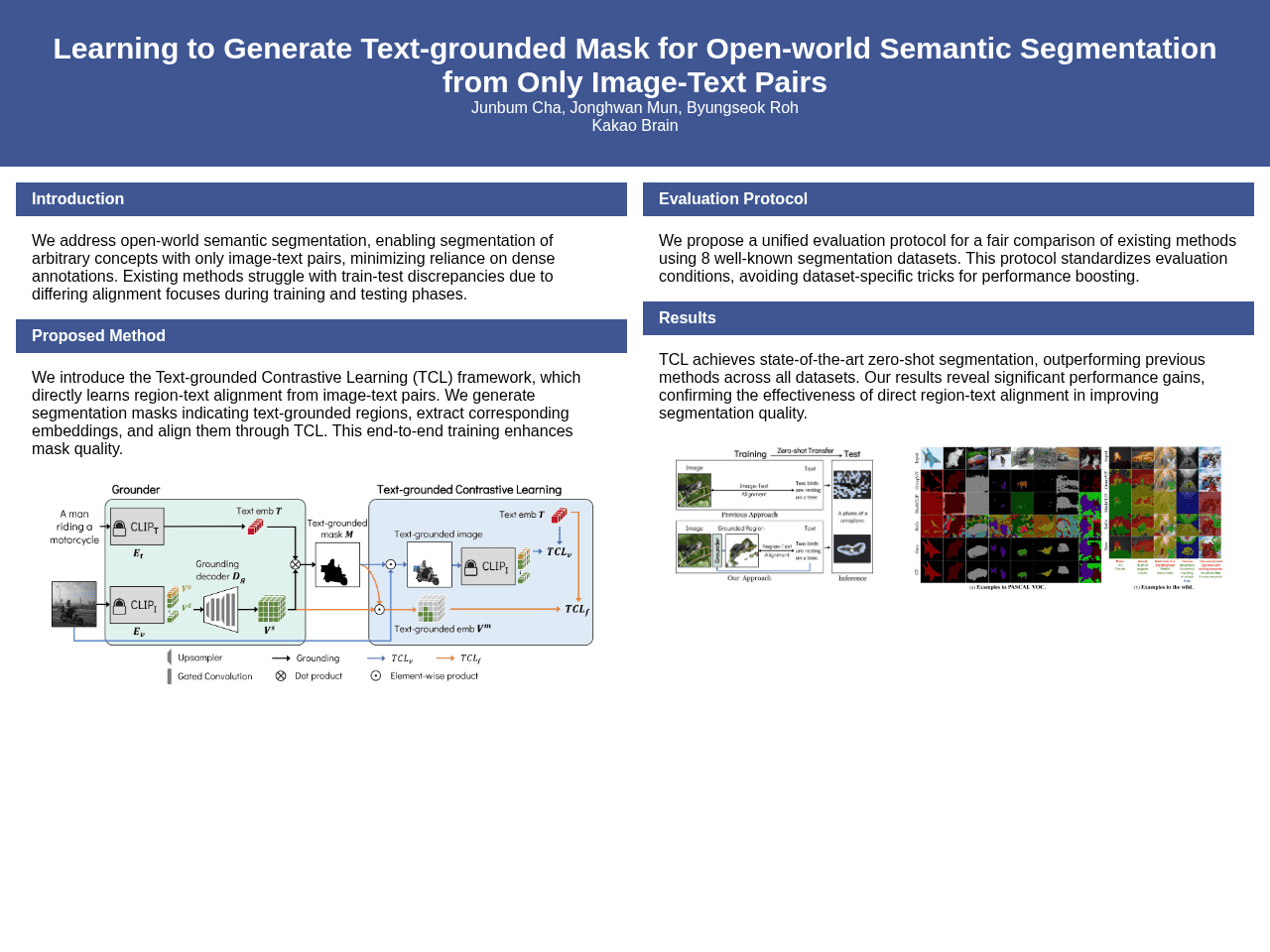}}
    \end{subfigure}
    \begin{subfigure}[t]{0.245\textwidth}
        \centering
        \fbox{\includegraphics[valign=c,width=\textwidth]{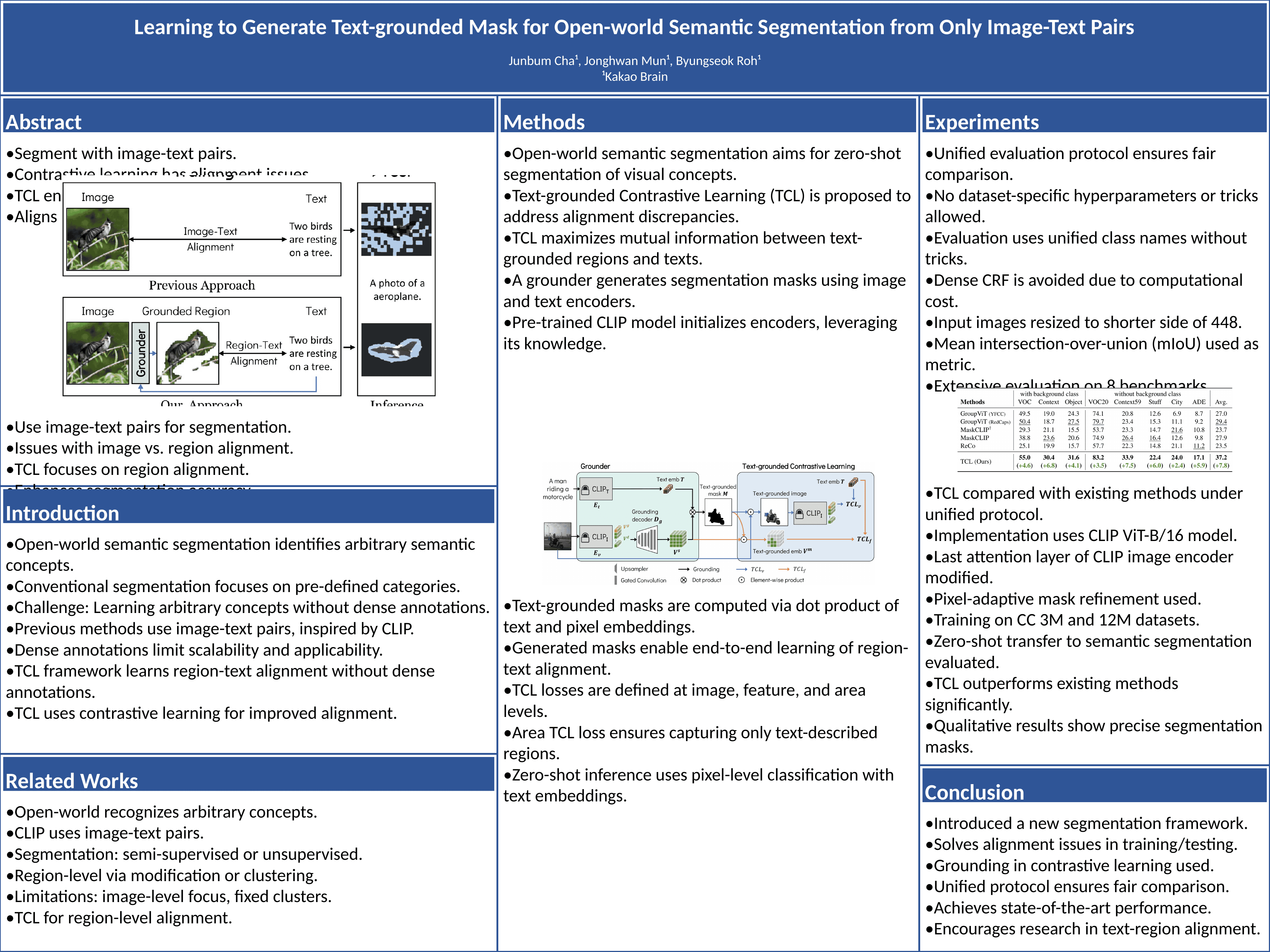}}
    \end{subfigure}
    \begin{subfigure}[t]{0.245\textwidth}
        \centering
        \fbox{\includegraphics[valign=c,width=\textwidth]{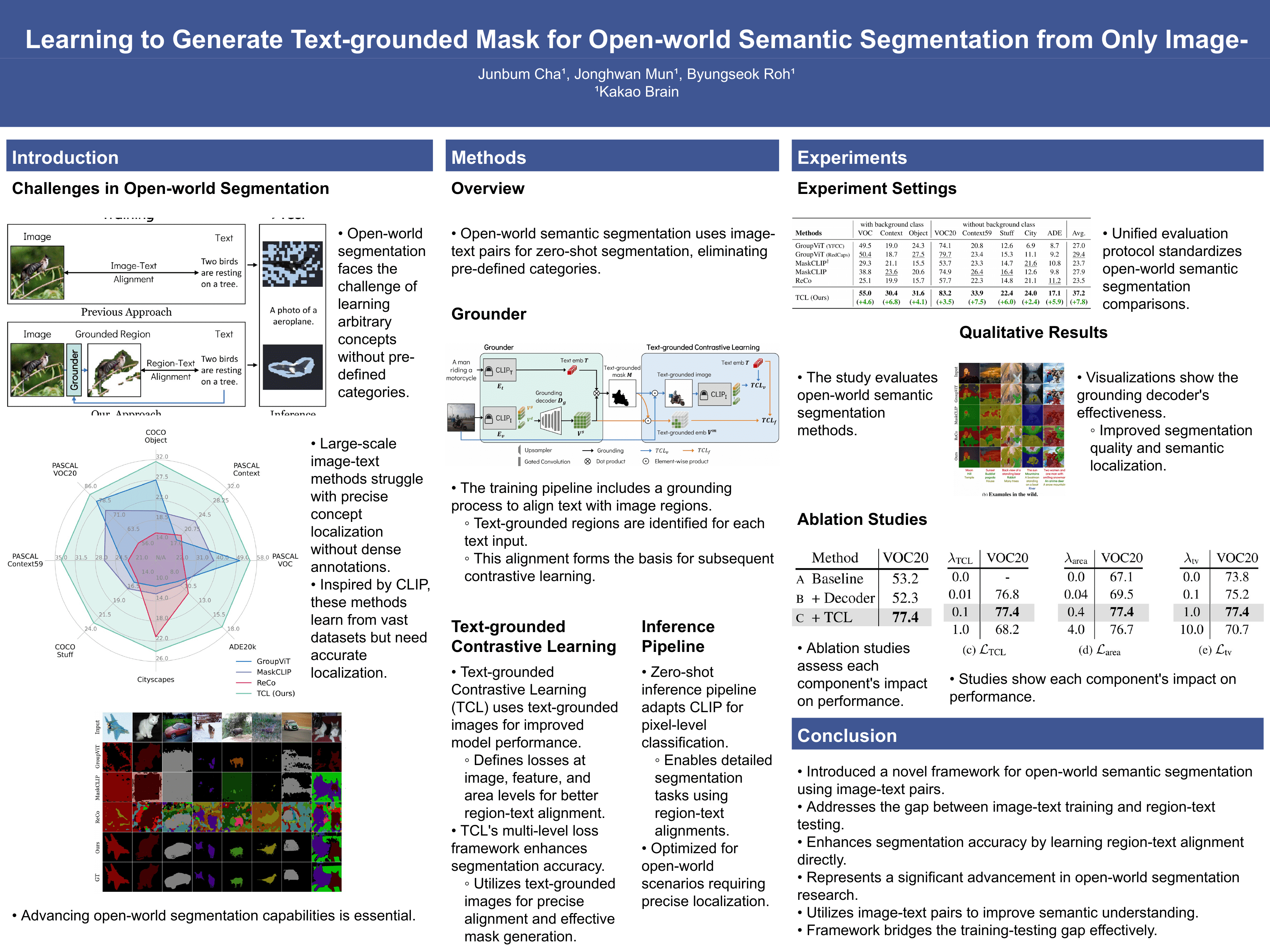}}
    \end{subfigure}
    

    \begin{subfigure}[t]{0.245\textwidth}
        \centering
        \fbox{\includegraphics[valign=c,width=\textwidth]{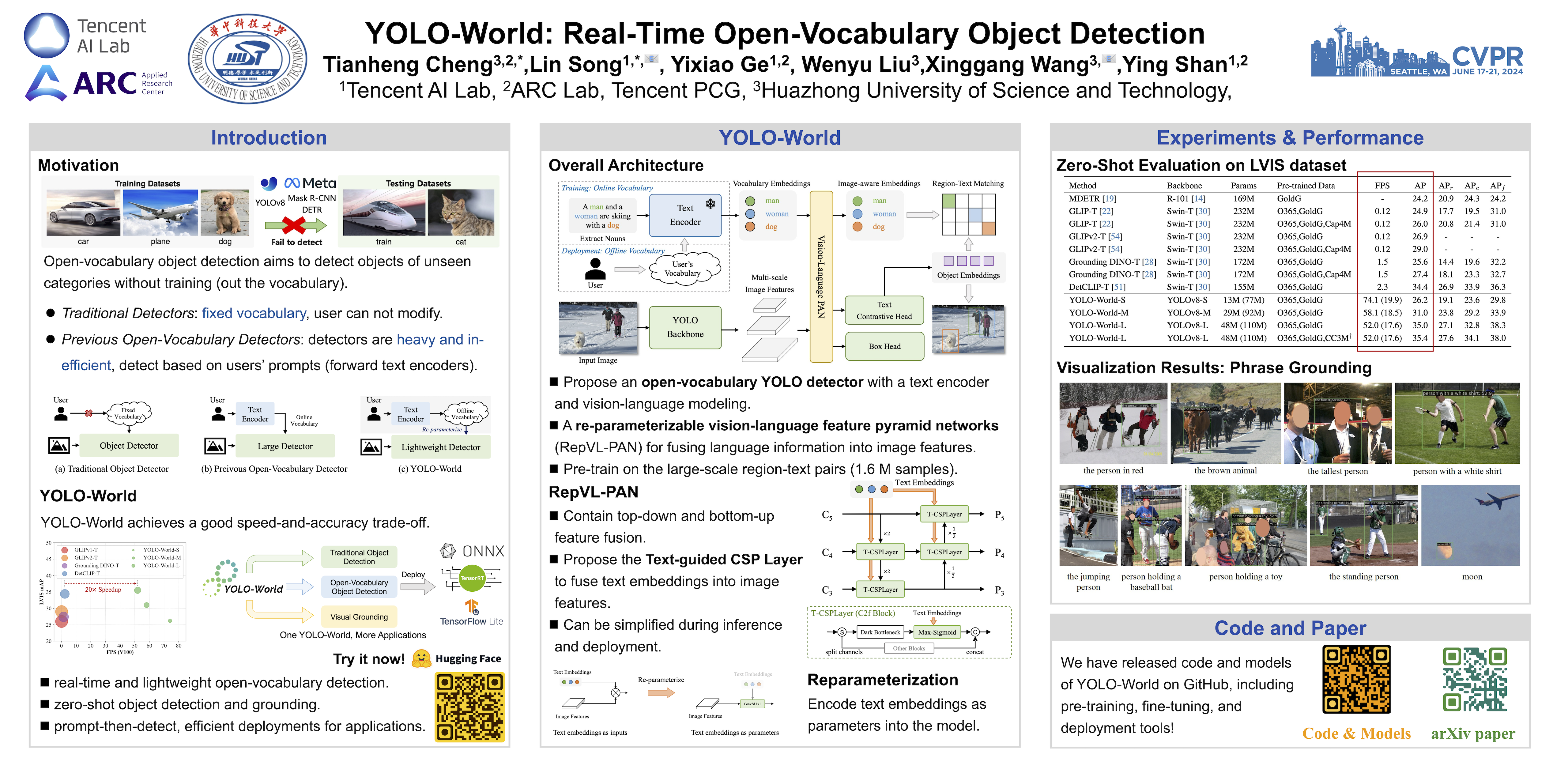}}
    \end{subfigure}
    \begin{subfigure}[t]{0.245\textwidth}
        \centering
        \fbox{\includegraphics[valign=c,width=\textwidth]{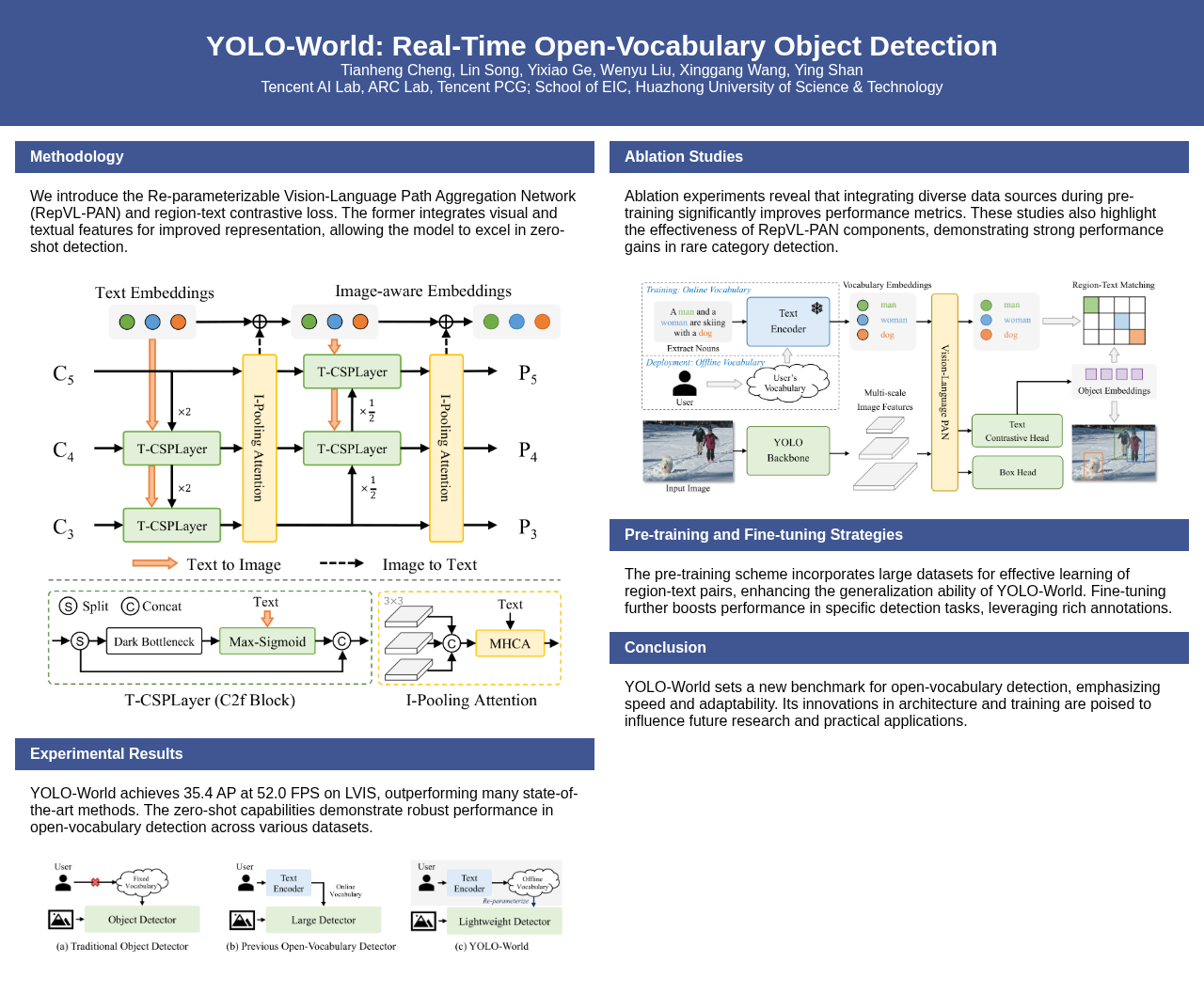}}
    \end{subfigure}
    \begin{subfigure}[t]{0.245\textwidth}
        \centering
        \fbox{\includegraphics[valign=c,width=\textwidth]{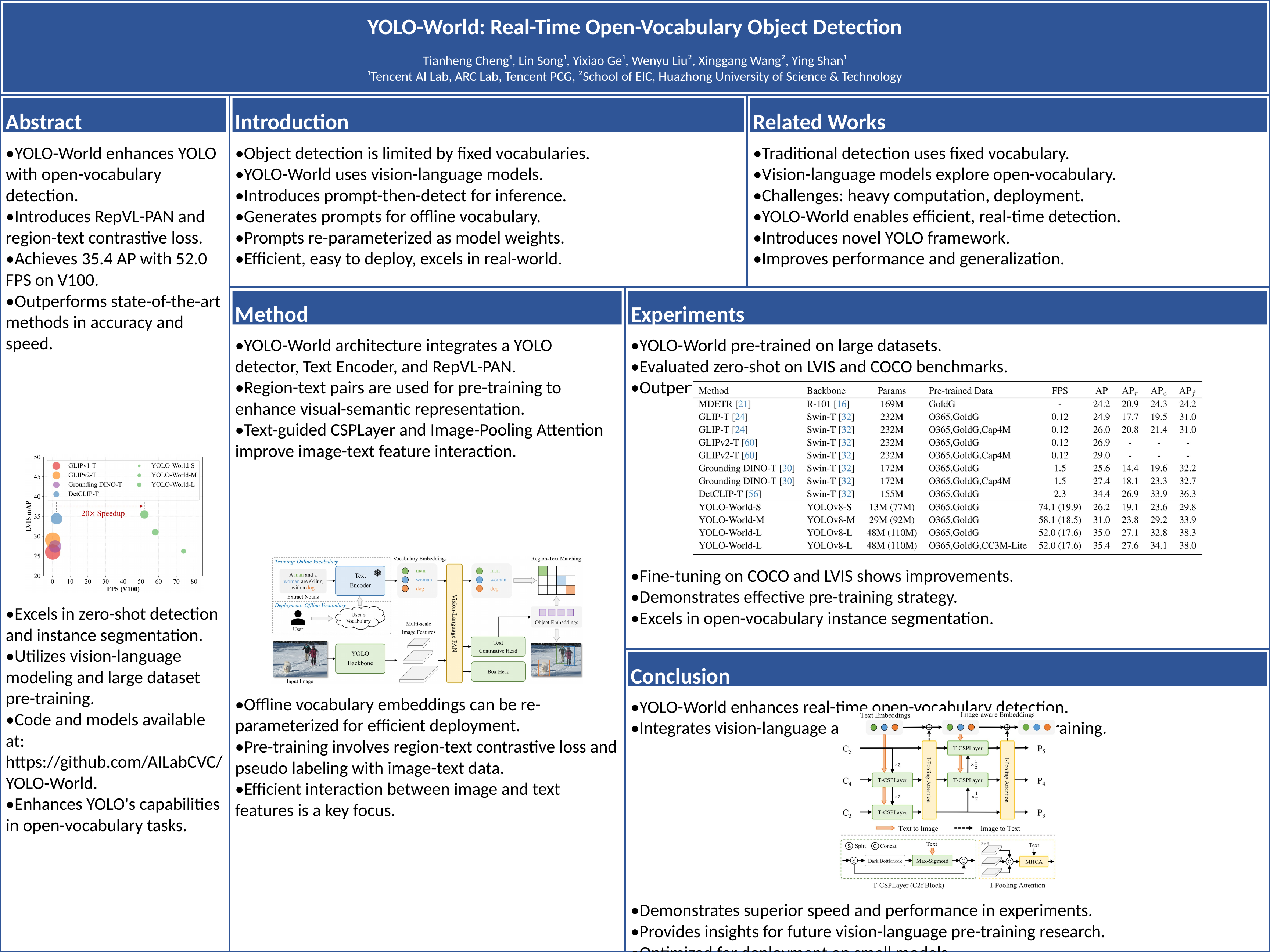}}
    \end{subfigure}
    \begin{subfigure}[t]{0.245\textwidth}
        \centering
        \fbox{\includegraphics[valign=c,width=\textwidth]{assets/06_supplementary_qualitative/CVPR2024_YOLOWorld/poster_ours_new.png}}
    \end{subfigure}

    
    \caption*{
        \makebox[0.245\textwidth]{(a) GT}\hfill
        \makebox[0.245\textwidth]{(b) P2P}\hfill
        \makebox[0.245\textwidth]{(c) Paper2Poster}\hfill
        \makebox[0.245\textwidth]{(d) Ours}\hfill
    }

    \caption{
        \textbf{Additional Qualitative Comparison [1/2].}
        Posters generated with the GPT-4o framework of baseline methods and \textit{PosterForest}, based on papers spanning different AI conferences, along with the original posters (GT) created by the authors.
    }
    \label{fig:appendix_qualitative_comparison_1}
\end{figure*}



\begin{figure*}[t]

    
    \begin{subfigure}[t]{0.245\textwidth}
        \centering
        \fbox{\includegraphics[valign=c,width=\textwidth]{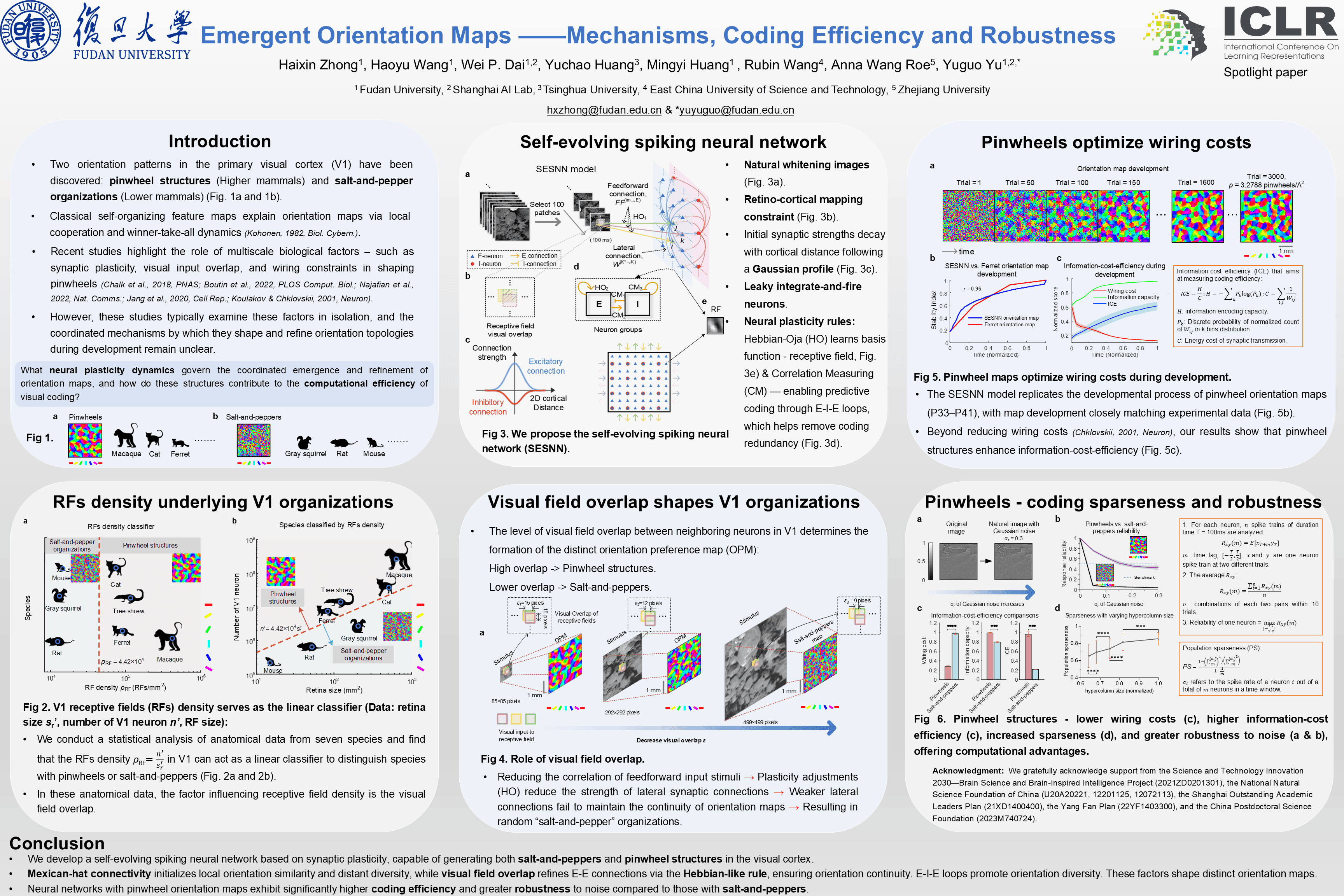}}
    \end{subfigure}
    \begin{subfigure}[t]{0.245\textwidth}
        \centering
        \fbox{\includegraphics[valign=c,width=\textwidth]{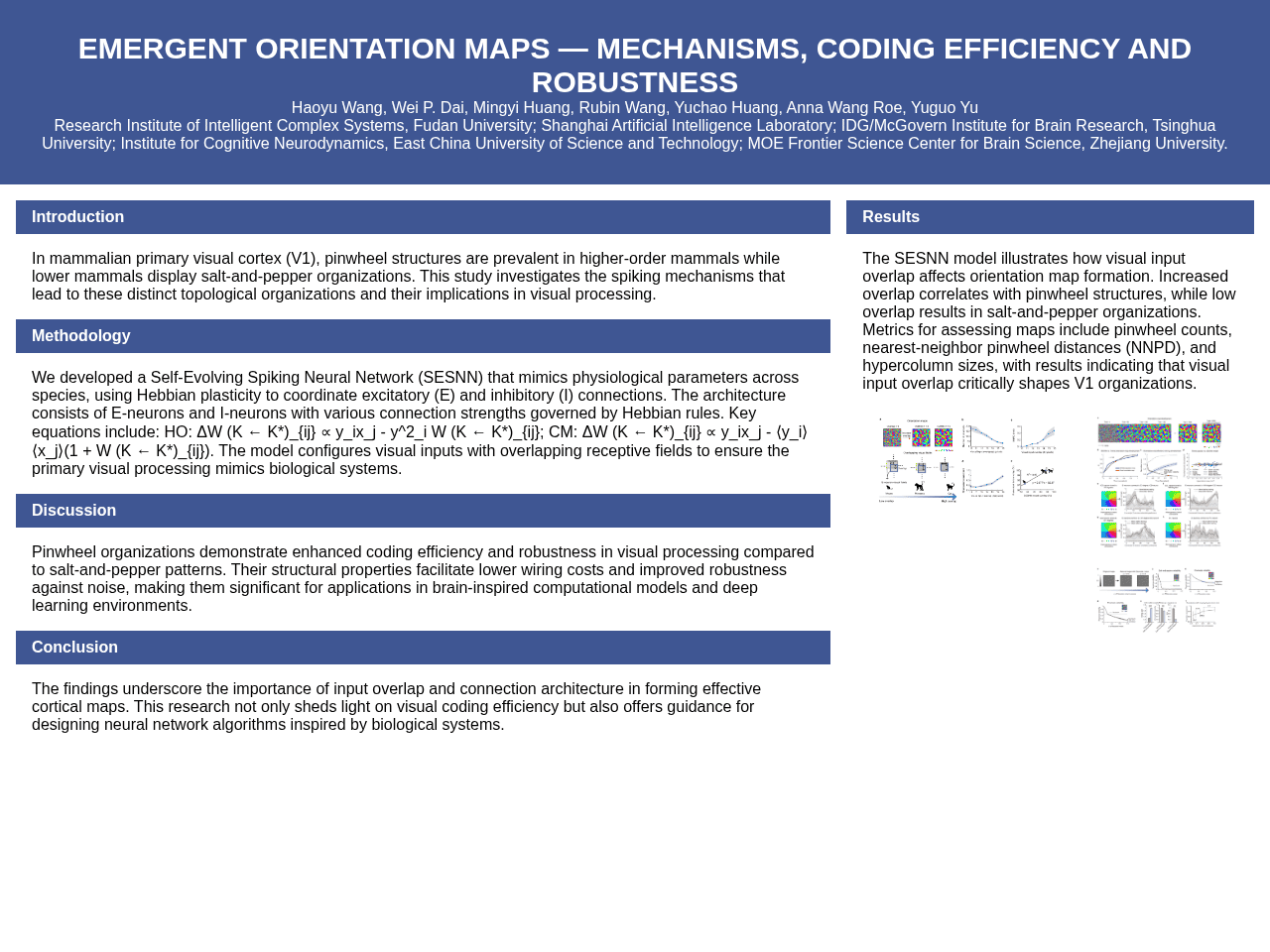}}
    \end{subfigure}
    \begin{subfigure}[t]{0.245\textwidth}
        \centering
        \fbox{\includegraphics[valign=c,width=\textwidth]{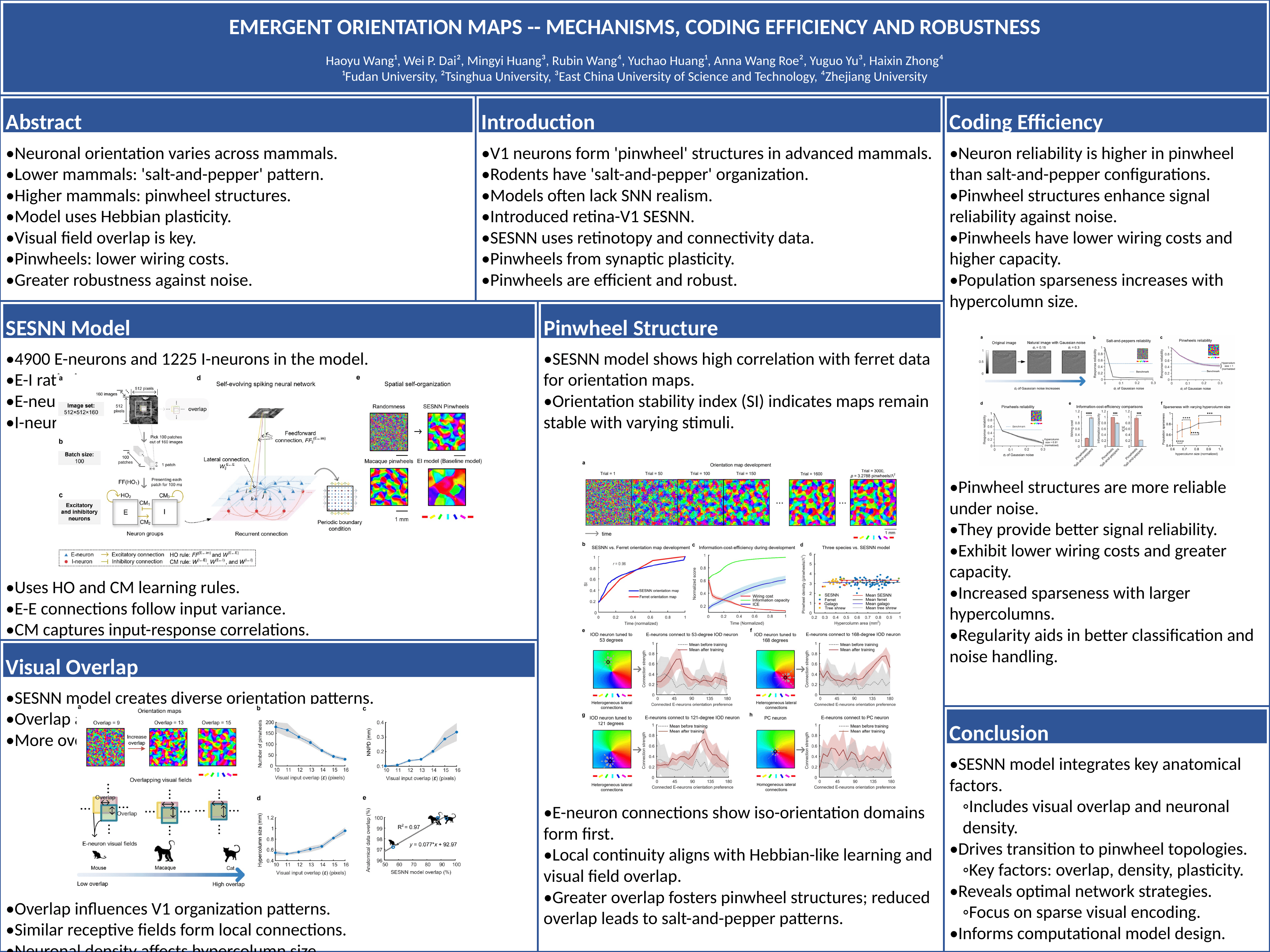}}
    \end{subfigure}
    \begin{subfigure}[t]{0.245\textwidth}
        \centering
        \fbox{\includegraphics[valign=c,width=\textwidth]{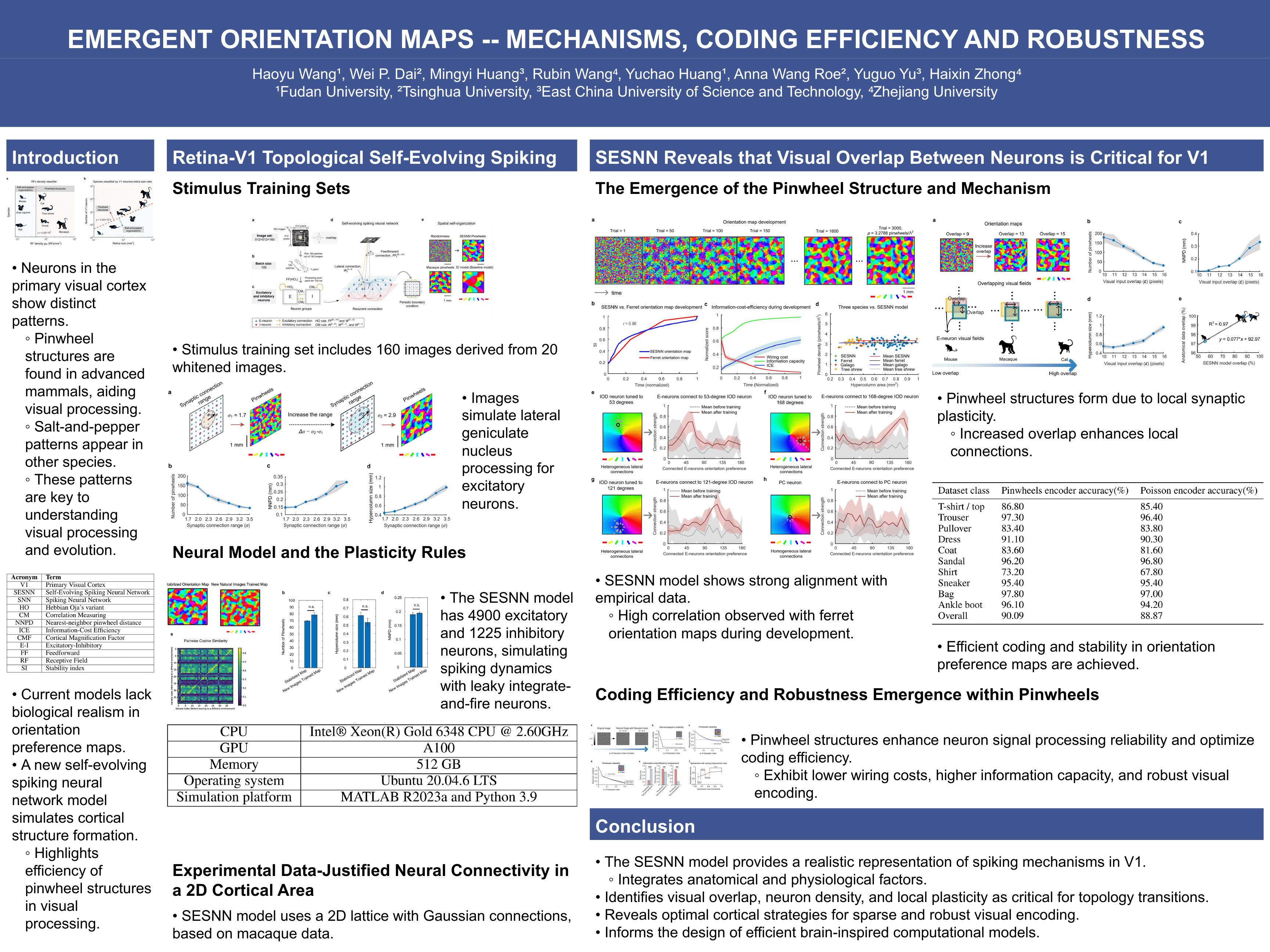}}
    \end{subfigure}


    \begin{subfigure}[t]{0.245\textwidth}
        \centering
        \fbox{\includegraphics[valign=c,width=\textwidth]{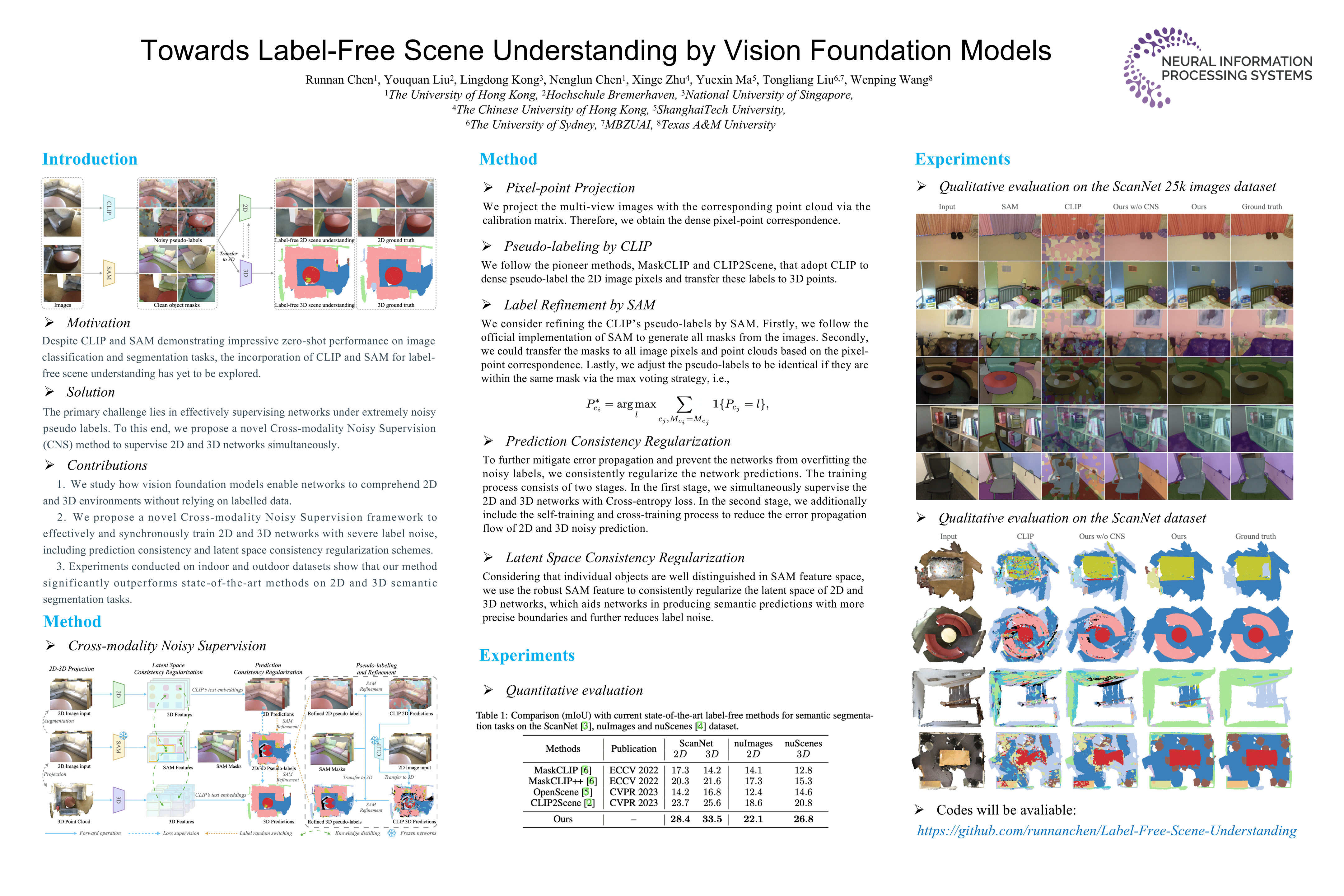}}
    \end{subfigure}
    \begin{subfigure}[t]{0.245\textwidth}
        \centering
        \fbox{\includegraphics[valign=c,width=\textwidth]{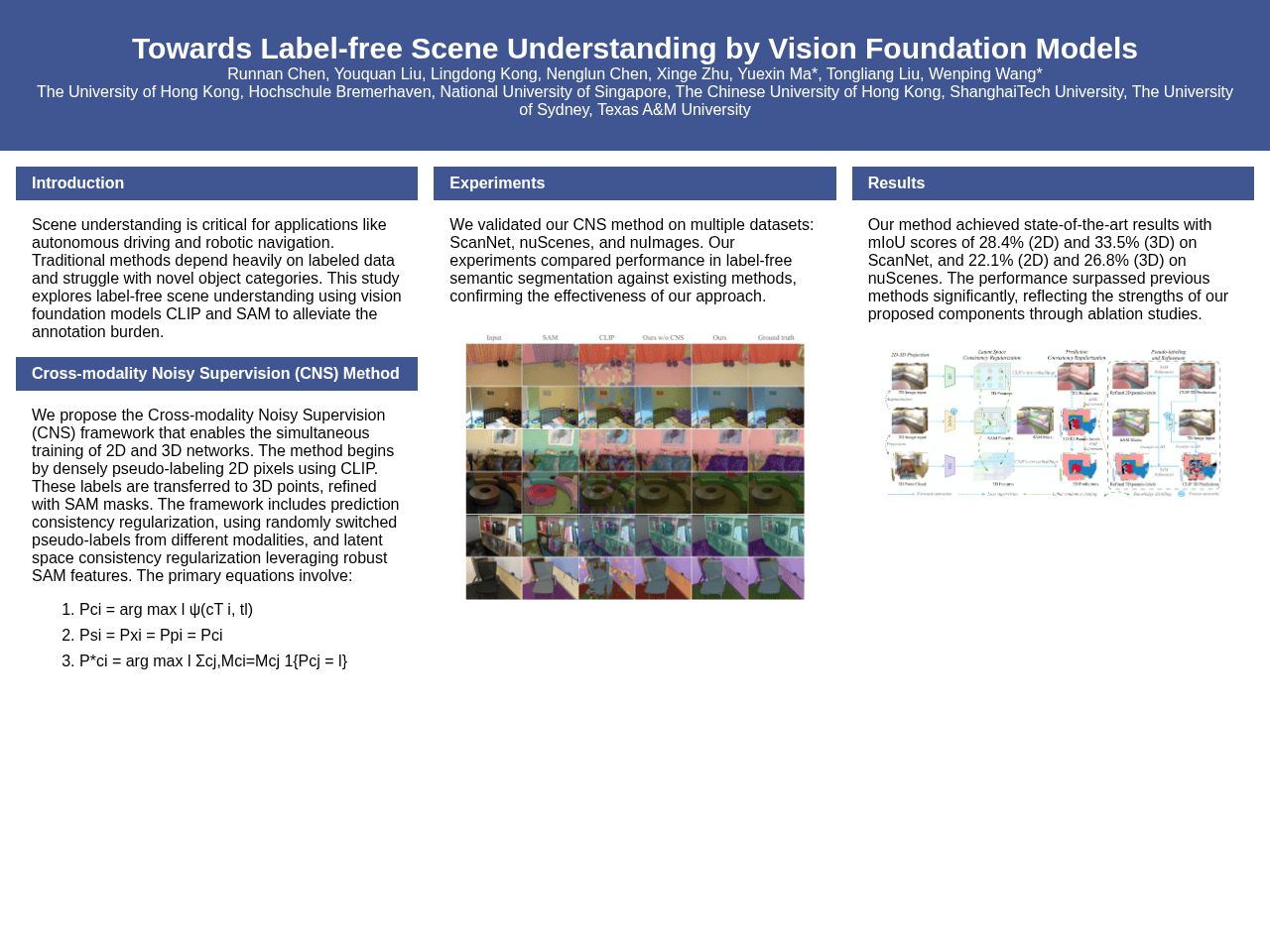}}
    \end{subfigure}
    \begin{subfigure}[t]{0.245\textwidth}
        \centering
        \fbox{\includegraphics[valign=c,width=\textwidth]{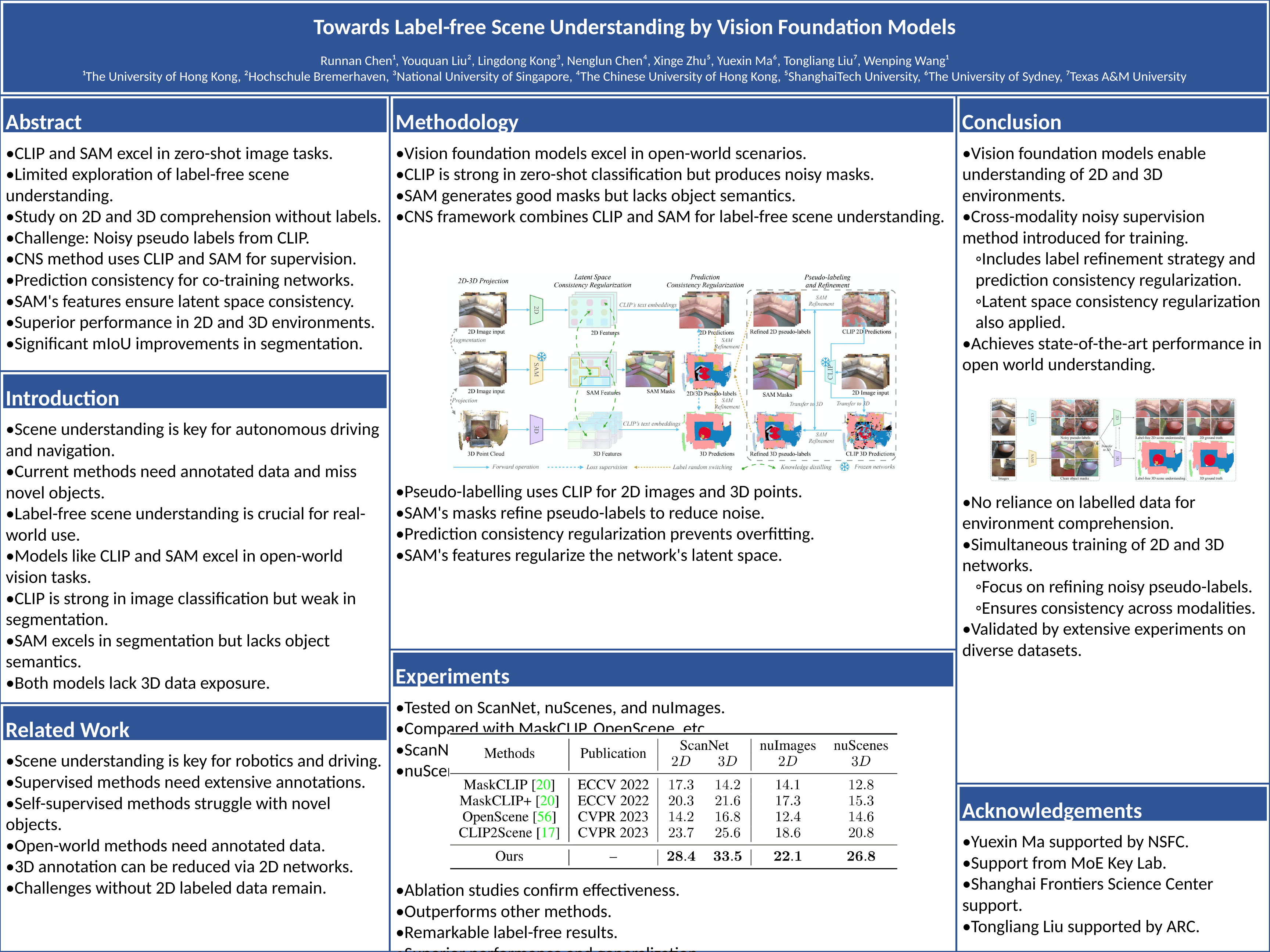}}
    \end{subfigure}
    \begin{subfigure}[t]{0.245\textwidth}
        \centering
        \fbox{\includegraphics[valign=c,width=\textwidth]{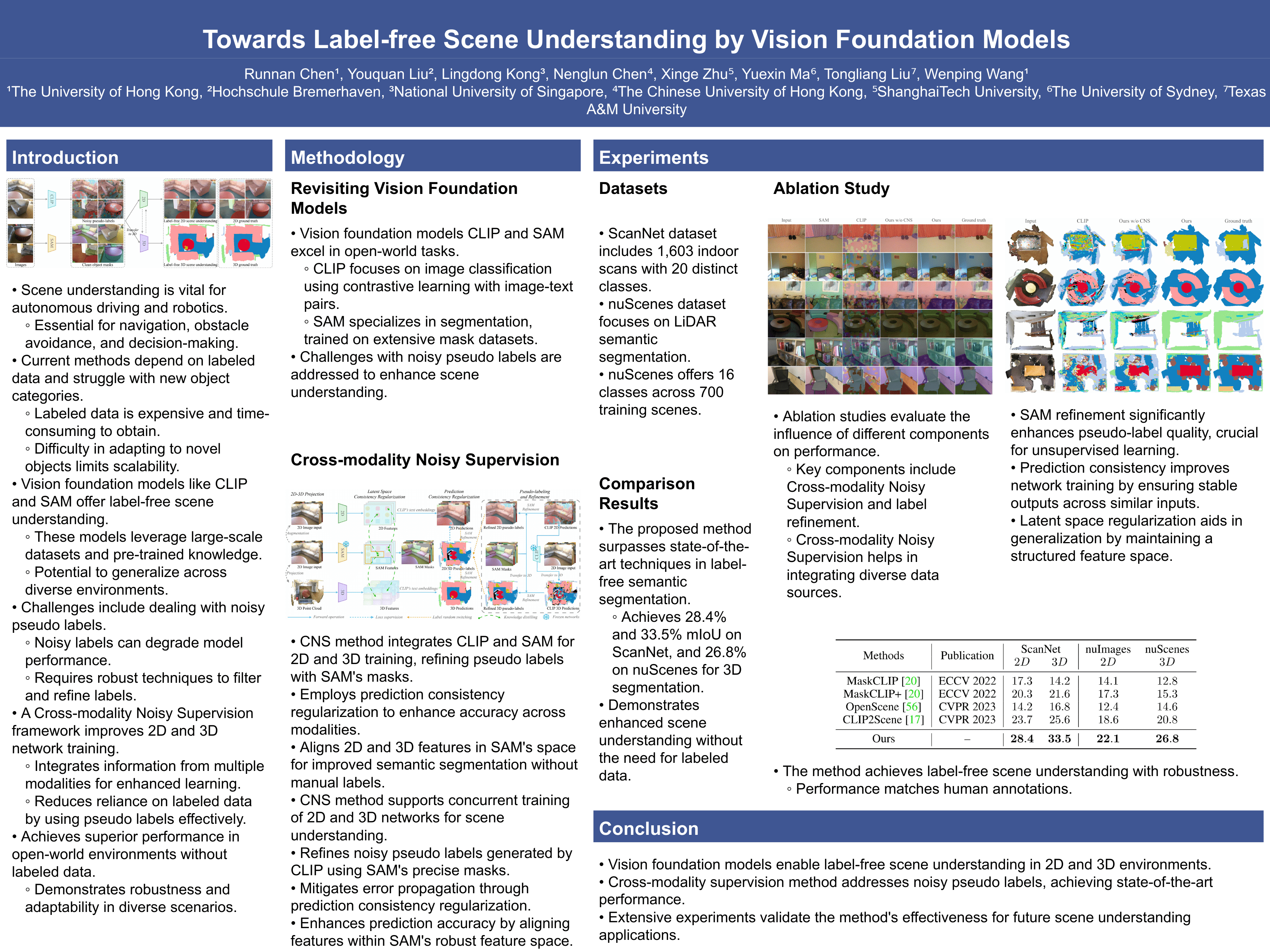}}
    \end{subfigure}
    

    \begin{subfigure}[t]{0.245\textwidth}
        \centering
        \fbox{\includegraphics[valign=c,width=\textwidth]{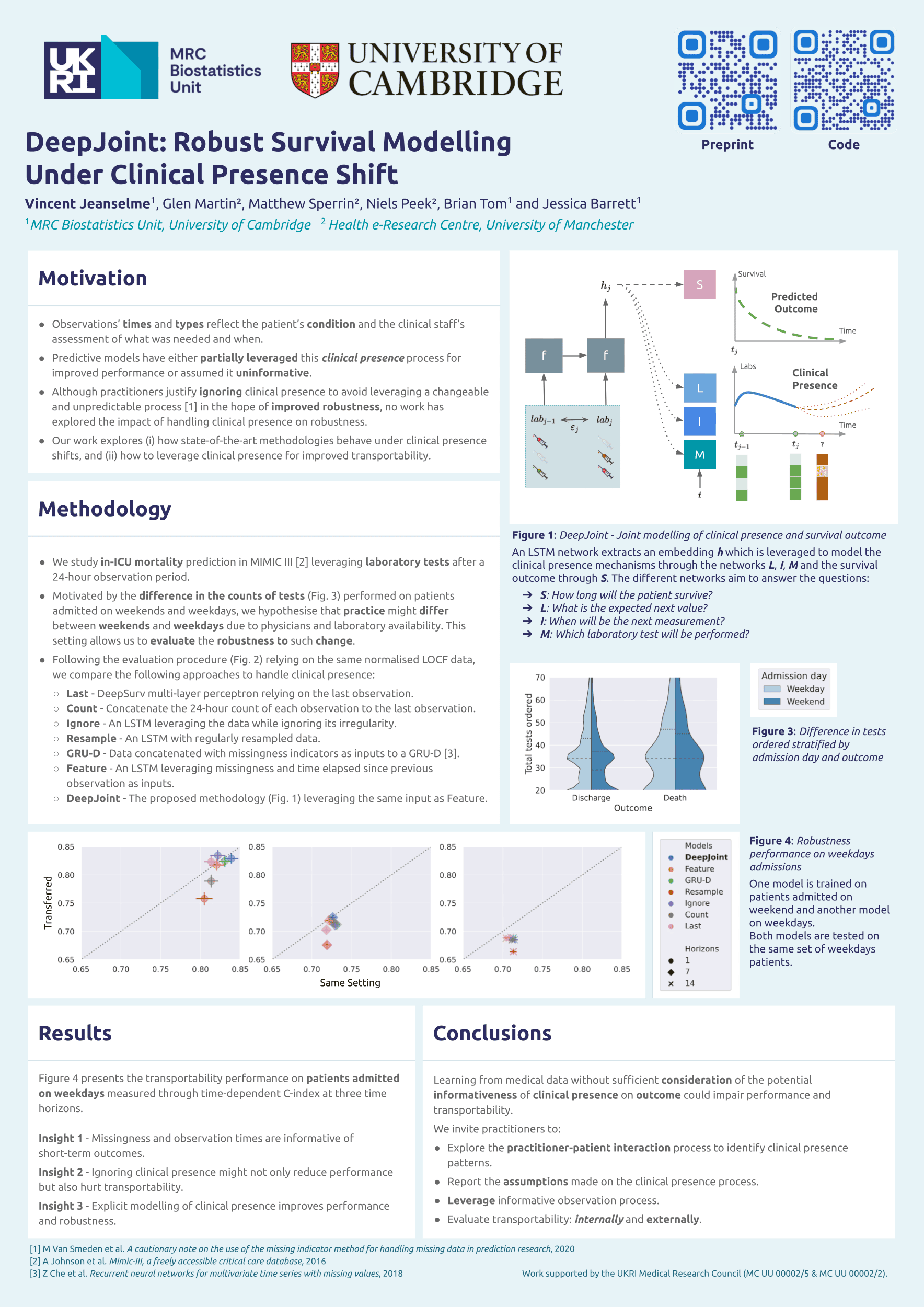}}
    \end{subfigure}
    \begin{subfigure}[t]{0.245\textwidth}
        \centering
        \fbox{\includegraphics[valign=c,width=\textwidth]{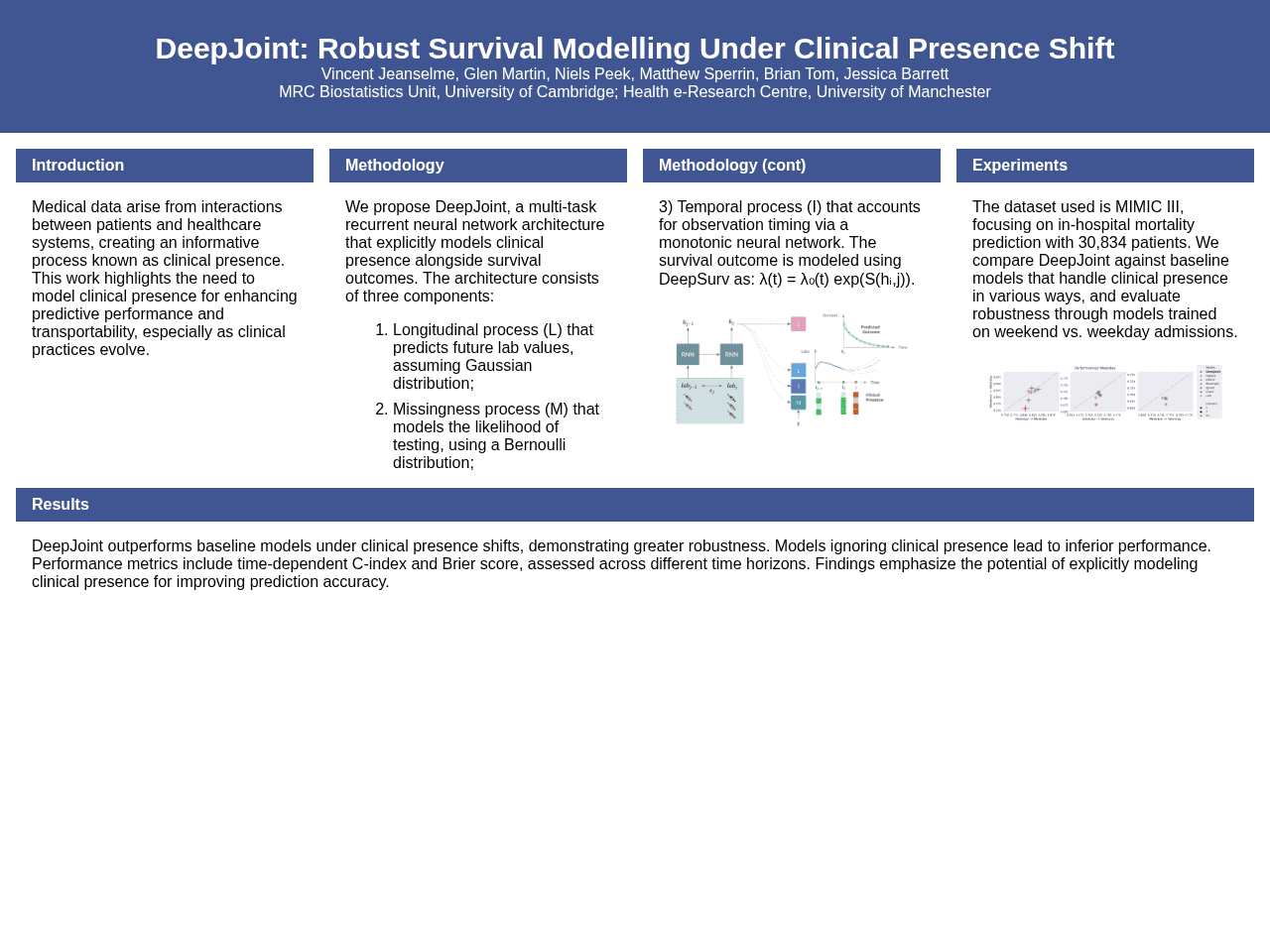}}
    \end{subfigure}
    \begin{subfigure}[t]{0.245\textwidth}
        \centering
        \fbox{\includegraphics[valign=c,width=\textwidth]{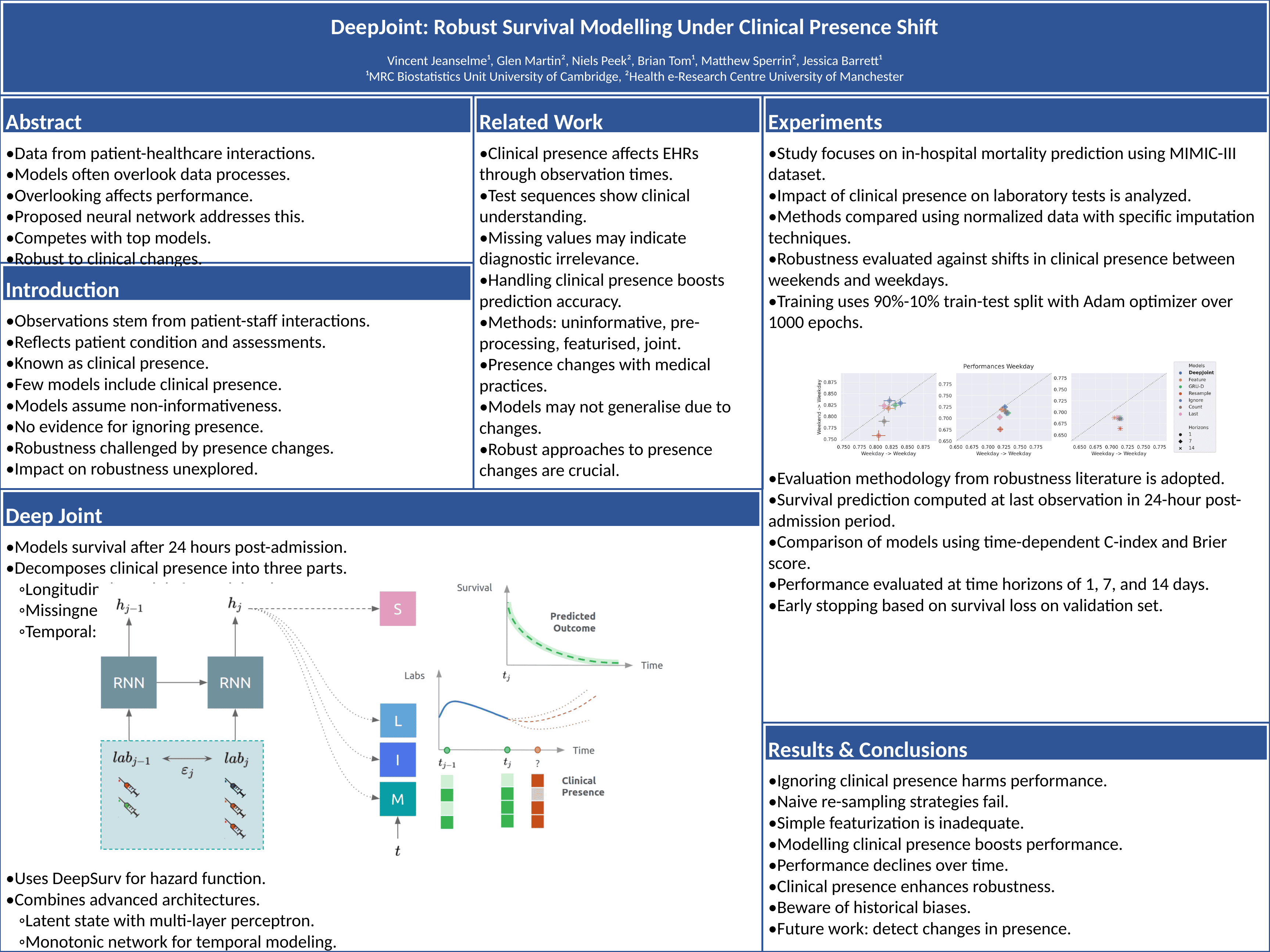}}
    \end{subfigure}
    \begin{subfigure}[t]{0.245\textwidth}
        \centering
        \fbox{\includegraphics[valign=c,width=\textwidth]{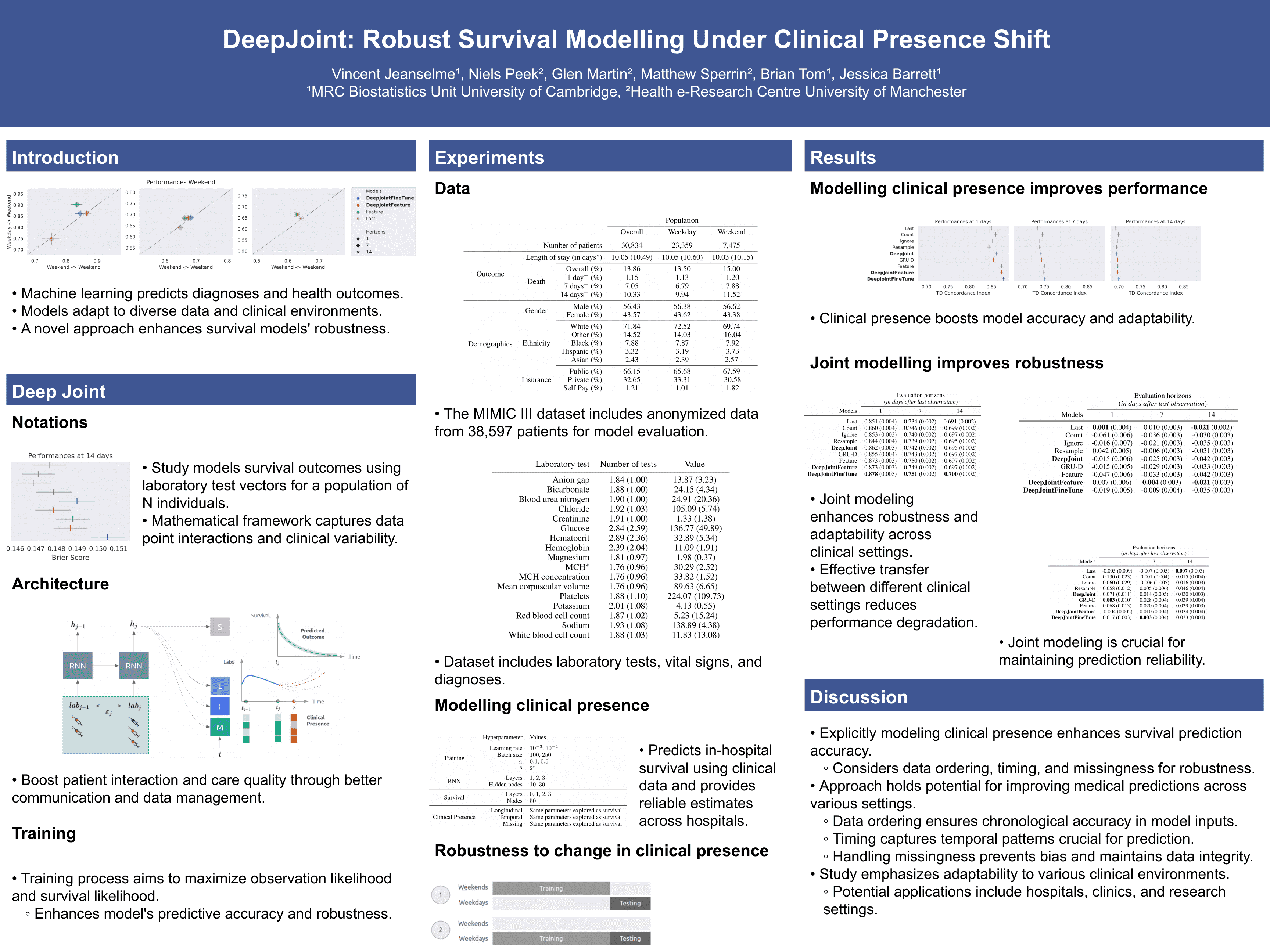}}
    \end{subfigure}
    


    \caption*{
        \makebox[0.245\textwidth]{(a) GT}\hfill
        \makebox[0.245\textwidth]{(b) P2P}\hfill
        \makebox[0.245\textwidth]{(c) Paper2Poster}\hfill
        \makebox[0.245\textwidth]{(d) Ours}\hfill
    }

    \caption{
        \textbf{Additional Qualitative Comparison [2/2].}
        Posters generated with the GPT-4o framework of baseline methods and \textit{PosterForest}, based on papers spanning different AI conferences, along with the original posters (GT) created by the authors.
    }
    \label{fig:appendix_qualitative_comparison_2}
\end{figure*}


\section{Additional Qualitative Results}

Additional results for scientific poster generation are presented in~\Cref{fig:appendix_qualitative_comparison_1} and~\Cref{fig:appendix_qualitative_comparison_2}.


\section{Experimental Details}


\subsection{Qualitative Experiments Setup}
\textbf{Standardization.}
To ensure a fair comparison across different methods, all posters were generated with a standardized width of 48 inches and height of 36 inches.
In addition, the color schemes and font styles were unified for all posters, eliminating visual biases related to design choices.
These controls allow the qualitative evaluation to focus solely on the content and layout quality produced by each model.

\subsection{Quantitative Experiments Setup}

\textbf{Dataset Details.}
For all quantitative experiments, we use the Paper2Poster benchmark~\citep{pang2025paper2poster}, which is the first large-scale dataset for scientific poster generation.
The benchmark consists of 100 paper–poster pairs curated from recent AI conferences, including NeurIPS, ICML, and ICLR (2022–2024).
Each pair includes a full-length research paper and its corresponding author-designed poster, enabling rigorous evaluation of poster generation models.

The dataset ensures high quality and diversity by selecting papers from a range of AI subfields such as computer vision, natural language processing, and reinforcement learning.
The input papers average 22.6 pages and 12,156 words, with an average of 22.6 figures per paper.
The corresponding posters contain about 774 words and 8.7 figures on average, resulting in a significant compression of both textual and visual content.

To avoid data leakage, only the test split of the POSTERSUM dataset~\citep{saxena2025postersum} was used when curating the benchmark.
Papers were filtered to include the latest camera-ready versions and to ensure a broad distribution across years and venues.
This benchmark provides a challenging setting for evaluating poster generation methods in terms of both informativeness and layout quality.

\subsection{Implementation Details}

All experiments were conducted on a server running Ubuntu 22.04 LTS, equipped with an AMD EPYC 7543 CPU and eight NVIDIA RTX A6000 GPUs (48 GB each).
Four GPUs were allocated to Qwen-2.5-7B-Instruct and four to Qwen2.5-VL-7B-Instruct, using PyTorch 2.6.0.
Random seeds were fixed for Python, NumPy, and PyTorch to ensure experimental reproducibility.

\subsection{Hyperparameters}

The maximum number of tree refinement iterations was set to $T_\texttt{max}=2$.
Pilot studies showed that additional exchanges between agents did not yield significant improvements.
These hyperparameters were chosen to balance generation quality and computational efficiency.


\subsection{Detailed Baseline Setup}

For figure extraction in P2P, we adopted DocLayout-YOLO~\cite{zhao2024doclayout}.
The PosterAgent model in Paper2Poster was implemented in two variants: one utilizing $\texttt{GPT-4o}$ and the other employing open-source Qwen models, specifically Qwen-2.5-7B and Qwen-2.5-VL-7B.

\section{Evaluation Metric and Rationale}

\subsection{Reason for Metric Selection}
Our evaluation metrics were selected to comprehensively assess the effectiveness of scientific poster generation from multiple perspectives. Human evaluation was conducted using four criteria: content fidelity, visual harmony, structural effectiveness, and overall completeness. These metrics reflect the need for posters to accurately summarize findings, present information attractively and logically, and maintain a professional appearance.

Although MLLM-as-a-judge metrics (e.g., GPT-4o) enable fine-grained assessment of informativeness and aesthetics~\citep{pang2025paper2poster}, they have limitations. Discrepancies between ground truth and generated posters are often better detected by human observers than by automated metrics. However, with advances in multimodal evaluation, MLLM-based metrics are expected to become more reliable.

\subsection{User Study Details}\label{user_study}

We recruited 25 graduate students with experience in academic conferences. Each participant evaluated 10 pairs of papers and posters generated by four methods: GPT-4o HTML, P2P~\citep{sun2025p2p}, Paper2Poster~\citep{pang2025paper2poster}, and PosterForest.
Posters were standardized for font and color.
Participants viewed the original poster and four generated posters in randomized order, judging based on content fidelity, aesthetic quality, structural clarity, and overall quality.

\section{Limitations and Future Work}






\subsection{Failure Cases}

While our hierarchical approach improves global understanding of paper structure, there remain notable limitations compared to human designers.
In particular, our framework struggles when processing papers containing a large number of figures or tables, especially when such assets are densely clustered or heavily interleaved within the text.
These scenarios often result in errors during asset parsing and content matching, which can propagate through subsequent stages of the pipeline.
As a result, missing or misaligned visual elements accumulate, further degrading the informativeness and visual coherence of the generated posters.

\begin{figure}[ht]
    \centering
    \fbox{
        \begin{subfigure}{\linewidth}
            \centering
            \includegraphics[width=0.5\textwidth]{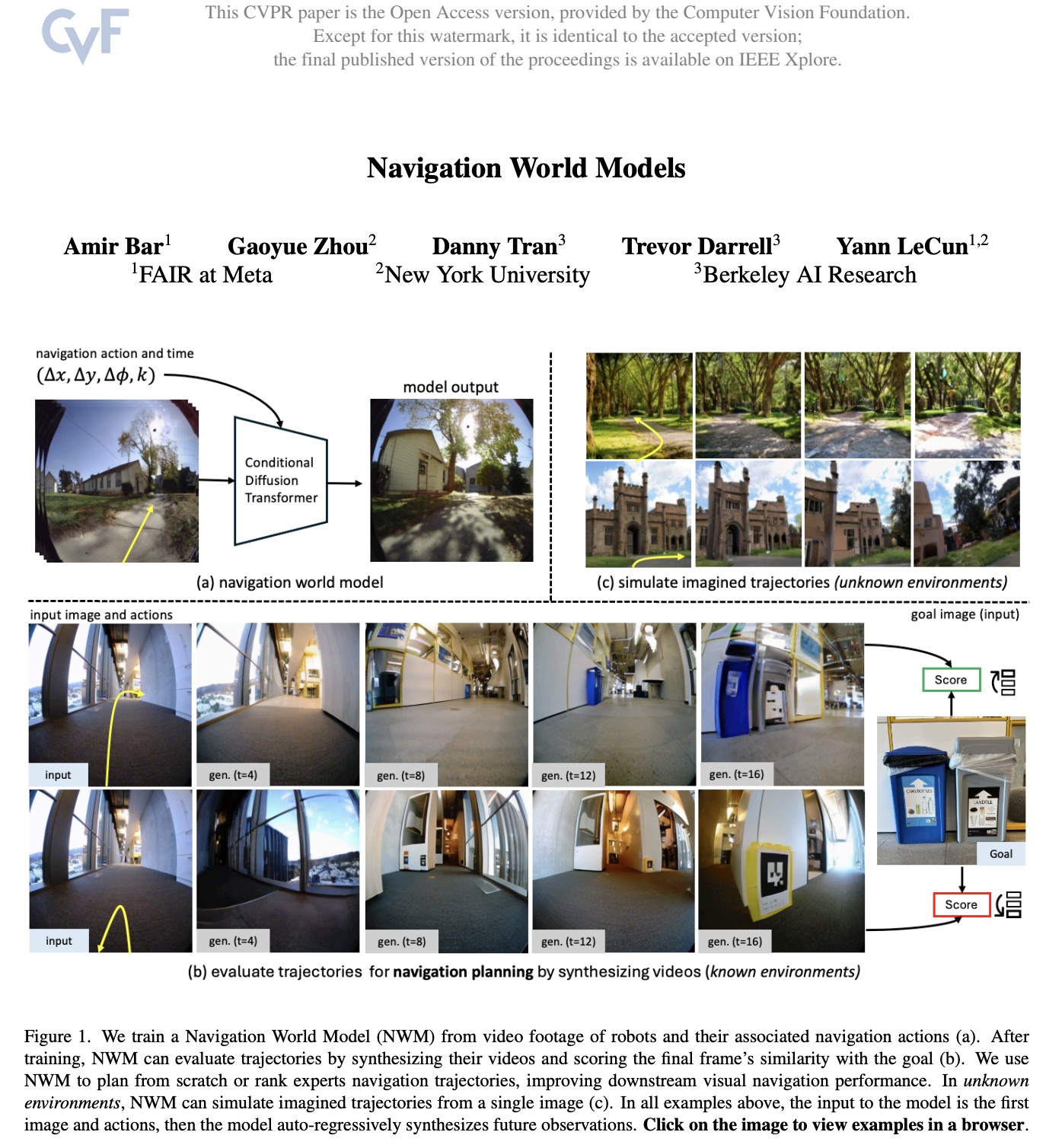}
            \subcaption{Example input paper containing a dense cluster of figures.}
            \label{fig:parser_failure_clustered_figures_subA}
        \end{subfigure}
    }
    \hfill
    \fbox{
        \begin{subfigure}{\linewidth}
            \includegraphics[width=1.0\textwidth]{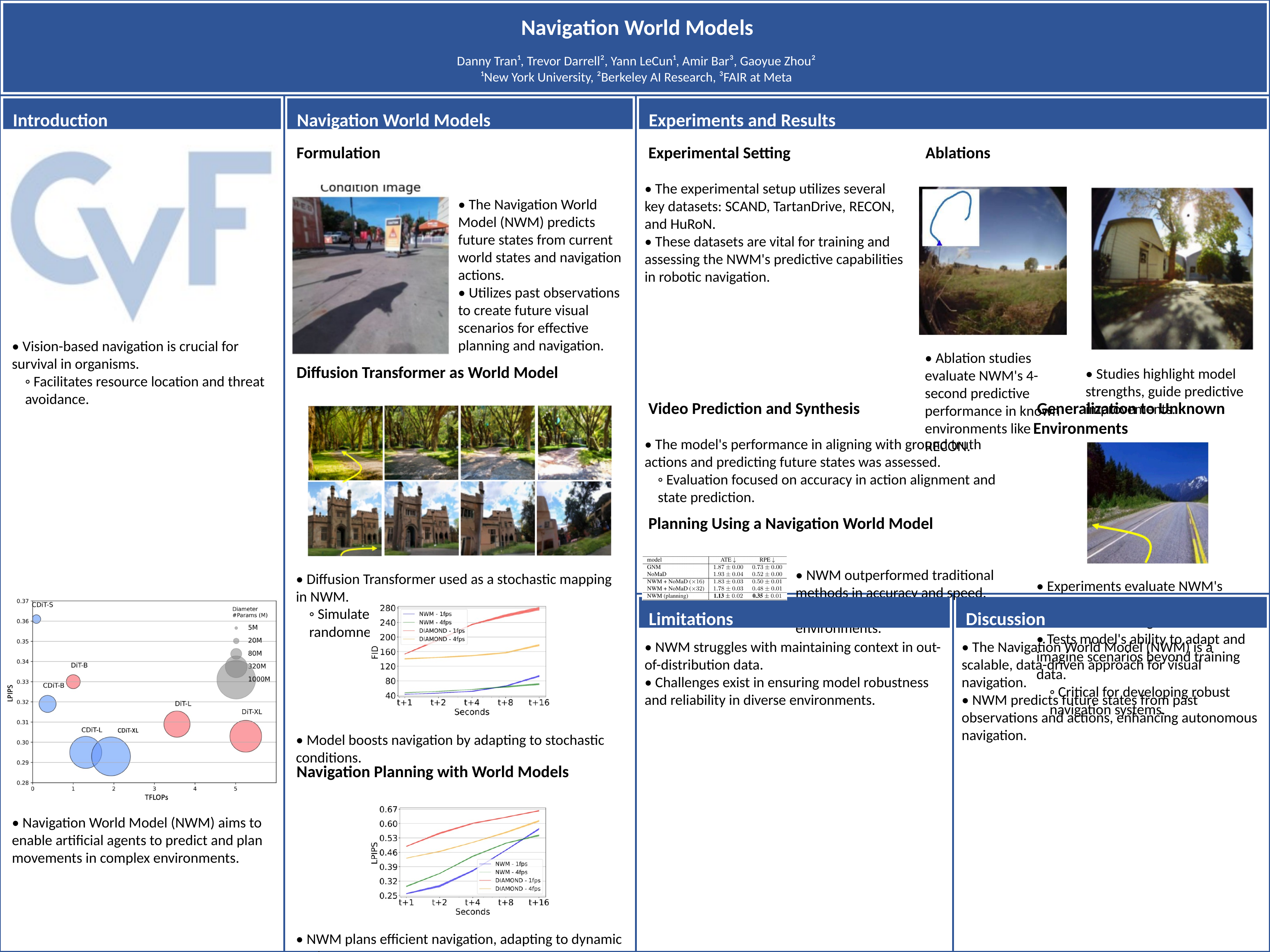}
            \subcaption{
            Generated poster with missing or misaligned visual elements due to parsing failures.
            }
            \label{fig:parser_failure_clustered_figures_subB}
        \end{subfigure}
    }
    \caption{
        An example of a failure case arising from clustered figures in the input paper.
        The parser fails to correctly extract and match all figures, leading to incomplete or incorrectly structured visual content in the poster.
    }
    \label{fig:parser_failure_clustered_figures}
\end{figure}

\Cref{fig:parser_failure_clustered_figures_subA} illustrates an example paper in which many figures are densely arranged within a short span of text.
As shown in~\Cref{fig:parser_failure_clustered_figures_subB}, the generated poster for this case exhibits clear failures, with several figures missing or incorrectly matched to their respective sections.
This example demonstrates how dense and complex visual layouts present a significant challenge for the current parsing and matching pipeline, ultimately limiting the quality of the generated output.

While improving the performance of parsers such as Docling is not the primary focus of this work, enhancing asset parsing and extraction remains an important avenue for future research.
Advances in these components could further improve the overall robustness and quality of the poster generation framework, especially when handling papers with complex and densely interleaved visual elements.

\subsection{Future Directions}

From a modeling perspective, future work will focus on enhancing the framework's ability to robustly parse and reason over complex layouts, particularly in papers with densely interleaved figures and tables.
Incorporating more advanced hierarchical modeling and visual-semantic alignment techniques may further narrow the gap with human-designed posters.
For example, human designers intuitively adjust the placement and scale of figures based on factors such as the amount of information conveyed, perceived importance, and the context of each figure, as well as considerations like font size and spatial balance.
However, our current framework does not yet fully capture the relative importance or semantic richness of individual figures and their optimal integration within the layout.
Modeling these nuanced factors in future work could enable more practical and human-aligned poster generation.

From the perspective of evaluation, we plan to build upon the metrics introduced in P2P~\citep{sun2025p2p} and Paper2Poster~\citep{pang2025paper2poster} to develop more reliable and comprehensive automated metrics for poster quality assessment.
Improving automated evaluation protocols is crucial for accurately measuring progress and guiding future research in scientific poster generation.



\section{Additional Materials}

\subsection{Parser Agent, $\mathcal{A}_\texttt{Parser}$.}

The Parser Agent parses the raw scientific document into a hierarchical tree structure, organizing components such as titles, sections, subsections, paragraphs, and visual assets for further processing by other agents.  
An example prompt is provided in \Cref{tbox:parser_agent}.


\medskip
\noindent\textbf{Collaborative Poster Optimization.}  
This criterion assesses the overall arrangement, alignment, and spacing of text and graphics to ensure a coherent and readable poster structure.  

\subsection{Summarizing Agent, $\mathcal{A}_\texttt{Summ}$.}
The Summarizing Agent is responsible for pruning, merging, and summarizing essential information from each section, subsection, and paragraph of the document.
It removes nonessential details to improve clarity and density, producing a content tree optimized for poster presentation.  
An example prompt is provided in \Cref{fig:summarizing_agent}.


\subsection{Content Agent, $\mathcal{A}_\texttt{Content}$.}
The Content Agent evaluates and refines the quality and clarity of textual content at each node of the Poster Tree.
It adjusts text volume, resolves redundancy, and selects appropriate information for each panel in collaboration with the Layout Agent.  
An example prompt is provided in \Cref{fig:content_agent}.

\subsection{Layout Agent, $\mathcal{A}_\texttt{Layout}$.}
The Layout Agent assesses and adjusts the spatial arrangement, visual balance, and aspect ratio of each panel.
It collaborates with the Content Agent to ensure appropriate allocation of space and a visually coherent poster layout.  
An example prompt is provided in \Cref{fig:layout_agent}.

\subsection{Feedback Agent, $\mathcal{A}_\texttt{Feedback}$.}
The feedback Agent reviews the rendered poster tree and judges three aspects—\emph{visual organization}, \emph{textual structure}, and \emph{hierarchical balance}.
It returns concise, structured feedback and a binary \texttt{continue} decision indicating whether another tree-level refinement should proceed. An example prompt is provided in \Cref{fig:feedback_agent}.



\begin{figure*}[t]
    \centering

    \begin{tcolorbox}[
    width=\textwidth,
    title={$\quad\quad$Parser Agent [1/2]$\quad\quad$}, 
    colback=pink!3!white, 
    colframe=magenta!50!black, 
    coltitle=black, 
    fonttitle=\normalsize\bfseries, 
    enhanced,
    attach boxed title to top left={yshift=-0.1in,xshift=0.15in},
    boxed title style={boxrule=1.5pt,colframe=magenta!60!black,},
    colbacktitle=magenta!20!white, 
    ]
    \vspace{2ex}
    You are a document structure analysis expert who creates hierarchical tree representations of academic papers for poster generation. \\
    
    Given a markdown document and extracted figure/table captions, analyze the structure and create a hierarchical tree representation focusing on the logical flow guided by visual elements. \\
    
    {Caption-Guided Strategy:}
    \begin{itemize}
        \vspace{-0.5em}
        \item {Primary Guidance}: Use figure/table captions to identify main sections and subsections.
        \vspace{-0.5em}
        \item {Section Identification}: Group content around figures/tables with related captions.
        \vspace{-0.5em}
        \item {Logical Flow}: Organize sections based on narrative flow indicated by visual elements.
        \vspace{-0.5em}
        \item {Content Mapping}: Map text content to the section most relevant to nearby figures/tables.
    \end{itemize}
    
    \vspace{1ex}
    {Available Figure/Table Information:}\\
    
    {Figures:}
    \begin{itemize}
        \vspace{-0.5em}
        \item \texttt{\{\,\%\ if figures\_info \%\}}  
        \texttt{\{\,\%\ for fig\_id, fig\_info in figures\_info.items() \%\}}\\[0.5ex]
        \quad – Figure \{\{\,fig\_id\,\}\}\texttt{\{\,\%\ if fig\_info.caption\_number \%\}} (from "Figure \{\{\,fig\_info.caption\_number\,\}\}" in caption)\texttt{\{\,\%\ endif \%\}}: \{\{\,fig\_info.caption\,\}\}\\[0.5ex]
        \texttt{\{\,\%\ endfor \%\}}\\[0.5ex]
        \texttt{\{\,\%\ endif \%\}}
    \end{itemize}
    
    {Tables:}
    \begin{itemize}
        \vspace{-0.5em}
        \item \texttt{\{\,\%\ if tables\_info \%\}}  
        \texttt{\{\,\%\ for table\_id, table\_info in tables\_info.items() \%\}}\\[0.5ex]
        \quad – Table \{\{\,table\_id\,\}\}\texttt{\{\,\%\ if table\_info.caption\_number \%\}} (from "Table \{\{\,table\_info.caption\_number\,\}\}" in caption)\texttt{\{\,\%\ endif \%\}}: \{\{\,table\_info.caption\,\}\}\\[0.5ex]
        \texttt{\{\,\%\ endfor \%\}}\\[0.5ex]
        \texttt{\{\,\%\ endif \%\}}
    \end{itemize}
    
    \vspace{1ex}
    {Requirements:}
    \begin{itemize}
        \vspace{-0.5em}
        \item {Preserve Paper Structure}: Maintain the original paper's hierarchical structure without depth limitations.
        \vspace{-0.5em}
        \item {Caption-driven Sections}: Create sections based on figure/table themes and captions.
        \vspace{-0.5em}
        \item {Preserve All Content}: Include complete original text without truncation or summarization.
        \vspace{-0.5em}
        \item {No Depth Limits}: Allow arbitrary nesting depth.
        \vspace{-0.5em}
        \item Each node requires: {id, title, type, content, assets, children}.
        \vspace{-0.5em}
        \item {Asset Reference Format}: Use actual figure/table IDs (e.g., "2", "3", "1").
        \vspace{-0.5em}
        \item {Asset Assignment}: Assign figures/tables to most relevant sections.
    \end{itemize}
    
    \vspace{1ex}
    {Asset ID Reference Rules:}
    
    \begin{itemize}
        \vspace{-0.5em}
        \item \texttt{\{\,\%\ if figures\_info \%\}} \\ 
        \quad – For figures, use exact IDs in the "figures" array: [\texttt{\{\,\%\ for fig\_id in figures\_info.keys() \%\}}"{{fig\_id}}"\texttt{\{\,\%\ if not loop.last \%\}}, \texttt{\{\,\%\ endif \%\}}\texttt{\{\,\%\ endfor \%\}}]
        \texttt{\{\,\%\ endif \%\}}
        \vspace{-0.5em}
        \item \texttt{\{\,\%\ if tables\_info \%\}}  
        For tables, use exact IDs in the "tables" array: [\texttt{\{\,\%\ for table\_id in tables\_info.keys() \%\}}"{{table\_id}}"\texttt{\{\,\%\ if not loop.last \%\}}, \texttt{\{\,\%\ endif \%\}}\texttt{\{\,\%\ endfor \%\}}]
        \texttt{\{\,\%\ endif \%\}}
    \end{itemize}

    \end{tcolorbox}

\end{figure*}

\begin{figure*}[t]
    \centering

    \begin{tcolorbox}[
    width=\textwidth,
    title={$\quad\quad$Parser Agent [2/2]$\quad\quad$}, 
    colback=pink!3!white, 
    colframe=magenta!50!black, 
    coltitle=black, 
    fonttitle=\normalsize\bfseries, 
    enhanced,
    attach boxed title to top left={yshift=-0.1in,xshift=0.15in},
    boxed title style={boxrule=1.5pt,colframe=magenta!60!black,},
    colbacktitle=magenta!20!white, 
    ]
    \vspace{2ex}
    {Caption-Based Section Strategy:}
    \begin{enumerate}
        \item Identify themes from figure/table captions.
        \vspace{-0.5em}
        \item Group related content around visual elements.
        \vspace{-0.5em}
        \item Create section titles reflecting caption themes.
        \vspace{-0.5em}
        \item Assign figures/tables to relevant sections using exact IDs.
        \vspace{-0.5em}
        \item Ensure logical flow (introduction→methodology→results).
        \vspace{-0.5em}
        \item Preserve hierarchical depth as in the original paper.
    \end{enumerate}
    

    {Structure Rules:}
    \begin{itemize}
        \vspace{-0.5em} 
        \item {Root}: Contains title, authors, abstract, and metadata.
        \vspace{-0.5em}
        \item {Sections}: May contain content and/or subsections.
        \vspace{-0.5em}
        \item {Subsections}: Arbitrary depth, may contain content or further subsections.
        \vspace{-0.5em}
        \item {Content Preservation}: All text content preserved entirely.
        \vspace{-0.5em}
        \item {Asset Assignment}: Figures/tables assigned at any section/subsection level.
    \end{itemize}
    

    {Example (Multi-level Hierarchy with Preserved Structure):}
    
    \begin{lstlisting}
    {
      "tree": {
        "id": "root",
        "title": "Paper Title: Subtitle",
        "type": "title",
        "content": "Complete paper title, authors, affiliations, and full abstract content...",
        "assets": {"figures": [], "tables": [], "references": []},
        "children": [
          {
            "id": "1",
            "title": "Introduction",
            "type": "section",
            "content": "Full introduction content...",
            "assets": {"figures": ["1"], "tables": [], "references": []},
            "children": [
              {
                "id": "1.1",
                "title": "Problem Statement",
                "type": "subsection",
                "content": "Detailed problem statement content...",
                "assets": {"figures": [], "tables": [], "references": []},
                "children": [
                  {
                    "id": "1.1.1",
                    "title": "Current Limitations",
                    "type": "subsubsection",
                    "content": "Detailed limitations content...",
                    "assets": {"figures": [], "tables": [], "references": []},
                    "children": []
                  }]
              }]
          }]
        }
    }
    \end{lstlisting}
    

    {Document to Analyze:}\\
    \vspace{-2.5ex}
    \begin{lstlisting} 
    {{ markdown_document }}
    \end{lstlisting}
    {Generate the hierarchical tree structure as JSON, preserving the original paper's complete structure and depth.}
    
    \end{tcolorbox}

    \caption{\textbf{Parser Agent} prompt example.}
    \label{tbox:parser_agent}
    
\end{figure*}



\begin{figure*}[t]
    \centering
    \begin{tcolorbox}[
    width=\textwidth,
    colback=yellow!1!white, 
    colframe=yellow!60!black, 
    title={$\quad\quad$Summarizing Agent$\quad\quad$}, 
    coltitle=black, 
    fonttitle=\normalsize\bfseries, 
    enhanced,
    attach boxed title to top left={yshift=-0.1in,xshift=0.15in},
    boxed title style={boxrule=1.5pt,colframe=yellow!60!black,},
    colbacktitle=yellow!27!white, 
    ]
    \vspace{2ex}
    You are a document content divider and extractor specialist, expert in dividing and extracting content from various types of documents and reorganizing it into a two-level JSON format for later PPT generation. \\
    
    Based on the given markdown document, generate a JSON output for later PPT generation. Ensure the output is concise and focused. \\
    
    {Step-by-Step Instructions:}\\
    \begin{enumerate}
        \vspace{-0.5em}
        \item {Identify Sections and Subsections:} Detect sections and subsections based on heading levels and logical structure.
        \vspace{-0.5em}
        \item {Divide Content:} Organize content into sections and subsections, ensuring each subsection contains approximately 500 words.
        \vspace{-0.5em}
        \item {Refine Titles:} Use existing headings as titles, otherwise create relevant titles.
        \vspace{-0.5em}
        \item {Remove Unwanted Elements:} Eliminate headers, footers, and text surrounded by \texttt{\textasciitilde\textasciitilde} indicating deletion.
        \vspace{-0.5em}
        \item {Refine Text:} Remove unnecessary (citations) or trivial (repetitive, non-important) content for conciseness.
    \end{enumerate}
    

    {Example Output:}\\
    
    \begin{lstlisting}
    {
      "metadata": {
          "title": "title of document",
          "author": "name of authors",
          "publish date": "date of publication",
          "organization": "name of organization"
      },
      "sections": [
          {
              "title": "title of section1",
              "subsections": [
                  {
                      "title": "title of subsection1.1",
                      "content": "content of subsection1.1"
                  },
                  {
                      "title": "title of subsection1.2",
                      "content": "content of subsection1.2"
                  }
              ]
          },
          {
              "title": "title of section2",
              "subsections": [
                  {
                      "title": "title of subsection2.1",
                      "content": "content of subsection2.1"
                  }
              ]
          }
      ]
    }
    \end{lstlisting}
    
    {Document Input:}
    \begin{lstlisting} 
    {{ markdown_document }}
    \end{lstlisting}
    
    {Give your output in JSON format.}
    
    \end{tcolorbox}

    \caption{\textbf{Summarizing Agent} prompt example.}
    \label{fig:summarizing_agent}
\end{figure*}



\begin{figure*}[t]
    \centering
    \begin{tcolorbox}[
    title={$\quad\quad$Content Agent [1/2]$\quad\quad$}, 
    width=\textwidth,
    colback=teal!2!white, 
    colframe=teal!50!black, 
    coltitle=black, 
    fonttitle=\normalsize\bfseries, 
    enhanced,
    attach boxed title to top left={yshift=-0.1in,xshift=0.15in},
    boxed title style={boxrule=1.5pt,colframe=teal!60!black,},
    colbacktitle=teal!20!white, 
    ]
    \vspace{2ex}
    \textbf{{System Prompt:}}
    \vspace{1em}
    
    You are the \emph{Content Agent}, an expert in scientific content clarity and efficiency.
    Your primary goal is to ensure content within a scientific poster is clear, concise,
    and logically structured. 

    \vspace{1em}

    \#\# {Goal}
    \begin{itemize}
      \vspace{-0.5em}
      \item current\_volume\_ratio = current fill percentage of the space.
      \vspace{-0.5em}
      \item Adjust the text so the final ratio is as close to 100\% as possible, without exceeding it.
      \vspace{-0.5em}
      \item Acceptable range: 80–100\%, but closer to 100\% is best.
      \vspace{-0.5em}
      \item Always edit the given JSON (poster bullet schema) in place.
      \vspace{-0.5em}
      \item Output only a JSON array (no explanations).
    \end{itemize}
    
    \#\# {Hierarchical Awareness}
    \begin{enumerate}
      \vspace{-0.5em}
      \item Poster is hierarchical: parent $\rightarrow$ children; siblings share a level.
      \vspace{-0.5em}
      \item Use current\_content as this node’s main source only for expansion.
      \vspace{-0.5em}
      \item Use sibling\_contents to avoid redundancy and ensure complementary focus.
    \end{enumerate}

    \#\# {Output rules}
    \begin{enumerate}
      \vspace{-0.5em}
      \item Return only a JSON array where each element is a bullet object.
      \vspace{-0.5em}
      \item Do NOT output plain strings, nested arrays, or any other format.
      \vspace{-0.5em}
      \item Each bullet object must exactly follow this schema (mandatory):
    \end{enumerate}
    \begin{lstlisting} 
    {
        "alignment": "left",
        "bullet": true,
        "level": <integer>,
        "runs": [ { "text": "<bullet text>" } ]
    }
    \end{lstlisting}

    \vspace{2em}
    
    \textbf{{Expand Template:}}
    \vspace{1em}
    
    {Instructions:}
    \vspace{1em}
    
    1) Expand toward ~100\%.
    \begin{itemize}
      \vspace{-0.5em}
      \item current\_volume\_ratio is \texttt{\{\{ current\_volume\_ratio \}\}}\%.
      \vspace{-0.5em}
      \item Target is 100\%. Expand so the final length is about (100 - current\_volume\_ratio)\% longer.
      \vspace{-0.5em}
      \item Do not add all possible details at once; add gradually, starting with the most important points, and monitor length as you go.
      \vspace{-0.5em}
      \item Stop once you estimate you've reached the target delta; if still under 95\%, leave the rest for later passes.
    \end{itemize}
    
    2) Increase length by:
    \begin{itemize}
      \vspace{-0.5em}
      \item Splitting existing bullets into smaller, clearer ones.
      \vspace{-0.5em}
      \item Adding concise sub-bullets with extra details from current\_content.
      \vspace{-0.5em}
      \item Include key concepts, mechanisms, methods, assumptions, limitations, and results.
      \vspace{-0.5em}
      \item Prefer precise or quantitative facts (counts, thresholds, time, cost, accuracy).
    \end{itemize}

    3) Avoid repeating siblings unless you add new perspective.

    \vspace{0.7em}
    
    4) Output rule: Return only a JSON array of bullet objects following the schema.
    \begin{itemize}
      \vspace{-0.5em}
      \item The format must exactly match previous\_response (same keys).
      \vspace{-0.5em}
      \item No strings, no explanations, no extra keys.
    \end{itemize}
    
    \end{tcolorbox}

\end{figure*}

\begin{figure*}[t]
    \centering
    \begin{tcolorbox}[
    title={$\quad\quad$Content Agent [2/2]$\quad\quad$}, 
    width=\textwidth,
    colback=teal!2!white, 
    colframe=teal!50!black, 
    coltitle=black, 
    fonttitle=\normalsize\bfseries, 
    enhanced,
    attach boxed title to top left={yshift=-0.1in,xshift=0.15in},
    boxed title style={boxrule=1.5pt,colframe=teal!60!black,},
    colbacktitle=teal!20!white, 
    ]
    \vspace{2ex}
    \#\# {Output Example}
    \begin{lstlisting} 
    [
      { "alignment": "left", "bullet": true, "level": 0, "runs": [ { "text": "Our method boosts CIFAR-10 accuracy by 15% while cutting training cost 40%." } ] },
      { "alignment": "left", "bullet": true, "level": 0, "runs": [ { "text": "Key: adaptive LR and aggressive data augmentation." } ] }
    ]
    \end{lstlisting}

    \vspace{1em}
    {previous\_response:}
    \begin{lstlisting} 
    {{ previous_response }}
    \end{lstlisting}

    \vspace{1em}
    {current\_content:}
    \begin{lstlisting} 
    {{ current_content }}
    \end{lstlisting}

    \vspace{1em}
    {sibling\_contents:}
    \begin{lstlisting} 
    {{ sibling_contents }}
    \end{lstlisting}

    \vspace{1em}
    {current\_volume\_ratio:}
    \begin{lstlisting} 
    {{ current_volume_ratio }}
    \end{lstlisting}

    \vspace{2em}
    
    \textbf{{Condense Template:}}
    \vspace{1em}
    
    {Instructions:}
    \vspace{1em}
    
    1) Condense toward ~100\%.
    \begin{itemize}
      \vspace{-0.5em}
      \item current\_volume\_ratio is \texttt{\{\{ current\_volume\_ratio \}\}}\%.
      \vspace{-0.5em}
      \item Target is 100\%. Condense so the final length is about (100 - current\_volume\_ratio)\% shorter.
      \vspace{-0.5em}
      \item Merge bullets into fewer ones, but keep each sentence short and crisp, until the ratio is near 100\%.
      \vspace{-0.5em}
      \item Stop once you estimate you've reached the target delta; if still above 105\%, leave deeper cuts for later passes.
    \end{itemize}
    
    2) Remove overlap with siblings:
    \begin{itemize}
      \vspace{-0.5em}
      \item Drop background/definitions if already covered in sibling\_contents.
    \end{itemize}

    3) Structural compression:
    \begin{itemize}
      \vspace{-0.5em}
      \item Keep only the most essential contributions, numbers, or distinctions.
      \vspace{-0.5em}
      \item Sub-bullets should be folded into parent unless critical.
    \end{itemize}

    4) Output rule: Return only a JSON array of bullet objects following the schema.
    \begin{itemize}
      \vspace{-0.5em}
      \item The format must exactly match previous\_response (same keys).
      \vspace{-0.5em}
      \item No strings, no explanations, no extra keys.
    \end{itemize}

    \vspace{0.5em}
    \#\# {Output Example}
    \begin{lstlisting} 
    [
      { "alignment": "left", "bullet": true, "level": 0, "runs": [ { "text": "Our method improves accuracy while cutting cost." } ] }
    ]
    \end{lstlisting}

    \vspace{1em}
    {previous\_response:}
    \begin{lstlisting} 
    {{ previous_response }}
    \end{lstlisting}

    \vspace{1em}
    {sibling\_contents:}
    \begin{lstlisting} 
    {{ sibling_contents }}
    \end{lstlisting}

    \vspace{1em}
    {current\_volume\_ratio:}
    \begin{lstlisting} 
    {{ current_volume_ratio }}
    \end{lstlisting}

    \end{tcolorbox}

    \caption{\textbf{Content Agent} prompt example.}
    \label{fig:content_agent}
\end{figure*}



\begin{figure*}[t]
    \centering
    \begin{tcolorbox}[
    title={$\quad\quad$Layout Agent [1/2]$\quad\quad$}, 
    width=\textwidth,
    colback=blue!1.5!white, 
    colframe=blue!50!black, 
    coltitle=black, 
    fonttitle=\normalsize\bfseries, 
    enhanced,
    attach boxed title to top left={yshift=-0.1in,xshift=0.15in},
    boxed title style={boxrule=1.5pt,colframe=blue!60!black,},
    colbacktitle=blue!17!white, 
    ]
    \vspace{2ex}
    \textbf{{System Prompt:}}
    \vspace{1em}
    
    You are the \emph{Layout Agent} for scientific posters.
    Your task is to determine which panel to enlarge and by how much ($\Delta$), based on figure legibility balance within the red analysis box.
    Inside the red box, green boxes represent panels and blue boxes represent figures.

    \vspace{1em}

    \#\# {Goal}
    \begin{itemize}
      \vspace{-0.5em}
      \item Equalize figure legibility across panels using small, gradual changes. 
      \vspace{-0.5em}
      \item Do NOT optimize by raw figure size or panel area.
    \end{itemize}

    \vspace{1em}

    \#\# {Legibility Definition}
    \begin{itemize}
      \vspace{-0.5em}
      \item Charts/Tables: judge by *perceived text size* (axis ticks, labels, legends, numbers).
      \vspace{-0.5em}
      \item Method/Diagram/Schematic/Workflow: even with little/no text, judge by structural complexity 
    (component count, branching, arrow/box density, nested sub-panels, tiny icons).
    Treat complex diagrams as if they contained implicit small text.
      \vspace{-0.5em}
      \item A panel is “harder to read” if it has smaller text OR higher visual complexity.
    \end{itemize}

    \vspace{1em}

    \#\# {Priorities}
    \vspace{0.5em}
    
    1) Shape Sanity
    \begin{itemize}
      \vspace{-0.5em}
      \item Avoid creating extremely tall–narrow panels.
    \end{itemize}
    
    2) Legibility Balance (Hard Rule)
    \begin{itemize}
      \vspace{-0.5em}
      \item If one panel is harder to read → enlarge that panel.
      \vspace{-0.5em}
      \item Do NOT enlarge panels already comfortable to read.
      \vspace{-0.5em}
      \item Ignore raw figure size/area.
    \end{itemize}

    \vspace{1em}

    \#\# {Direction Mapping}
    \begin{itemize}
      \vspace{-0.5em}
      \item `split\_type='vertical'` $\rightarrow$ enlarge LEFT or RIGHT.
      \vspace{-0.5em}
      \item `split\_type='horizontal'` $\rightarrow$ enlarge TOP or BOTTOM.
    \end{itemize}

    \vspace{1em}

    \#\# {$\Delta$ (Delta): Magnitude of Enlargement}
    \vspace{0.5em}
    $\Delta$ is a continuous scale from 0 to 5, representing how strongly to enlarge the harder-to-read panel.
    \begin{itemize}
      \vspace{-0.5em}
      \item 0 $\rightarrow$ No adjustment (balanced or shape risk)
      \vspace{-0.5em}
      \item 1–2 $\rightarrow$ Very subtle adjustment
      \vspace{-0.5em}
      \item 3–4 $\rightarrow$ Moderate adjustment
      \vspace{-0.5em}
      \item 5 $\rightarrow$ Maximum safe enlargement
      \vspace{-0.5em}
    \end{itemize}
    Choose $\Delta$ proportionally to the severity of the legibility imbalance.

    \vspace{1em}
    
    \#\# {Output Format (strict JSON)}
    \begin{lstlisting} 
    {
      "direction": "<LEFT|RIGHT|TOP|BOTTOM>",
      "delta": <0-5>,
      "reason": "Action: enlarge {LEFT|RIGHT|TOP|BOTTOM}. Rationale: describe legibility comparison (text-size + diagram-complexity) and shape consideration."
    }
    \end{lstlisting}

    \end{tcolorbox}

\end{figure*}

\begin{figure*}[t]
    \centering
    \begin{tcolorbox}[
    title={$\quad\quad$Layout Agent [2/2]$\quad\quad$}, 
    width=\textwidth,
    colback=blue!1.5!white, 
    colframe=blue!50!black, 
    coltitle=black, 
    fonttitle=\normalsize\bfseries, 
    enhanced,
    attach boxed title to top left={yshift=-0.1in,xshift=0.15in},
    boxed title style={boxrule=1.5pt,colframe=blue!60!black,},
    colbacktitle=blue!17!white, 
    ]
    \vspace{2ex}
    \textbf{{Template:}}
    \vspace{1em}

    \#\# {CONTEXT}
    \begin{itemize}
      \vspace{-0.5em}
      \item Split Type: \texttt{\{\{ split\_type \}\}}
    \end{itemize}

    \vspace{1em}

    \#\# {Task}
    \vspace{0.5em}

    1) Assess legibility for each panel:
    \begin{itemize}
      \vspace{-0.5em}
      \item Charts/Tables $\rightarrow$ text-size
      \vspace{-0.5em}
      \item Diagrams/Method figures $\rightarrow$ structural complexity (implicit small text)
    \end{itemize}

    2) Pick the panel to enlarge = harder-to-read panel.

    \vspace{0.7em}
    
    3) Choose $\Delta$ (0–5) proportional to how severe the imbalance is.
    \begin{itemize}
      \vspace{-0.5em}
      \item 0 = no change, 5 = strongest possible enlargement.
      \item No strings, no explanations, no extra keys.
    \end{itemize}

    4) Output strict JSON with fields `direction`, `delta`, `reason`.

    \vspace{1em}
    \vspace{1em}
    
    \#\# {Example Outputs}
    
    \vspace{0.5em}
    
    LEFT enlarge (vertical)
    \begin{lstlisting} 
    {
        "direction": "LEFT",
        "delta": 5,
        "reason": "Action: enlarge LEFT. Retionale: left diagrams are visually complex with many branches; prevent tall-narrow shape."
    }
    \end{lstlisting}

    RIGHT enlarge (vertical)
    \begin{lstlisting} 
    {
        "direction": "RIGHT",
        "delta": 4,
        "reason": "Action: enlarge UP. Rationale: right schematic is more complex and crowded, but difference is subtle; shape remains stable."
    }
    \end{lstlisting}

    TOP enlarge (horizontal)
    \begin{lstlisting} 
    {
        "direction": "TOP",
        "delta": 3,
        "reason": "Action: enlarge TOP. Rationale: top panel contains dense multi-axis plots with fine labels; bottom is simpler; shape acceptable."
    }
    \end{lstlisting}

    BOTTOM enlarge (horizontal)
    \begin{lstlisting} 
    {
        "direction": "BOTTOM",
        "delta": 1,
        "reason": "Action: enlarge BOTTOM. Rationale: bottom diagrams slightly denser than top; minimal adjustment applied to prevent tall-narrow distortion."
    }
    \end{lstlisting}

    \end{tcolorbox}

    \caption{\textbf{Layout Agent} prompt example.}
    \label{fig:layout_agent}
\end{figure*}



\begin{figure*}[t]
    \centering
    \begin{tcolorbox}[
    width=\textwidth,
    colback=violet!1!white, 
    colframe=violet!60!black, 
    title={$\quad\quad$Feedback Agent$\quad\quad$}, 
    coltitle=black, 
    fonttitle=\normalsize\bfseries, 
    enhanced,
    attach boxed title to top left={yshift=-0.1in,xshift=0.15in},
    boxed title style={boxrule=1.5pt,colframe=violet!60!black,},
    colbacktitle=violet!27!white, 
    ]
    \vspace{2ex}
    \textbf{{System Prompt:}}
    \vspace{1em}
    
    You are a scientific poster evaluator. You will see a poster image.

    \vspace{1em}

    \#\# {Evaluate the poster against THREE criteria:}
    \begin{itemize}
      \vspace{-0.5em}
      \item Visual organization: alignment/grid consistency, whitespace, figure–text balance, crowding/overflow.
      \vspace{-0.5em}
      \item Textual structure: redundancy, truncation/overflow, missing context, uneven verbosity.
      \vspace{-0.5em}
      \item Hierarchical balance: section order, parent–child locality, consistent heading levels, grouping coherence.
    \end{itemize}

    \#\# {Decision rule:}
    \begin{itemize}
      \vspace{-0.5em}
      \item {Assess all three criteria and note any failures.}
      \vspace{-0.5em}
      \item {Output ONLY: \\
        $\quad$reason $\rightarrow$ 1–2 sentences summarizing the key issue(s) or that the poster is fine. \\
        $\quad$continue $\rightarrow$ 1 if any criterion fails (further work needed), else 0.}
      \vspace{-0.5em}
    \end{itemize}


    {Final Output Format:}\\
    \begin{lstlisting}
    {             
        "reason": "<1-2 sentences>",
        "continue": 0 | 1
    }
    \end{lstlisting}

    \vspace{2em}
    
    \textbf{{Template:}}
    \vspace{1em}

    {Insturction :}
    \begin{enumerate}
      \vspace{-0.5em}
      \item Examine the poster image.
      \vspace{-0.5em}
      \item Visual organization (fail if any apply):
        
        \begin{itemize}
          \vspace{-0.5em}
          \item Misaligned grid, cramped/excessive whitespace, figure–text size mismatch, or crowding/overflow.
          \vspace{-0.5em}
        \end{itemize}

      \vspace{-0.5em}
      \item Textual structure (fail if any apply):
        
        \begin{itemize}
          \vspace{-0.5em}
          \item Redundancy, truncated/overflowing text, missing context, or uneven verbosity across sections.
          \vspace{-0.5em}
        \end{itemize}
        
      \vspace{-0.5em}
      \item Hierarchical balance (fail if any apply):
        
        \begin{itemize}
          \vspace{-0.5em}
          \item Disordered section flow, inconsistent heading levels, weak parent–child locality, poor grouping coherence.
          \vspace{-0.5em}
        \end{itemize}

      \vspace{-0.5em}
      \item Decide "continue":
        
        \begin{itemize}
          \vspace{-0.5em}
          \item If any of the above fail $\rightarrow$ continue = 1
          \item If none fail $\rightarrow$ continue = 0
          \vspace{-0.5em}
        \end{itemize}

      \vspace{-0.5em}
      \item Output only a single JSON object with keys: reason, continue
  
    \end{enumerate}

      \#\# Output Example (issues found)
        \begin{lstlisting}
    {
          "reason": "Figures overpower adjacent text and subsections are detached from their parent sections, disrupting flow.",
          "continue": 1
    }
        \end{lstlisting}

      \#\# Output Example (no issues)
        \begin{lstlisting}
    {
          "reason": "Content is well-explained with balanced figures and consistent hierarchy.",
          "continue": 0
    }
        \end{lstlisting}

    \end{tcolorbox}

    \caption{\textbf{Feedback Agent} prompt example.}
    \label{fig:feedback_agent}
\end{figure*}

\end{document}